\documentclass{article}

\usepackage{microtype}
\usepackage{graphicx}
\usepackage{subcaption}
\usepackage{booktabs} 
\usepackage{multirow}
\usepackage{amssymb}
\usepackage{hyperref}
\usepackage{pgfplots}
\pgfplotsset{compat=1.18}
\usepackage[table]{xcolor}

\usepackage[accepted]{icml2026}

\usepackage{amsmath}
\usepackage{amssymb}
\usepackage{mathtools}
\usepackage{amsthm}
\usepackage{cancel}
\usepackage{enumitem}

\usepackage[capitalize,noabbrev]{cleveref}

\theoremstyle{plain}

\theoremstyle{definition}

\theoremstyle{remark}

\usepackage{float}
\usepackage{comment}
\usepackage{bm}
\usepackage[most]{tcolorbox}
\tcbuselibrary{skins,breakable}
\usepackage{xcolor}
\definecolor{lightlavender}{HTML}{E6E6FA}   
\definecolor{lavenderframe}{HTML}{9B8CFF}   
\newtcolorbox{prompt}{
  enhanced,breakable,
  colback=black!5,
  colframe=black!40,
  boxrule=0.4pt,
  borderline west={1.75pt}{0pt}{black!60},
  title=Prompt,
  left=6pt,right=6pt,top=6pt,bottom=6pt
}
\newtcolorbox{answer}{
  enhanced,breakable,
  colback=black!5,
  colframe=black!40,
  boxrule=0.4pt,
  borderline west={1.75pt}{0pt}{black!60},
  title=Answer,
  left=6pt,right=6pt,top=6pt,bottom=6pt
}

\usepackage[textsize=tiny]{todonotes}

\icmltitlerunning{GuidedBridge: Training-freely Improving Bridge Models with Prior Guidance}

\begin{document}

\twocolumn[
\icmltitle{GuidedBridge: Training-freely Improving Bridge Models with Prior Guidance}

\icmlsetsymbol{equal}{*}

\begin{icmlauthorlist}
\icmlauthor{Zehua Chen}{equal,thu}
\icmlauthor{Yucheng Yang}{equal,thu}
\icmlauthor{Binjie Yuan}{thu}
\icmlauthor{Kaiwen Zheng}{thu}
\icmlauthor{Jun S. Liu}{thu}
\icmlauthor{Jun Zhu}{thu} 
\end{icmlauthorlist}

\icmlaffiliation{thu}{Tsinghua University, Beijing, China}
\icmlcorrespondingauthor{Zehua Chen}{zhc23thuml@mail.tsinghua.edu.cn}
\icmlcorrespondingauthor{Jun Zhu}{dcszj@mail.tsinghua.edu.cn}

\icmlkeywords{Machine Learning, ICML}

\vskip 0.3in
]

\printAffiliationsAndNotice{\icmlEqualContribution \\
}

\begin{abstract}
Guidance methods, such as classifier-free guidance (CFG) and auto-guidance (AG), have advanced noise-to-data generation in diffusion models. 
Recently, bridge models have introduced a data-to-data generative process that can exploit an instructive clean prior.
In this work, inspired by previous methods creating quality difference between denoising results as guidance, we propose a training-free bridge guidance method, termed Prior Guidance (PG). 
Specifically, we introduce a weak prior, which is unseen during bridge pre-training, hindering prior exploitation and thereby degrading denoising result. 
Then, we contrast it with the seen prior to highlight and enhance prior exploitation via a scaling factor.
Moreover, we analyze the underlying mechanism of prior exploitation in the bridge process and design frequency-modulated prior guidance (FMPG), which tailors the guidance scale to low- and high-frequency bands coherent with bridge generative dynamics.
To address prior exploitation in image in-painting, we develop a cascaded framework, CFG-FMPG, which first generates a noisy hidden representation via CFG and then exploits it as a generative prior with FMPG, fulfilling their complementary strengths without compromising inference efficiency. Experiments demonstrate that our PG methods consistently improve pre-trained bridge models across diverse image translation tasks.
\end{abstract}

\section{Introduction}
Diffusion models~\cite{NEURIPS2020_4c5bcfec,SGM} have been a widely adopted framework for generative modeling~\cite{InferGrad,BinauralGrad,ResGrad,audioldm,Tiva,DiffGAP,RespDiff,freeaudio,Omni2Sound,UniCardio}, where they can faithfully reconstruct the target distribution from a known prior distribution,~\textit{e.g.}, standard Gaussian, with time-dependent score functions learned at model training stage.
In conditional generation tasks, a guidance method, classifier-free guidance (CFG)~\cite{ho2022classifier}, was developed, which enhances condition alignment by extrapolating two denoising results,~\textit{i.e.}, an unconditional and a conditional one, at each inference step.
Recently, another guidance method, auto-guidance (AG)~\cite{NEURIPS2024_5ee7ed60}, has been developed, which improves the accuracy of score estimation by extrapolating two denoising results: a result estimated with a well-trained network and a result estimated with an under-trained one.
By creating quality differences in denoising results, these guidance methods have emphasized condition alignment and score accuracy, respectively, strengthening the generation quality of diffusion models in diverse generation tasks, such as image~\cite{stablediffusion,sd3}, audio~\cite{audioldm,freeaudio} and video generation~\cite{svd,framebridge}.

Given the noise-to-data sampling nature of diffusion models, they often suffer from an increased sampling burden in tasks that have already been provided with strong prior information~\cite{FBSDEBridge,bridge-tts,GSBM}, such as image-to-image translation~\cite{liu2023i2sb}. 
To address this limitation, bridge models~\cite{liu2023i2sb,bridge-tts,zhou2024ddbm,zheng2025diffusion,CDBM,zhang2025exploring} have introduced a data-to-data generative framework that is naturally aligned with such tasks. 
By directly exploiting the instructive information contained in clean prior representations, bridge models alleviate the sampling burden and achieve improved performance over conditional diffusion models in applications including image-to-video generation~\cite{framebridge}, audio super-resolution~\cite{Bridge-SR,audiolbm} as well as signal restoration~\cite{voicebridge, RefineBridge}.

\begin{figure*}[t!]
    \centering
    \includegraphics[width=\textwidth]{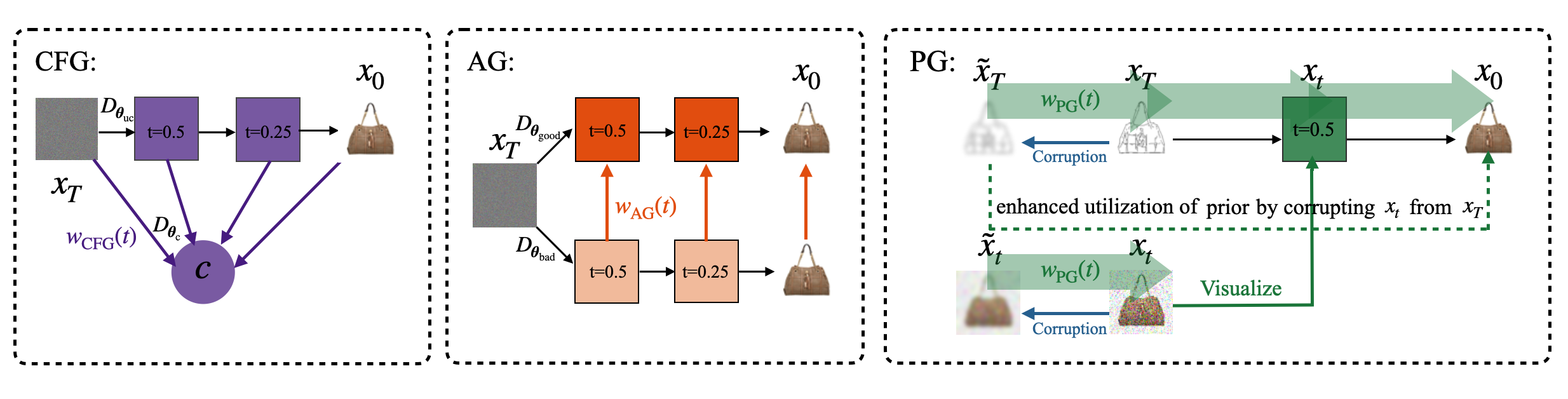}
    \caption{\textbf{Overview of guidance strategies.} Classifier-free guidance (CFG, left) enhances condition alignment by extrapolating an unconditional denoising result $D_{\bm{\theta_{\text{uc}}}}$ and a conditional denoising result $D_{\bm{\theta_{\text{c}}}}$. Auto-guidance (AG, middle) improves sample quality by contrasting a full-capacity denoiser $D_{\bm{\theta_{\text{good}}}}$ against a less-capable denoiser $D_{\bm{\theta_{\text{bad}}}}$. Our proposed prior guidance (PG, right) encourages prior exploitation by corrupting the prior representation at each bridge sampling step from $t=T$, forcing the model to strengthen its utilization of informative prior throughout the sampling trajectory. The bottom-right panel visualizes the effect of corruption on $\bm{x}_t$.}
    \label{fig:overview}
\end{figure*}

Although diffusion guidance methods can be applied to bridge models, \textit{e.g.}, improving text alignment with CFG in bridge-based image-to-video generation~\cite{framebridge}, guidance methods tailored to bridge generative dynamics remain largely unexplored. 
To address this gap, we propose a novel \textbf{training-free} guidance method for bridge generative frameworks, termed \textbf{Prior Guidance (PG)}. 
Previous work shows that CFG and AG create quality differences between two denoising results to enable guidance: in CFG, unconditional denoising typically fits the data worse than conditional denoising, while in AG under-trained models produce larger errors than well-trained ones~\cite{ho2022classifier,NEURIPS2024_5ee7ed60,audiomog}. 
Motivated by the strong prior exploitation ability of bridge models, PG further amplifies this property as a guidance method to improve generation quality. 
Given a pre-trained bridge model that generates the target from an informative prior, we introduce an additional \textbf{weak prior}, which can be constructed via degradation operations, e.g., additional noise, blurring, or JPEG compression, in a training-free manner. 
As this weak prior is unseen during bridge training and provides limited target information, it increases the difficulty of prior exploitation and produces a lower-quality denoising result. 
We then extrapolate the two denoising results from the seen prior and the weak prior, respectively, to encourage stronger exploitation of instructive prior information at each inference step.

Furthermore, we investigate the mechanism of prior exploitation along the bridge generative process and strengthen PG through frequency modulation, termed \textbf{FMPG}. Unlike diffusion models, whose signal-to-noise ratio (SNR) increases monotonically along the noise-to-data sampling trajectory~\cite{NEURIPS2020_4c5bcfec,SGM}, bridge models~\cite{liu2023i2sb,bridge-tts,zhou2024ddbm,zheng2025diffusion} exhibit a U-shaped SNR due to clean representations at both boundary distributions~\textit{i.e.}, data-to-data generation. Consequently, high-frequency (HF) prior information is mainly exploited at early and late sampling stages, where hidden representations are moderately corrupted by noise. In contrast, low-frequency (LF) prior information remains relatively stable even at the time step with a minimum SNR value and can therefore be exploited throughout the sampling trajectory. Based on this observation, we design FMPG, which compresses the guidance scale for HF prior information with a U-shaped schedule while amplifying the guidance scale for LF prior information with an inverted U-shaped profile, aligning guidance with the generative dynamics of bridge models.

As discussed, given an informative prior, PG enhances prior exploitation for pre-trained bridge models by constructing a weak prior that produces a lower-quality denoising result. 
However, when the provided prior itself is difficult to exploit, for example, the masked region in the image inpainting task~\cite{zheng2025diffusion}, it may be difficult to construct a meaningful weak prior. 
To address this limitation, we propose a cascaded guidance framework, \textbf{CFG-FMPG}, which fulfills the complementary strengths of CFG and FMPG along the sampling trajectory. 
Specifically, at the early sampling stage, CFG reconstructs masked regions under semantic guidance, producing a noisy hidden representation that captures the coarse structure of the target. Given this representation, the guidance method is switched to FMPG to fully exploit the prior information in the following inference steps. This design assigns time-varying scaling factors to CFG and FMPG, respectively, guiding large-scale structure reconstruction with semantic guidance and high-quality refinement with prior exploitation, without sacrificing sampling efficiency.

Our contributions are summarized as follows.
\begin{itemize}[nosep]
\item We introduce \textbf{PG}, a training-free guidance method that boosts bridge generation performance by emphasizing prior exploitation.
\item We propose \textbf{FMPG}, which aligns PG with the prior exploitation mechanism of bridge generative dynamics, through frequency modulation.
\item We develop \textbf{CFG-FMPG}, a cascaded framework that leverages CFG to provide coarse prior information for FMPG, thereby complementing their strengths without compromising inference speed.
\item Extensive experiments demonstrate that PG methods consistently improve pre-trained bridge models, including DDBM~\cite{zhou2024ddbm} and DBIM~\cite{zheng2025diffusion}, across diverse image translation tasks.
\end{itemize}

\section{Background}
\paragraph{Diffusion Models.}
Diffusion generative framework~\cite{NEURIPS2020_4c5bcfec,SGM} is composed of two mirror processes. A forward process transforms a data distribution $p_0(\bm{x}_0)$ into a known prior distribution $p_T(\bm{x}_T)$,~\textit{e.g.}, standard Gaussian distribution $\mathcal{N}(\bm{0},\bm{I})$, through a stochastic differential equation (SDE)~\cite{SGM}: 
\begin{equation}
\label{diffsde}
    \mathrm{d} \bm{x}_t = \bm{f}(t)\bm{x}_t\mathrm{d}t + g(t)\mathrm{d} \bm{w}_t,
\end{equation}
where $t\in [0, T]$, $\bm{f}$ and $g$ are predefined coefficients, $\bm{w}_t$ and $\bm{x}_t$ are the standard Wiener process and the noisy hidden representations, respectively, sharing the same data dimension with the generation target $\bm{x}_0\in \mathbb{R}^d$.
In sampling, a reverse process reconstructs the data distribution $p_0(\bm{x}_0)$ from the prior distribution $p_T(\bm{x}_T)=\mathcal{N}(\bm{0},\bm{I})$ by solving a reverse SDE, which shares the same marginal distribution with $p_t(\bm{x}_t)$ defined in Equation~\eqref{diffsde}:  
\begin{equation}
\label{diffisde}
    \mathrm{d} \bm{x}_t = [\bm{f}(t)\bm{x}_t - g^2(t)\nabla_{\bm{x}_t}\log p_t(\bm{x}_t)]\mathrm{d}t + g(t)\mathrm{d} \bar{\bm{w}}_t.
\end{equation}
In the model training stage, time-dependent score functions $\nabla_{\bm{x}_t}\log p_t(\bm{x}_t)$ can be learned by a denoising network $D_{\bm{\theta}}$ with:
\begin{equation}
\label{diffobj}
    \arg\min_{\bm{\theta}} \mathbb{E}_{\bm{x}_0, t, \bm{\epsilon}} \left\| \bm x_0 - D_{\bm{\theta}} \left( \bm{x}_t, t\right) \right\|_2^2,
\end{equation}
where the data predictor $D_{\bm{\theta}} \left( \bm{x}_t, t\right)$ is an alternative reparameterization method of score estimation~\cite{NEURIPS2020_4c5bcfec,SGM,NEURIPS2022_a98846e9}.

\paragraph{Classifier-free guidance.}
In conditional generation tasks, diffusion models can be trained with an additional network input, namely condition signal $\bm{c}$, learning the conditional denoising results $D_{\bm{\theta}} \left(\bm{x}_t, t, \bm{c}\right)$.
However, in practice, approximation errors caused by limited network capacity can lead to unlikely generated samples,~\textit{i.e.}, outliers, which inevitably restricts conditional generation quality~\cite{NEURIPS2024_5ee7ed60,audiomog}. 
As a solution, one of the most popular methods is CFG~\cite{ho2022classifier}, a guidance algorithm that jointly models conditional and unconditional denoising results in the training stage, and then extrapolates them with a scaling parameter $w_{\text{CFG}}$ at each inference step:
\begin{equation}
\begin{aligned}
\label{CFG}
    & D_{\bm{\text{CFG}}}(\bm{x}_t, t, \bm c) \\
    &= D_{\bm{\theta_{\text{uc}}}}(\bm{x}_t, t) + w_{\text{CFG}} \left( D_{\bm{\theta_{\text{c}}}}(\bm{x}_t, t, \bm c) -  D_{\bm{\theta_{\text{uc}}}}(\bm{x}_t, t) \right).
\end{aligned}
\end{equation}
Since $D_{\bm{\theta_{\text{uc}}}}$ tackles a more difficult task with a small training ratio, it often fits the data distribution worse and can be viewed as a low-quality denoising result~\cite{NEURIPS2024_5ee7ed60}.
By extrapolating high-quality conditional and low-quality unconditional denoising results, CFG strengthens condition alignment and eliminates outliers, thereby improving generation results across conditional tasks~\cite{ho2022classifier,stablediffusion,audioldm}.

\paragraph{Auto-guidance.}
Recently, AG has been proposed~\cite{NEURIPS2024_5ee7ed60}, a guidance method that achieves quality improvement by extrapolating the high-quality denoising result of a well-trained model $D_{\bm{\theta_{\text{good}}}}$ and the low-quality denoising result of an under-trained model $D_{\bm{\theta_{\text{bad}}}}$:
\begin{equation}
\begin{aligned}
\label{AG}
& D_{\bm{\text{AG}}}(\bm x_t, t, \bm{c}) \\
& = D_{\bm{\theta_{\text{bad}}}}(\bm x_t, t, \bm{c}) + w_{\text{AG}} \left(D_{\bm{\theta_{\text{good}}}}(\bm x_t, t, \bm{c}) - D_{\bm{\theta_{\text{bad}}}}(\bm x_t, t, \bm{c}) \right).
\end{aligned}
\end{equation}
Specifically, AG additionally trains an under-trained denoising network $D_{\bm{\theta_{\text{bad}}}}$ on the same task, condition signal, and data distribution as the well-trained network $D_{\bm{\theta_{\text{good}}}}$.
These settings ensure that these two models make similar approximation errors, and $D_{\bm{\theta_{\text{bad}}}}$ makes the error even stronger.
Therefore, measuring and emphasizing their differences can generally indicate and lead to a direction towards better generation results~\cite{NEURIPS2024_5ee7ed60,audiomog}.

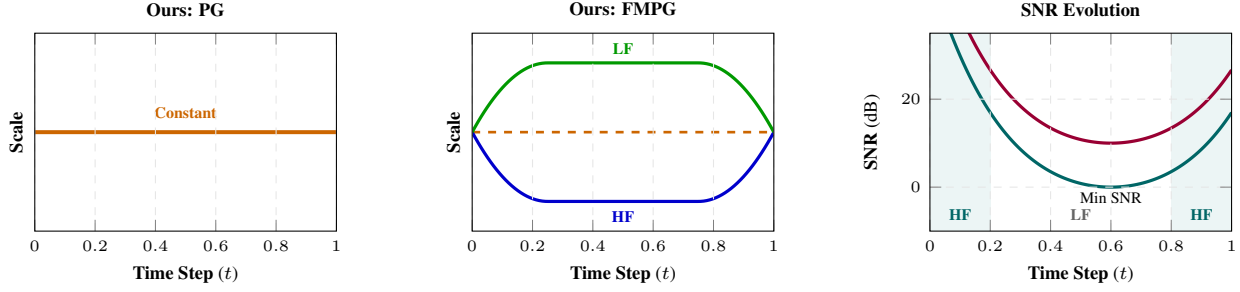
\begin{figure*}[t!]
    \centering
    \setlength{\tabcolsep}{1pt}
    \begin{minipage}[b]{0.325\linewidth}
        \centering
        \begin{tikzpicture}
            \begin{axis}[
                width=\linewidth,
                height=4.2cm, 
                xlabel={\textbf{Time Step} ($t$)},
                ylabel={\textbf{Scale}},
                xmin=0, xmax=1,
                ymin=14, ymax=22,
                ytick=\empty,
                enlarge x limits=false,
                grid=major,
                grid style={dashed, gray!20},
                title={\textbf{Ours: PG}},
                title style={font=\scriptsize, yshift=-3pt},
                label style={font=\scriptsize, yshift=1pt},
                tick label style={font=\tiny},
                axis on top
            ]
                \addplot[domain=0:1, samples=2, color=orange!80!black, line width=1.5pt] { 18.0 };
                \node[anchor=south, color=orange!80!black, font=\tiny] at (axis cs: 0.5, 18.2) {\textbf{Constant}};
            \end{axis}
        \end{tikzpicture}
    \end{minipage}
    \hfill
    \begin{minipage}[b]{0.325\linewidth}
    \centering
    \begin{tikzpicture}
        \begin{axis}[
            width=\linewidth,
            height=4.2cm,
            xlabel={\textbf{Time Step} ($t$)},
            ylabel={\textbf{Scale}},
            xmin=0, xmax=1,
            ymin=8, ymax=28, 
            ytick=\empty,
            enlarge x limits=false,
            grid=major,
            grid style={dashed, gray!20},
            title={\textbf{Ours: FMPG}},
            title style={font=\scriptsize, yshift=-3pt},
            label style={font=\scriptsize, yshift=1pt},
            tick label style={font=\tiny},
            axis on top
        ]
            \addplot[domain=0:1, samples=2, color=orange!80!black, line width=1.0pt, dashed, forget plot] { 18.0 };
            
            \addplot[domain=0:1, samples=100, color=blue!80!black, line width=1.2pt]
            { (x < 0.25) * (11.0 + 7.0*((0.25-x)/0.25)^2) + (x > 0.75) * (11.0 + 7.0*((x-0.75)/0.25)^2) + (x >= 0.25 && x <= 0.75) * 11.0 };
            \node[anchor=north, color=blue!80!black, font=\tiny\bfseries] at (axis cs: 0.5, 11.0) {HF};
            
            \addplot[domain=0:1, samples=100, color=green!60!black, line width=1.2pt]
            { (x < 0.25) * (25.0 - 7.0*((0.25-x)/0.25)^2) + (x > 0.75) * (25.0 - 7.0*((x-0.75)/0.25)^2) + (x >= 0.25 && x <= 0.75) * 25.0 };
            \node[anchor=south, color=green!60!black, font=\tiny\bfseries] at (axis cs: 0.5, 25.0) {LF};
            
        \end{axis}
    \end{tikzpicture}
\end{minipage}
    \hfill
    \begin{minipage}[b]{0.325\linewidth}
        \centering
        \begin{tikzpicture}
            \begin{axis}[
                width=\linewidth,
                height=4.2cm,
                xlabel={\textbf{Time Step} ($t$)},
                ylabel={\textbf{SNR} (dB)},
                xmin=0, xmax=1,
                ymin=-10, ymax=35,
                enlarge x limits=false,
                grid=major,
                grid style={dashed, gray!20},
                title={\textbf{SNR Evolution}},
                title style={font=\scriptsize, yshift=-3pt},
                label style={font=\scriptsize, yshift=1pt},
                tick label style={font=\tiny},
                axis on top
            ]
                \fill[teal!15, opacity=0.5] (axis cs:0,-10) rectangle (axis cs:0.2,35);
                \fill[teal!15, opacity=0.5] (axis cs:0.8,-10) rectangle (axis cs:1,35);

                \node[anchor=south, font=\tiny\bfseries, color=teal!80!black] at (axis cs: 0.1, -9.5) {HF};
                \node[anchor=south, font=\tiny\bfseries, color=teal!80!black] at (axis cs: 0.9, -9.5) {HF};
                \node[anchor=south, font=\tiny\bfseries, color=gray!80!black] at (axis cs: 0.5, -9.5) {LF};

                \addplot[domain=0:1, samples=100, color=purple!80!black, line width=1.2pt, forget plot] { 10 + 80 * (x-0.6)^2 + 150 * (x-0.6)^4 };

                \addplot[domain=0:1, samples=100, color=teal!80!black, line width=1.2pt] { 82 * (x-0.6)^2 + 150 * (x-0.6)^4 };
                \node[anchor=north, align=center, color=black, font=\tiny] at (axis cs: 0.6, 1.0) {Min SNR};
            \end{axis}
        \end{tikzpicture}
    \end{minipage}

    \caption{Guidance scale comparison and signal-to-noise (SNR) evolution. FMPG (Middle) adapts guidance to high-frequency (HF) and low-frequency (LF) bands, mirroring the typical U-shaped SNR (Right) observed in bridge models, whereas PG (Left) employs a constant scale and does not account for frequency dynamics.}
    \label{fig:guidance_schedules}
\end{figure*}

\begin{figure}[t!]
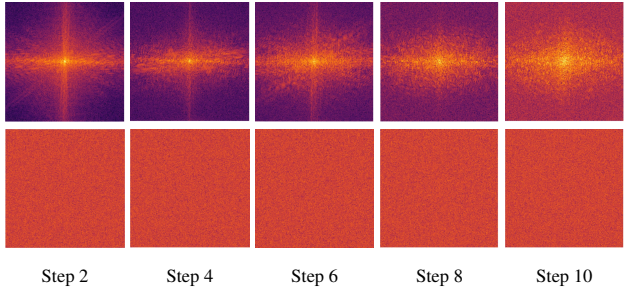

    \centering
    \foreach \step in {2,4,6,8,10} {%
        \begin{subfigure}[b]{0.19\linewidth}
            \centering
            \includegraphics[width=\linewidth, height=\linewidth]{img/x0_step\step.jpg}%
        \end{subfigure}\hfill%
    }
    
    \vspace{2pt}
    
    \foreach \step in {2,4,6,8,10} {%
        \begin{subfigure}[b]{0.19\linewidth}
            \centering
            \includegraphics[width=\linewidth, height=\linewidth]{img/xt_step\step.jpg}%
            \caption*{\scriptsize Step \step}
        \end{subfigure}\hfill%
    }
    \caption{Frequency energy distribution of residuals. This plot maps the energy transfer from the input residual after extra noise addition ($\Delta \bm{x}_t$ shown in the second row of figures) to the output residual ($\Delta \bm{x}_0$ shown in the first row of figures). Brighter colors indicate higher energy.}
    \label{fig:visual_frequency_evolution}
\end{figure}

\section{Method}

\subsection{Motivation}
\label{motivation}
As discussed, CFG and AG measure and boost quality differences arising from conditional information and model training, respectively, thereby strengthening condition alignment and score accuracy. 
Hence, their advantages are theoretically orthogonal to the~\textit{noise-to-data} generative framework of diffusion models~\cite{NEURIPS2020_4c5bcfec,SGM} and the~\textit{data-to-data} one of bridge models~\cite{zhou2024ddbm,bridge-tts}, and can be leveraged to guide bridge generation~\cite{framebridge}.
However, the key difference between diffusion and bridge models, namely~\textit{prior exploitation}, which could expand the design space for guidance strategies, has not been thoroughly explored in existing approaches. 
In this work, we present PG, a bridge guidance framework that creates quality differences by constructing a weak prior unseen during bridge pre-training. 
By further amplifying this difference and aligning it with the underlying mechanism of prior exploitation in the bridge generative process, PG explicitly emphasizes prior utilization along the sampling trajectory, leading to \textit{training-free} improvements of bridge generation quality without sacrificing sampling efficiency~\cite{zhou2024ddbm,zheng2025diffusion}.

\subsection{Bridge models}
\label{bridgemodels}
\textbf{Generative framework.} 
Different from the~\textit{noise-to-data} diffusion generation process shown in Equation~\eqref{diffsde} and Equation~\eqref{diffisde}, bridge models~\cite{bridge-tts,zhou2024ddbm,zheng2025diffusion} learn a~\textit{data-to-data} process between a prior $p_T(\bm{x}_T)\sim p_{\text{prior}}$ and the target $p_0(\bm{x}_0)\sim p_{\text{data}}$.
Specifically, conditioned on the clean prior representation $\bm{x}_T$, the forward process is modified by a drift term, ensuring that the trajectory reaches $\bm{x}_T$ at the endpoint $t=T$. 
\begin{equation}
\label{bridgesde}
    \mathrm{d} \bm{x}_t = [\bm{f}(\bm{x}_t, t) + g^2(t)\bm{h}(\bm{x}_t, t, \bm{x}_T)]\mathrm{d}t + g(t)\mathrm{d} \bm{w}_t,
\end{equation}
where $\bm{h}(\bm{x}_t, t, \bm{x}_T) = \nabla_{\bm{x}_t}\log p(\bm{x}_T|\bm{x}_t)$ serves as a guiding drift derived from the pre-defined transition kernel.
In generation, we reverse this process starting from $\bm{x}_T$. The time-reversed SDE is given by:
\begin{equation}
\label{bridgeisde}
\begin{split}
    \mathrm{d} \bm{x}_t = & [\bm{f}(\bm{x}_t, t) -  g^2(t)(\bm{s_{\theta}}(\bm{x}_t, t, \bm{x}_T) \\
    & - \bm{h}(\bm{x}_t, t, \bm{x}_T))]\mathrm{d}t 
     + g(t)\mathrm{d} \bar{\bm{w}}_t,
\end{split}
\end{equation}
where $\bm{s_{\theta}}(\bm{x}_t, t, \bm{x}_T)$ is the score function approximating $\nabla_{\bm{x}_t}\log p_t(\bm{x}_t|\bm{x}_T)$. 
Similar to the diffusion training objective shown in Equation~\eqref{diffobj}, the bridge score function can be reparameterized with data prediction $D_{\bm{\theta}}$ and trained with a denoising objective:
\begin{equation}
\label{bridgeobj}
    \arg\min_{\bm{\theta}} \mathbb{E}_{\bm{x}_0, \bm{x}_T, t, \bm{\epsilon}} \left\| \bm{x}_0 - D_{\bm{\theta}} \left( \bm{x}_t, t, \bm{x}_T \right) \right\|_2^2.
\end{equation}

\textbf{Prior Exploitation.} 
As shown in Equation~\eqref{diffisde} and Equation~\eqref{bridgeisde}, one of the key differences between the diffusion and bridge generative frameworks lies in~\textit{prior exploitation}. 
Namely, by conditioning the marginal distribution on clean prior representation $\bm{x}_T$, bridge models disentangle the additive Gaussian noise with prior distribution, allowing strong exploitation of prior information provided by $p_T(\bm{x}_T)$ at each generation step $p_{s|t}(\bm{x}_s|\bm{x}_t,\bm{x}_T), 0\leq s<t\leq T$.
Therefore, in tasks with an informative prior, such as image-to-image translation~\cite{liu2023i2sb}, image-to-video generation~\cite{framebridge}, and audio super-resolution~\cite{audiolbm}, bridge models can directly start sampling from the provided prior, reducing the burden of generative modeling and leading to improved results.

\begin{table*}[t]
\caption{Quantitative comparison on Edges$\to$Handbags and DIODE datasets. We report FID ($\downarrow$), IS ($\uparrow$), LPIPS ($\downarrow$), and MSE ($\downarrow$). Our method achieves superior performance across multiple metrics.}
\label{tab:combined_quantitative_detailed}
\begin{center}
\begin{scriptsize} 
\begin{sc}
\setlength{\tabcolsep}{5pt}
\begin{tabular}{l c cccc c cccc}
\toprule
& & \multicolumn{4}{c}{Edges$\to$Handbags ($64 \times 64$)} & & \multicolumn{4}{c}{DIODE-Outdoor ($256 \times 256$)} \\
\cmidrule(r){3-6} \cmidrule(l){8-11}
Method & NFE & FID $\downarrow$ & IS $\uparrow$ & LPIPS $\downarrow$ & MSE $\downarrow$ & & FID $\downarrow$ & IS $\uparrow$ & LPIPS $\downarrow$ & MSE $\downarrow$ \\
\midrule
\rowcolor{gray!10} DDIB ~\cite{su2023dual} & $\ge 40^\dagger$ & 186.84 & 2.04 & 0.869 & 1.05 & & 242.3 & 4.22 & 0.798 & 0.794 \\
\rowcolor{gray!10} SDEdit ~\cite{meng2022sdedit} & $\ge 40$ & 26.5 & 3.58 & 0.271 & 0.510 & & 31.14 & 5.70 & 0.714 & 0.534 \\
Pix2Pix \citep{Isola_2017_CVPR} & 1 & 74.8 & 3.24 & 0.356 & 0.209 & & 82.4 & 4.22 & 0.556 & 0.133 \\
I$^2$SB \citep{liu2023i2sb} & $\ge 40$ & 7.43 & 3.40 & 0.244 & 0.191 & & 9.34 & 5.77 & 0.373 & 0.145 \\
DDBM \citep{zhou2024ddbm} & 118 & 1.83 & \textbf{3.73} & 0.142 & 0.040 & & 4.43 & \textbf{6.21} & 0.244 & 0.084 \\
DDBM \citep{zhou2024ddbm} & 200 & 0.88 & 3.69 & 0.110 & 0.006 & & 3.34 & 5.95 & 0.215 & 0.020 \\
DBIM ~\cite{zheng2025diffusion} & 20 & 1.74 & 3.63 & 0.095 & 
0.005 & & 4.99 & 6.10 & 0.201 & 0.017 \\
\textbf{DBIM+FMPG (Ours)} & 20 & 1.07 & 3.69 & \textbf{0.093} & \textbf{0.005} & & 3.20 & 6.09 & 0.199 & \textbf{0.017} \\
DBIM ~\cite{zheng2025diffusion} & 100 & 0.91 & 3.62 & 0.100 & 0.006 & & 2.58 & 6.06 & 0.198 & 0.018 \\
\textbf{DBIM+FMPG (Ours)} & 100 & \textbf{0.78} & 3.66 & 0.101 & \textbf{0.005} & & \textbf{2.06} & 6.03 & \textbf{0.197} & \textbf{0.017} \\
\bottomrule
\end{tabular}
\end{sc}
\end{scriptsize}
\end{center}
\vspace{-0.3cm}
\end{table*}

\subsection{Prior Guidance}
\label{priorguidance}

\textbf{Advantage emphasis.} 
Considering the advantage of bridge models over diffusion models in prior exploitation, we develop the PG method, further emphasizing
this advantage to improve bridge generation quality.
Given a pre-trained bridge model that has learned to generate the target $\bm x_{0}$ from the prior $\bm x_{T}$, we first construct a weak prior representation, $\mathcal{H}(\bm x_{T})$, at the endpoint $t=T$ with a degradation operator $\mathcal{H}$.
Then, at each following inference step $p_{\bm \theta}(\bm{x}_{s}|\bm{x}_{t},\bm{x}_T)$, as $\bm{x}_t$ has been an updated prior for the generation target $\bm x_{0}$, we construct $\mathcal{H}(\bm x_{t})$ as the weak prior for each hidden representation.
By constructing $\mathcal{H}(\bm x_{T})$ from the first sampling step and $\mathcal{H}(\bm x_{t})$ along sampling trajectory, we degrade the prior information that could be exploited by the pre-trained bridge model, resulting in a degraded denoising result $D_{\bm{\theta}}(\mathcal{H}(\bm x_{t}), t, \bm x_T)$.

In practice, we measure the produced quality difference and extrapolate the high-quality denoising result $D_{\bm{\theta}}(\bm x_{t}, t)$ and the low-quality one $D_{\bm{\theta}}(\mathcal{H}(\bm x_{t}), t)$ with a scaling factor $w_{\text{PG}}$, to emphasize prior exploitation. 
\begin{equation}
\begin{aligned}
\label{PG}
D_{\bm{\text{PG}}}(\bm x_t, t, \bm x_T) & = D_{\bm{\theta}}(\mathcal{H}(\bm x_{t}), t, \bm x_T) + w_{\text{PG}} \\ 
& \left(D_{\bm{\theta}}(\bm x_{t}, t, \bm x_T) 
- D_{\bm{\theta}}(\mathcal{H}(\bm x_{t}), t, \bm x_T) \right),
\end{aligned}
\end{equation}
where $t\in [T,0)$. 
The condition signal $\bm x_T$ in the constructed low-quality term $D_{\bm{\theta}}(\mathcal{H}(\bm x_{t}), t, \bm x_T)$ is preserved as a clean representation without degradation, which controls the same condition information as the high-quality term $D_{\bm{\theta}}(\bm x_{t}, t, \bm x_T)$, ensuring that the quality difference is caused by prior degradation.

\paragraph{Degradation operator.}
Different from AG~\cite{NEURIPS2024_5ee7ed60} that requires an additional network to make stronger errors than the good model, PG can construct $\mathcal{H}(\bm x_{t})$ in a training-free manner, where the degradation operator $\mathcal{H}$ can be selected from a group of methods, such as additional noise injection, blurring, and JPEG compression.
In this work, considering the Gaussian transition kernel of bridge models~\cite{liu2023i2sb,zhou2024ddbm,zheng2025diffusion,zhang2025exploring}, we align $\mathcal{H}$ with it, namely, adding extra Gaussian noise $\bm{\epsilon}\sim \mathcal{N}(\bm{0},\bm{I})$ to the clean prior representation $\bm x_{T}$ and each generated noisy hidden representation $\bm x_{t}$.

As shown in Figure~\ref{fig:visual_frequency_evolution}, Gaussian noise in the time domain corresponds to a flat power spectral density in the frequency domain, meaning that it injects equal expected energy across all frequency bands. Consequently, adding extra Gaussian noise to the clean prior and the noisy hidden representations uniformly corrupts all spectral components of the signal, effectively weakening the ability of pretrained bridge models to exploit prior information and producing low-quality denoising result.
Empirically, we find that PG effects are robust to other degradation methods, such as blurring and JPEG compression. These additional experimental results are provided in Appendix~\ref{app:corruption_details}.

\begin{table*}[t]
\caption{Quantitative comparison on Edges$\to$Handbags and DIODE datasets. We report FID ($\downarrow$), demonstrating that our guidance method outperforms baseline inference methods across both datasets given the same NFE.}
\label{tab:combined_quantitative_compact}
\begin{center}
\begin{scriptsize} 
\begin{sc}
\setlength{\tabcolsep}{4pt} 
\begin{tabular}{lc cccc c cccc}
\toprule
& & \multicolumn{4}{c}{\textbf{Edges$\to$Handbags (NFE)}} & & \multicolumn{4}{c}{\textbf{DIODE (NFE)}} \\
\cmidrule(r){3-6} \cmidrule(l){8-11} 
Method & \textbf{Checkpoint} & 10 & 20 & 40 & 100 & & 10 & 20 & 40 & 100 \\
\midrule
DDBM ~\cite{zhou2024ddbm}   & DDBM & 137.15 & 46.74 & 7.79 & 2.40 & & 151.93 & 41.03 & 15.19 & 3.34 \\
DBIM ~\cite{zheng2025diffusion}    & DDBM & 2.49 & 1.74 & 1.26 & 0.91 & & 7.99 & 4.99 & 3.35 & 2.58 \\
ECSI ~\cite{zhang2025exploring}    & DDBM & 2.25 & 1.54 & - & - & & 6.83 & 4.12 & - & - \\
\textbf{DBIM+FMPG (OURS)} & DDBM & \textbf{1.42} & \textbf{1.07} & \textbf{0.89} & \textbf{0.78} & & \textbf{5.28} & \textbf{3.20} & \textbf{2.62} & \textbf{2.06} \\
\bottomrule
\end{tabular}
\end{sc}
\end{scriptsize}
\end{center}
\end{table*}

\begin{figure*}[t]
    \centering
    \begin{minipage}{\linewidth}
        \centering
        \setlength{\tabcolsep}{1pt}
        \begin{subfigure}[b]{0.18\linewidth}
            \centering
            \includegraphics[width=\linewidth, height=\linewidth]{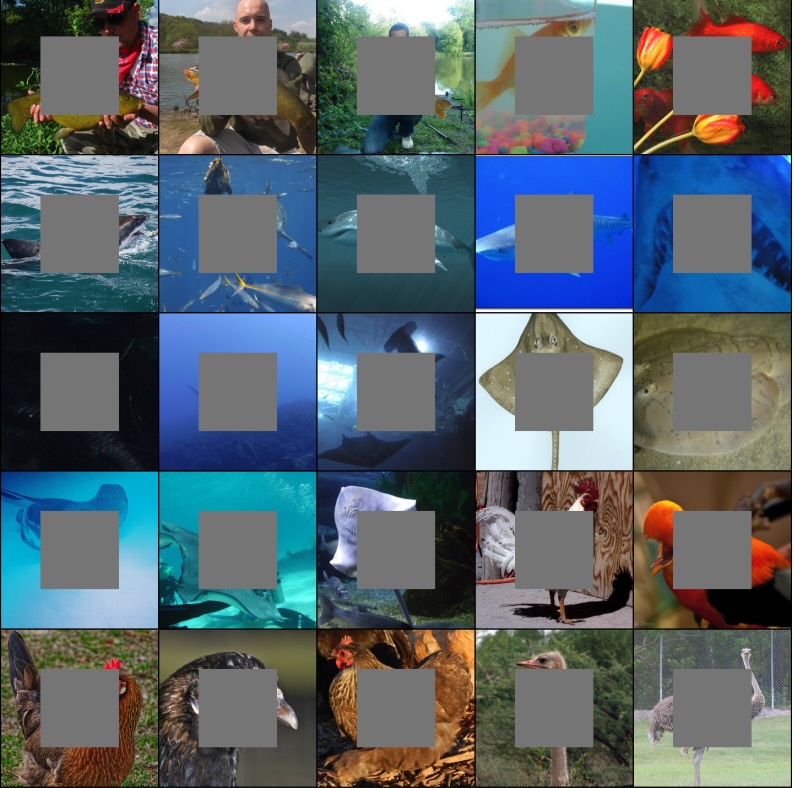}
            \caption{Input $\bm x_T$}
        \end{subfigure}\hfill
        \begin{subfigure}[b]{0.18\linewidth}
            \centering
            \includegraphics[width=\linewidth, height=\linewidth]{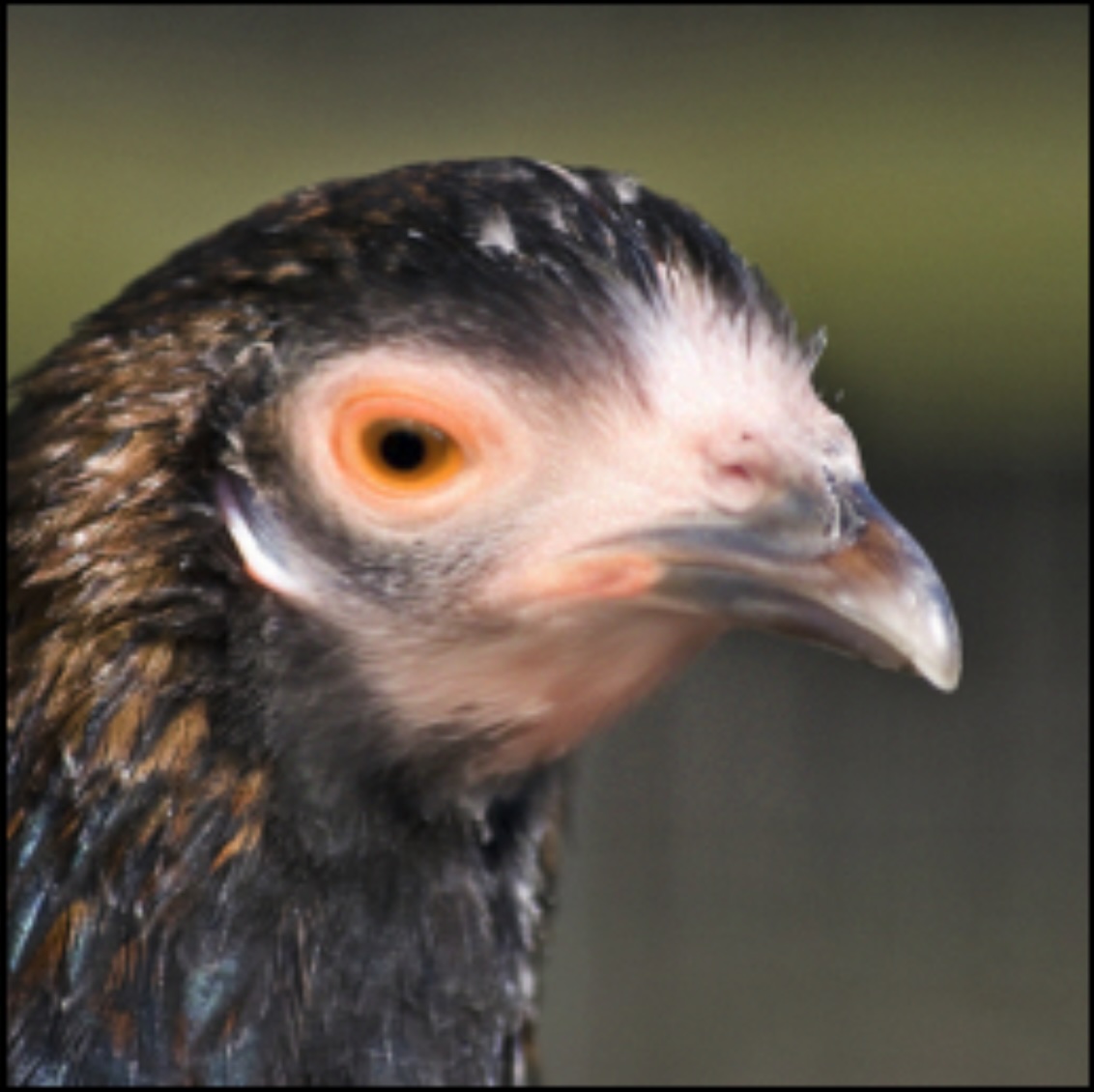}
            \caption{CFG 10}
        \end{subfigure}\hfill
        \begin{subfigure}[b]{0.18\linewidth}
            \centering
            \includegraphics[width=\linewidth, height=\linewidth]{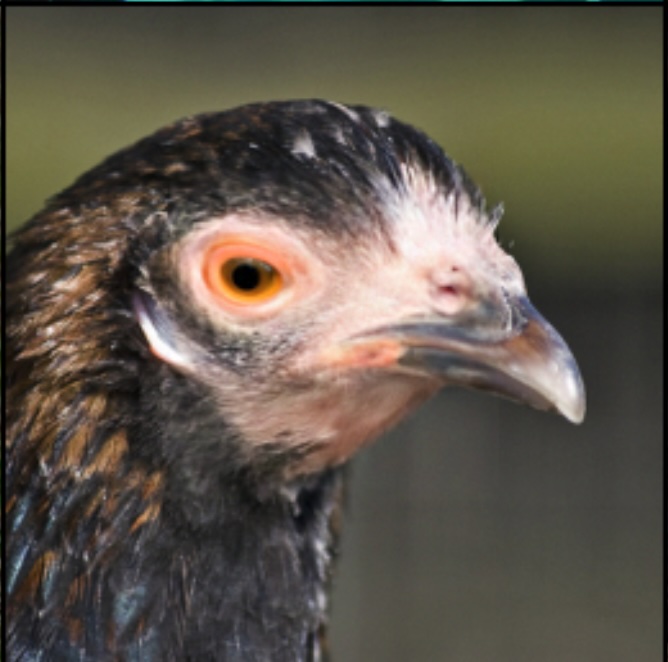}
            \caption{\textbf{Ours }10}
        \end{subfigure}\hfill
        \begin{subfigure}[b]{0.18\linewidth}
            \centering
            \includegraphics[width=\linewidth, height=\linewidth]{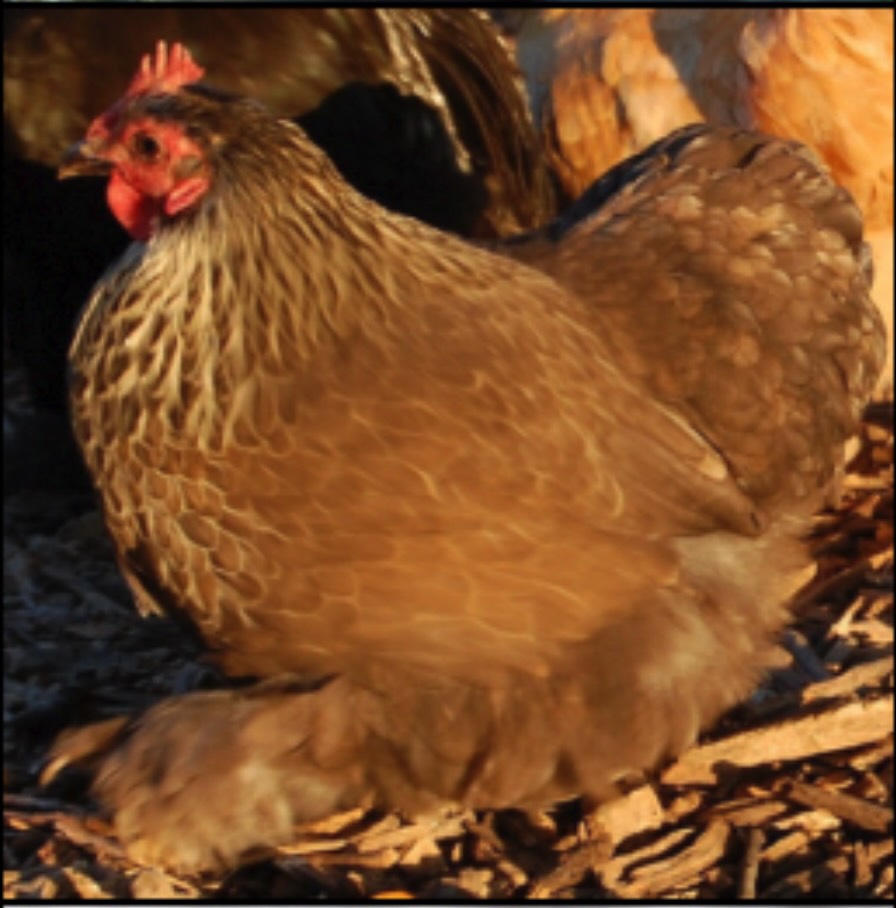}
            \caption{CFG 10}
        \end{subfigure}\hfill
        \begin{subfigure}[b]{0.18\linewidth}
            \centering
            \includegraphics[width=\linewidth, height=\linewidth]{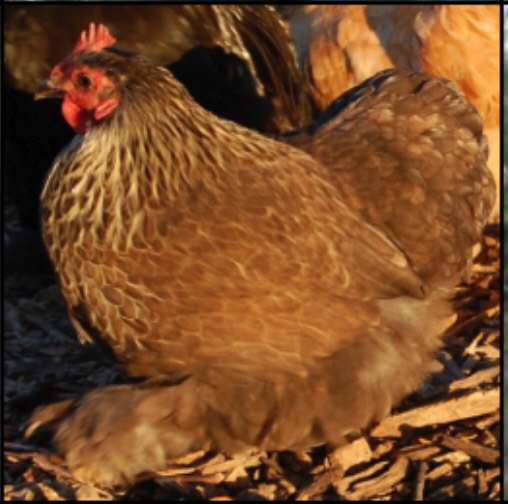}
            \caption{\textbf{Ours }10}
        \end{subfigure}

        \begin{subfigure}[b]{0.18\linewidth}
            \centering
            \includegraphics[width=\linewidth, height=\linewidth]{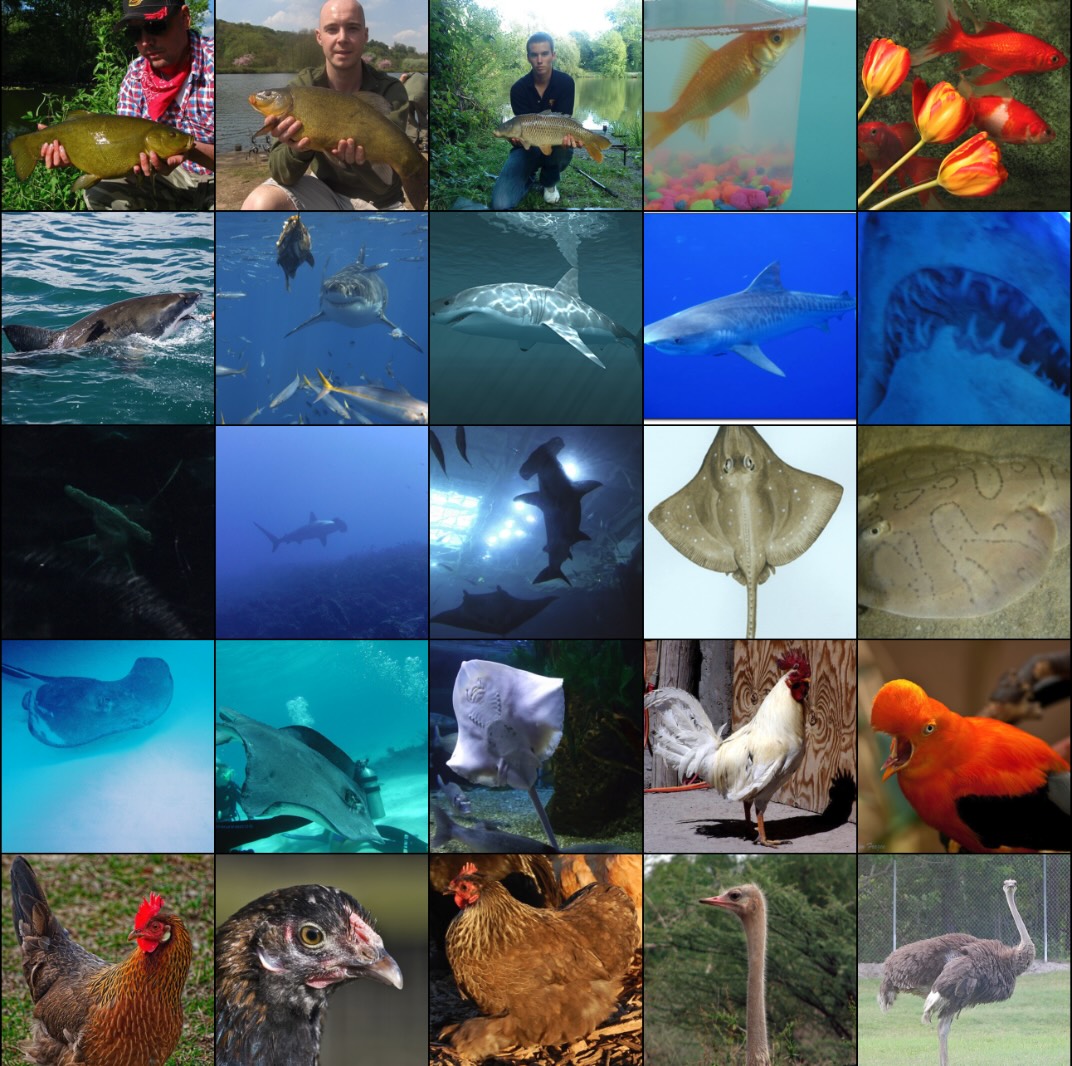}
            \caption{Target $\bm x_0$}
        \end{subfigure}\hfill
        \begin{subfigure}[b]{0.18\linewidth}
            \centering
            \includegraphics[width=\linewidth, height=\linewidth]{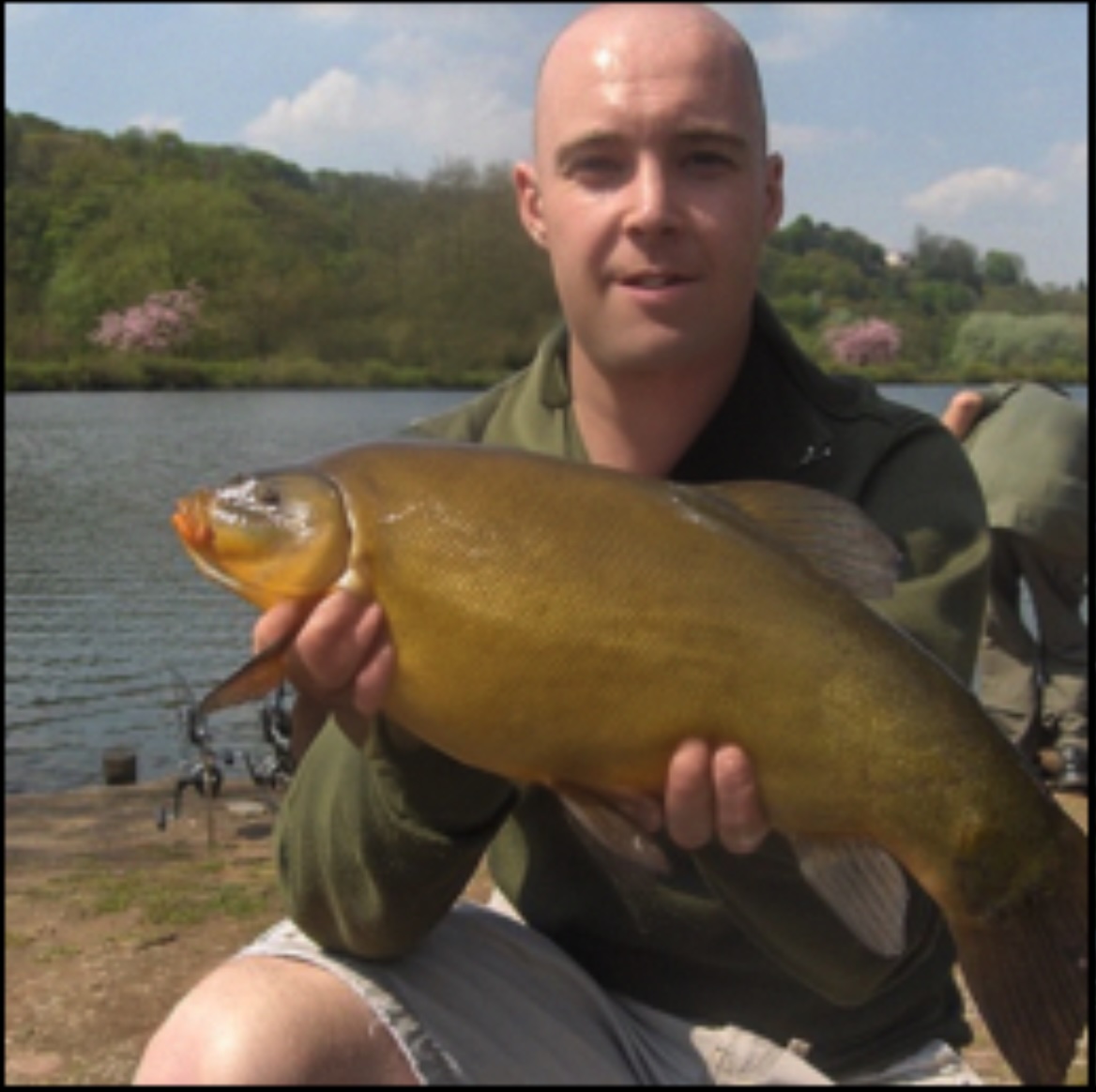}
            \caption{CFG 20}
        \end{subfigure}\hfill
        \begin{subfigure}[b]{0.18\linewidth}
            \centering
            \includegraphics[width=\linewidth, height=\linewidth]{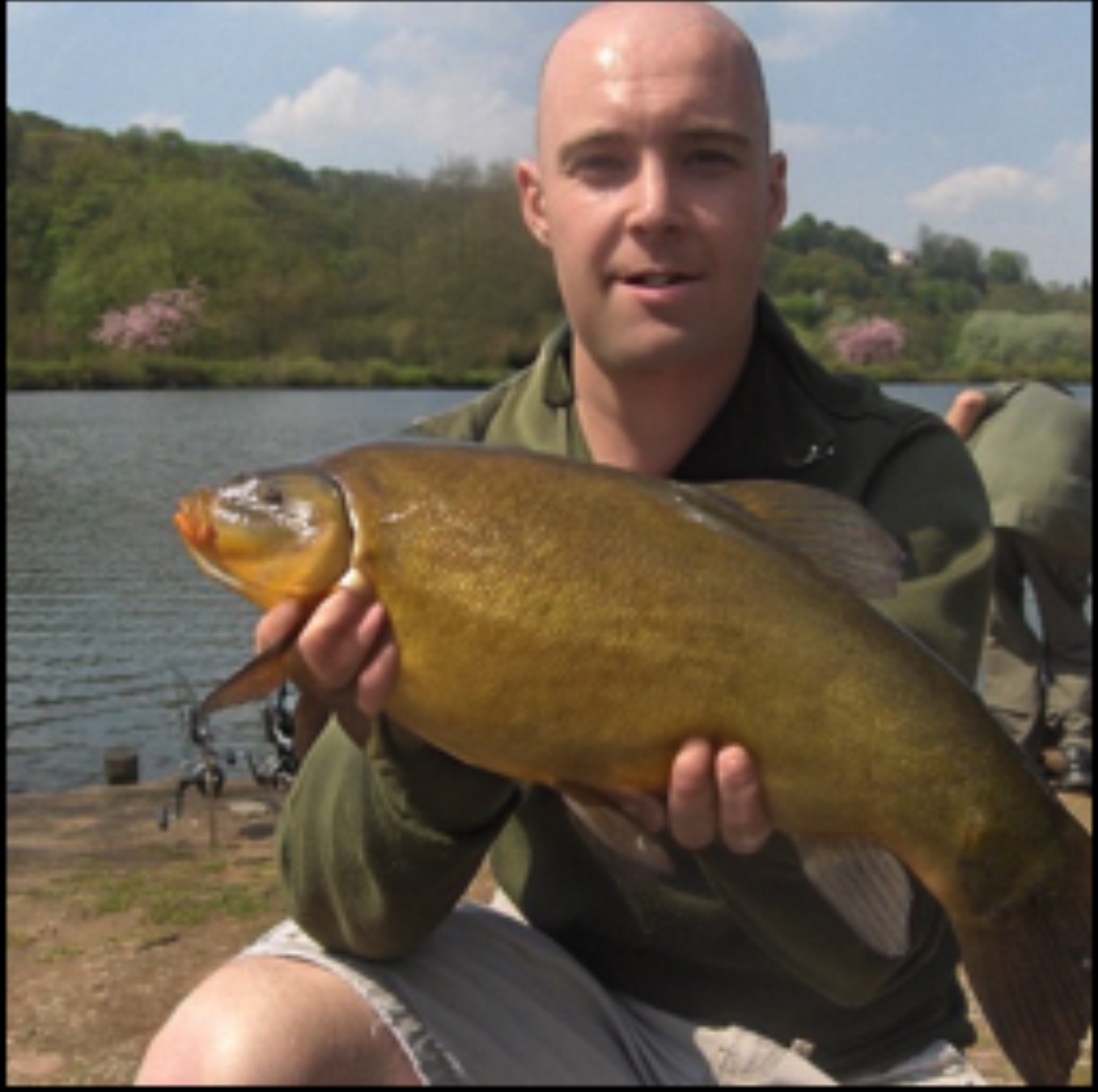}
            \caption{\textbf{Ours }20}
        \end{subfigure}\hfill
        \begin{subfigure}[b]{0.18\linewidth}
            \centering
            \includegraphics[width=\linewidth, height=\linewidth]{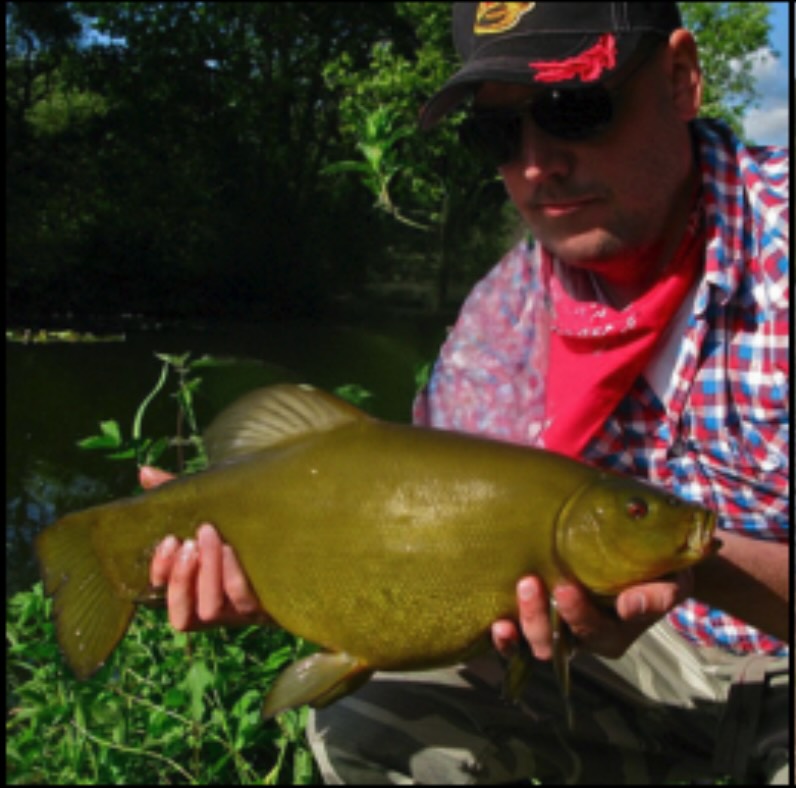}
            \caption{CFG 20}
        \end{subfigure}\hfill
        \begin{subfigure}[b]{0.18\linewidth}
            \centering
            \includegraphics[width=\linewidth, height=\linewidth]{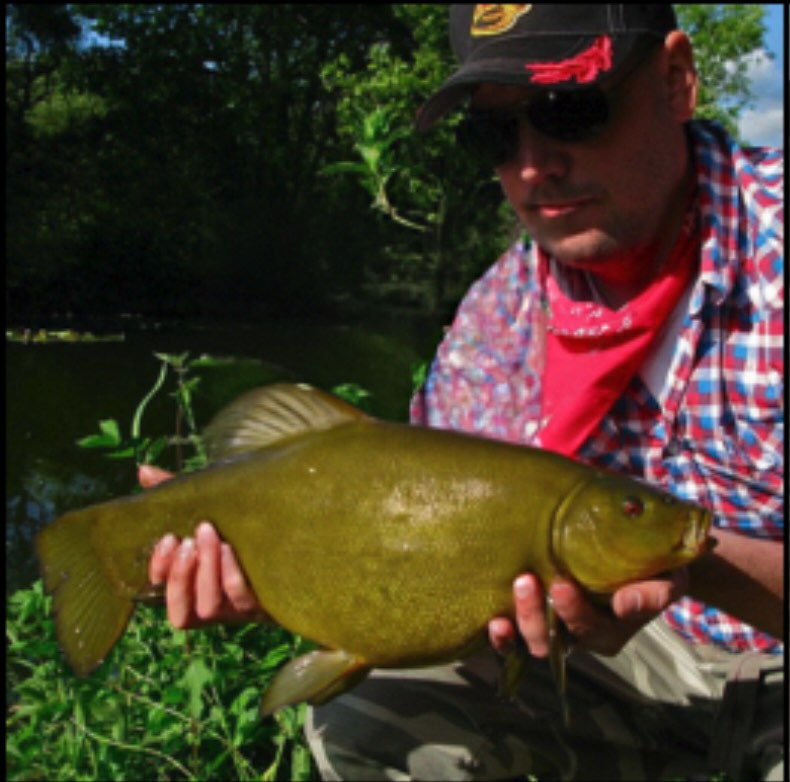}
            \caption{\textbf{Ours }20}
        \end{subfigure}
    \end{minipage}

    \caption{\textbf{Qualitative comparison on ImageNet restoration.} 
    The corruption is simulated using a center \textbf{128$\times$128 mask}. 
    Our proposed hybrid guidance strategy recovers \textbf{semantic layout} and \textbf{high-frequency details}, simultaneously. 
    Notably, our method outperforms the standard CFG baseline.}
    \label{fig:full_comparison_10images}
\end{figure*}

\subsection{Frequency-modulated Prior Guidance}
\label{fmpg}
Building upon our designed PG, we further investigate the underlying mechanism of prior exploitation in the bridge generative process and enhance PG with frequency modulation, named \textbf{FMPG}.
As discussed earlier, a key difference between diffusion and bridge models lies in their exploitation of clean prior representations. 
Consequently, while diffusion models,~\textit{i.e.}, noise-to-data generation, exhibit a monotonically increasing SNR along the sampling trajectory~\cite{NEURIPS2020_4c5bcfec,SGM}, bridge models,~\textit{i.e.}, data-to-data generation, demonstrate a U-shaped SNR profile~\cite{liu2023i2sb,zhou2024ddbm,bridge-tts}, where the SNR decreases in the early sampling stage and gradually increases in late-stage sampling steps.
From the perspective of prior exploitation, this U-shaped SNR trajectory implies that sampling steps close to $\bm x_T$ and $\bm x_0$ operate under relatively high SNR conditions. As a result, both the LF structural information and the HF details are more accessible for effective prior exploitation at these stages. 

Figure~\ref{fig:visual_frequency_evolution} illustrates how the corruption introduced by additional Gaussian noise on $\bm x_t$ propagates to $\bm x_0$. 
We observe that during the intermediate time steps, where the SNR is relatively low, the additional noise on the HF components does not propagate from the prior $\bm x_t$ to the denoising result $\bm x_0$.
This indicates that the model itself cannot effectively exploit the HF information at these steps, since it has already been severely corrupted by the large noise in the bridge process. 
In contrast, at the time steps on both sides with high SNR, the change in $\bm x_0$ resembles the noise injected at $\bm x_t$ and spans the full frequency spectrum, indicating that the model exploits the prior information well across all frequency bands at these steps.

Motivated by this underlying mechanism, we design FMPG, rescaling PG by assigning a time-varying scaling factor, $w^{\text{LF}}_{\text{PG}}$ and $w^{\text{HF}}_{\text{PG}}$, for the LF and HF bands, respectively, as shown in Figure~\ref{fig:guidance_schedules}.
For the LF band, we have:
\begin{equation}
\begin{aligned}
\label{PG}
D^{\text{LF}}_{\bm{\text{FMPG}}}& (\bm x_t, t, \bm x_T) = I^{\text{LF}}\left [ D_{\bm{\theta}}(\mathcal{H}(\bm x_{t}), t, \bm x_T) \right ] + w^{\text{LF}}_{\text{PG}}\\
&(I^{\text{LF}} \left [ D_{\bm{\theta}}(\bm x_{t}, t, \bm x_T) \right ]
- I^{\text{LF}} \left [ D_{\bm{\theta}}(\mathcal{H}(\bm x_{t}), t, \bm x_T) \right ] ),
\end{aligned}
\end{equation}
where $I^{\text{LF}}$ is the LF component extracted with a low-pass filter, and $w^{\text{LF}}_{\text{PG}}$ has an inverted U-shape profile to strengthen the prior exploitation in LF band during intermediate sampling steps.
For the HF band, we have
\begin{equation}
\begin{aligned}
\label{PG}
D^{\text{HF}}_{\bm{\text{FMPG}}}& (\bm x_t, t, \bm x_T) = I^{\text{HF}}\left [ D_{\bm{\theta}}(\mathcal{H}(\bm x_{t}), t, \bm x_T) \right ] + w^{\text{HF}}_{\text{PG}}\\
&(I^{\text{HF}} \left [ D_{\bm{\theta}}(\bm x_{t}, t, \bm x_T) \right ]
- I^{\text{HF}} \left [ D_{\bm{\theta}}(\mathcal{H}(\bm x_{t}), t, \bm x_T) \right ] ),
\end{aligned}
\end{equation}
where $I^{\text{HF}}$ is the remaining HF component, and $w^{\text{HF}}_{\text{PG}}$ has a U-shaped profile similar to the trend of bridge SNR.
Namely, considering that the prior information on the HF band has been corrupted at the time steps with a SNR value, we compress PG in the HF band during the intermediate sampling steps, which is aligned with the mechanism of prior exploitation in the bridge generative process. 

Empirically, we find that a U-shaped design for $w^{\text{HF}}_{\text{PG}}$ and $w^{\text{LF}}_{\text{PG}}$ has already been able to improve the generation result compared to the constant guidance scale, and it does not require a precise alignment with the pre-defined SNR of pre-trained bridge models.

\subsection{Integration with Classifier-free Guidance}
\label{cfg-fmpg}

The designs of PG methods emphasize the advantage of bridge generative process on tasks with strong prior.
However, in tasks such as image in-painting, the provided information may not be a strong prior for the generation target, as shown in Figure~\ref{fig:full_comparison_10images}. 
Therefore, when constructing a weak version of prior, the quality of denoising result may not be distinctively decreased, leading to marginal improvement.
In this scenario, PG methods can naturally be integrated with CFG, which provides complementary advantages.
Specifically, at the early sampling stage $t\in [T, t_s)$, the prior $\bm x_t$ has not been able to provide an instructive prior for the generation target $\bm x_0$. Hence, CFG can be leveraged to guide the in-painting process with global semantic information,~\textit{e.g.}, image class $\bm l$, generating a noisy representation already containing large-scale features of $\bm x_0$ at the time step $t_s$:
\begin{equation}
\begin{aligned}
\label{CFG}
    D_{\bm{\text{CFG}}}(\bm{x}_t, t, \bm l, \bm x_T) &= D_{\bm{\theta}}(\bm{x}_t, t, \bm x_T) + w_{\text{CFG}} \\
    & \left( D_{\bm{\theta}}(\bm{x}_t, t, \bm l, \bm x_T) -  D_{\bm{\theta}}(\bm{x}_t, t, \bm x_T) \right),
\end{aligned}
\end{equation}
where $t\in [T, t_s)$.
In the following sampling steps $t\in [t_s, 0)$, PG methods can stand on the CFG-guided result, emphasizing prior exploitation rather than continually strengthening the semantic alignment with CFG:
\begin{equation}
\begin{aligned}
\label{PG}
D_{\bm{\text{PG}}}(\bm x_t, t, \bm l, \bm x_T) & = D_{\bm{\theta}}(\mathcal{H}(\bm x_{t}), t, \bm l, \bm x_T) + w_{\text{PG}} \\ 
& \left(D_{\bm{\theta}}(\bm x_{t}, t, \bm l, \bm x_T) 
- D_{\bm{\theta}}(\mathcal{H}(\bm x_{t}), t, \bm l, \bm x_T) \right).
\end{aligned}
\end{equation}
By cascading CFG and FMPG methods along the sampling trajectory, the advantages of condition alignment and prior exploitation can be fulfilled concurrently, without compromising the inference speed~\cite{audiomog}.

\section{Experiment}

\subsection{Experimental setting}

\textbf{Datasets and Tasks.}
We evaluate our method on three diverse image-to-image translation and restoration benchmarks.
Edges$\to$Handbags~\cite{Isola_2017_CVPR} is an edge-to-image synthesis task containing 138,567 training images. We evaluate on the entire dataset to ensure statistical significance.
Then, following the experimental protocols of DDBM~\cite{zhou2024ddbm} and DBIM~\cite{zheng2025diffusion}, we utilize the Outdoor subset of the DIODE dataset~\cite{vasiljevic2019diodedenseindooroutdoor} for scene restoration. 
We perform evaluation on the complete outdoor validation split to ensure strict consistency with the baselines.
ImageNet ($256 \times 256$)~\cite{5206848} is a large-scale class-conditional generation task. For the inpainting experiments, we apply a center mask of $128 \times 128$ to evaluate the model's ability to handle high-resolution semantic restoration.

\textbf{Backbones and Baselines.} 
We implement our guidance strategies on top of two state-of-the-art diffusion bridge models.
DDBM~\cite{zhou2024ddbm} is a representative bridge framework that constructs a diffusion bridge process between two arbitrary distributions.
DBIM~\cite{zheng2025diffusion} is an accelerated sampling method for bridge models, analogous to DDIM~\cite{song2021denoising} in standard diffusion.
For DDBM, we employ the recommended hybrid SDE-ODE sampler. 
For DBIM, we adopt the pure ODE mode with $\eta=0.0$.
For fair comparison, we utilize the official pre-trained checkpoints for both backbones without any modifications.
We compare our methods with popular and strong baseline methods including DDIB~\cite{su2023dual}, SDEdit~\cite{meng2022sdedit}, PIX2PIX~\cite{Isola_2017_CVPR}, I2SB~\cite{liu2023i2sb}, and recently proposed ECSI~\cite{zhang2025exploring}.

\textbf{Corruption Methods and Guidance Scale.}
When training-freely employing PG methods in the generation process, we use a two-stage tuning strategy. 
In the first stage, we search an appropriate corruption method, such as noise addition, blurring, JEPG compression, and pooling.
In the second stage, we tune an appropriate guidance scale, which is commonly required by guidance methods, such as CFG or AG.
We provide an investigation of different corruption methods in Appendix.
Empirically, simple noise addition has been a robust method for image translation tasks. 

\textbf{Evaluation Metrics.} We evaluate generation quality using FID~\cite{NIPS2017_8a1d6947} and IS~\cite{NIPS2016_8a3363ab} for distributional distance and diversity, and also reporting LPIPS~\cite{Zhang_2018_CVPR} and MSE to assess reconstruction fidelity. To ensure a fair comparison, we align our evaluation protocol with DBIM~\cite{zheng2025diffusion}. 
We also provide results across a broader range of the number of function evaluations (NFE) to comprehensively evaluate efficiency. In our experiments, we keep the NFE exactly the same as the original baseline, namely using fewer sampling steps in the guided sampling process than the generation process without guidance.

\begin{table}[t!]
    \centering
    \begin{scriptsize}
    \begin{sc}
    
    \caption{The effect of applying PG to DDBM sampler on the Edges$\to$Handbags dataset. We report FID ($\downarrow$) given a large number of inference steps. Noise addition is employed as the degradation method of PG.}
    \label{tab:edges2handbags_fid_ddbm}
    \setlength{\tabcolsep}{8pt}
    \begin{tabular}{lcc}
        \toprule
        & \multicolumn{2}{c}{NFE} \\
        \cmidrule(lr){2-3} 
        Method & 150 & 300 \\
        \midrule
        DDBM ~\cite{zhou2024ddbm} & 1.30 & 0.65   \\  
        \textbf{DDBM+PG} & \textbf{1.23}  & \textbf{0.59}   \\ 
        \bottomrule
    \end{tabular}
    
    \par\rule{0pt}{1.5em}\par
    
    \caption{Ablation study comparing PG and FMPG with DBIM sampler as the baseline on the DIODE dataset. We report FID ($\downarrow$) across NFEs.}
    \label{tab:diode_fid}
    \setlength{\tabcolsep}{5pt}
    \begin{tabular}{lcccc}
        \toprule
        & \multicolumn{4}{c}{NFE} \\
        \cmidrule(lr){2-5} 
        Method  & 10 & 20 & 40 & 100  \\
        \midrule
        DBIM ~\cite{zheng2025diffusion}& 7.99 & 4.99 & 3.35 & 2.58  \\ 
        ECSI ~\cite{zhang2025exploring}& 6.83 & 4.12 & - & - \\
        DBIM+PG (Blur) & 7.33 & 3.89 & 2.64 & \textbf{2.06} \\ 
        DBIM+PG (Noise) & 6.25 & 3.77 & 2.96 & 2.63 \\ 
        \textbf{DBIM+FMPG(noise)} & \textbf{5.28} & \textbf{3.20} & \textbf{2.62} & 2.37 \\ 
        \bottomrule
    \end{tabular}
    
    \end{sc}
    \end{scriptsize}
\end{table}

\begin{table}[t!]
    \centering
    \begin{scriptsize}
    \begin{sc}
    
    \caption{Ablation study comparing PG and FMPG with DBIM sampler as the baseline on the Edges$\to$Handbags (\textbf{10000 images}) dataset. We report FID ($\downarrow$) across NFEs.}
    \label{tab:edges2handbags_fid_10k}
    \setlength{\tabcolsep}{3.5pt} 
    \begin{tabular}{lccccc} 
        \toprule
        & \multicolumn{5}{c}{NFE} \\ 
        \cmidrule(lr){2-6} 
        Method & 6 & 10 & 20 & 40 & 100  \\
        \midrule
        DBIM ~\cite{zheng2025diffusion}         & 4.86 & 4.23 & 3.69 & 3.38 & 3.16 \\ 
        ECSI ~\cite{zhang2025exploring}& \textbf{4.10} & 3.90 & 3.59 & - & -  \\
        DBIM+PG(blur) & 4.72 & 4.01 & 3.50 & 3.26 & 3.15 \\ 
        DBIM+PG(noise) & 4.86 & 3.74 & 3.44 & 3.34 & 3.20  \\ 
        DBIM+FMPG(blur) & 4.27 & 3.75 & 3.36 & 3.20 & 3.14  \\ 
        \textbf{DBIM+FMPG(noise)} & 4.54 & \textbf{3.51} & \textbf{3.23} & \textbf{3.13} & \textbf{3.08}  \\ 
        \bottomrule
    \end{tabular}
    
    \end{sc}
    \end{scriptsize}
\end{table}

\subsection{Main Results}

\textbf{Generation Quality.} 
Given the same NFE, namely with fewer sampling steps as a guidance method, as shown in Table~\ref{tab:combined_quantitative_detailed}, DBIM+FMPG substantially improves DBIM~\cite{zheng2025diffusion} and outperforms the baseline methods, achieving superior image translation quality.
As shown in Table~\ref{tab:combined_quantitative_compact}, DBIM+FMPG achieves higher quality than simply using fast bridge samplers, DBIM and ECSI~\cite{zhang2025exploring}, across different NFEs.

\begin{table*}[t]
\caption{Ablation study on ImageNet dataset comparing different guidance strategies: (1) baseline DBIM sampler without guidance, (2) full CFG guidance, (3) FMPG-first hybrid, and (4) CFG-first hybrid. We report FID ($\downarrow$) across different number of inference steps. Note that the additional NFE for obtaining the low-quality denoising term has been included. Blur is the searched degradation method.}
\label{tab:ablation_imagenet}
\begin{center}
\begin{scriptsize} 
\begin{sc}
\begin{tabular}{lccccccc}
\toprule
& \multicolumn{6}{c}{NFE} \\
\cmidrule(r){2-7}
Method & 8 & 10 & 20 & 40 & 100 & 200  \\
\midrule
DBIM ~\cite{zheng2025diffusion}        & 7.59 & 4.54 & 4.14 & 4.05 & 3.94 & 3.94  \\
DBIM+CFG ~\cite{ho2022classifier, zheng2025diffusion}        & 5.10 & 4.34 & 3.69 & 3.54 & 3.49 & 3.48 \\
DBIM+1/2FMPG+1/2CFG & 5.98 & 5.41 & 3.80 & 3.77 & - & - \\
\textbf{DBIM+1/2CFG+1/2FMPG} & \textbf{4.35} & \textbf{3.86} & \textbf{3.53} & \textbf{3.48} & \textbf{3.47} & \textbf{3.47} \\
\bottomrule
\end{tabular}
\end{sc}
\end{scriptsize}
\end{center}
\end{table*}

Furthermore, we validate the ability of PG to enhance the generation quality on the DDBM sampler, which prioritizes fidelity over speed. 
As shown in Table~\ref{tab:edges2handbags_fid_ddbm}, PG consistently improves the generation quality of DDBM sampler when using a large number of inference steps.

\textbf{Sampling Efficiency.} 
As shown in Table~\ref{tab:diode_fid} and Table~\ref{tab:edges2handbags_fid_10k}, we further evaluate the efficiency of our proposed PG methods, where we systematically compare both PG and FMPG with recently proposed fast samplers for bridge models, namely DBIM~\cite{zheng2025diffusion} and ECSI~\cite{zhang2025exploring}. 
On both the Edges$\to$Handbags and DIODE benchmarks, we can achieve a substantial 2$\times$ acceleration compared to the vanilla DBIM baseline, consistently delivering superior restoration quality at lower NFE regimes. 

Furthermore, the incorporation of frequency modulation into PG, which aligns the mechanism of prior exploitation in bridge models, yields substantial internal improvements. 
By tailoring the guidance scales to specific frequency bands, FMPG is fundamentally more efficient than uniform pixel-level guidance. 
This significant efficiency gain is robust and consistently observed across different datasets, as evidenced in Table~\ref{tab:diode_fid} and Table~\ref{tab:edges2handbags_fid_10k}.

\textbf{Integration with CFG.}
We evaluate our combination with CFG, namely CFG-FMPG, on ImageNet.
For the challenging large-scale ImageNet in-painting task, we introduce the cascaded CFG-FMPG strategy. 
We report FID, IS, and CA in Table~\ref{tab:ablation_imagenet} and Table~\ref{tab:imagenet_is_ca}, respectively.
As shown, this strategy achieves an unprecedented 20$\times$ overall speedup compared to the current state-of-the-art DBIM sampler. 
Notably, CFG-FMPG attains an FID of 3.86 at only 10 NFE, significantly outperforming the baseline which requires significantly higher computational cost to reach comparable performance.

\subsection{Ablation Study}

\textbf{Analysis of Frequency Modulation.}
We further observe that decoupled strategies, such as fixing HF to a constant while applying an inverted U-shape to LF (or vice versa), also yield effective improvements, with the performance gain varying according to the peak magnitude of the curve.
Detailed quantitative results and parameter configurations for the Edges$\to$Handbags and DIODE datasets are provided in Appendix~\ref{app:fmpg_calibration}.
Regarding the curve design, as illustrated in Figure~\ref{fig:guidance_schedules}, the mechanism proves highly robust: it does not necessitate meticulous engineering. A generic convex function that ascends or descends to a peak plateau at a specific timestep is sufficient to achieve significant gains.

\textbf{Integration with CFG.}
To incorporate CFG~\cite{ho2022classifier}, we employ a lightweight fine-tuning strategy for only 10,000 steps (negligible cost).
To ensure fairness, we compare against the optimal CFG hyperparameters identified via an extensive grid search detailed in Appendix~\ref{app:hyperparameter_tuning}.
Furthermore, in Table ~\ref{tab:ablation_imagenet}, a reversed (FMPG-first) schedule performed worse than standalone CFG, validating our analysis regarding the distinct roles of semantic establishment and details refinement.

\begin{table}[t!]
    \centering
    \scriptsize
    \caption{Quantitative results on the ImageNet ($256 \times 256$) task. Methods are grouped by NFE (40, 20, 10). IS ($\uparrow$) and CA ($\uparrow$) denote Inception Score and Classification Accuracy, respectively. Blur is the searched degradation method.}
    \label{tab:imagenet_is_ca}
    \setlength{\tabcolsep}{6pt} 
    \begin{tabular}{lccc}
        \toprule
        \multirow{2}{*}{Method} & \multicolumn{3}{c}{ImageNet ($256 \times 256$)} \\
        \cmidrule(lr){2-4}
        & NFE & IS $\uparrow$ & CA $\uparrow$ \\
        \midrule
        DBIM ~\cite{zheng2025diffusion} & \multirow{3}{*}{10} & 137.6 & 71.6 \\
        DBIM+CFG & & 149.4 & 74.5 \\
        DBIM+\textbf{CFG-FMPG (OURS)} & & \textbf{154.4} & \textbf{75.1} \\
        \midrule
        DBIM ~\cite{zheng2025diffusion} & \multirow{3}{*}{20} & 141.7 & 72.2 \\
        DBIM+CFG & & 158.1 & 76.0 \\
        DBIM+\textbf{CFG-FMPG (OURS)} & & \textbf{160.1} & \textbf{76.1} \\
        \midrule
        DBIM ~\cite{zheng2025diffusion} & \multirow{3}{*}{40} & 143.3 & 72.5 \\
        DBIM+CFG & & 159.1 & \textbf{75.8} \\
        DBIM+\textbf{CFG-FMPG (OURS)} & & \textbf{159.3} & 75.6 \\
        \bottomrule
    \end{tabular}
\end{table}

\section{Related Work}

\paragraph{Guidance methods.}
Existing guidance methods for diffusion models mainly explore two directions. 
One line, exemplified by CFG~\citep{ho2022classifier}, focuses on better exploiting condition information to improve condition alignment. 
Subsequent works such as NAG~\cite{chen2025normalized}, DCFG~\cite{anonymous2026dynamic}, and FDG~\cite{sadat2025guidancefrequencydomainenables} further refine this paradigm by making use of condition information,~\textit{e.g.} text, more stable and effective across different layers, timesteps, or frequency bands. 
Another line, exemplified by AG~\cite{NEURIPS2024_5ee7ed60}, improves the generation quality by constructing degraded predictions to provide a corrective guidance signal. Related methods such as SAG~\cite{10378223}, PAG~\cite{10.1007/978-3-031-73464-9_1}, and SEG~\cite{NEURIPS2024_7b3f7b66} similarly construct negative branches by perturbing the model's internal attention behavior or weakening its effective capacity, and then use the resulting contrast to steer sampling toward higher-quality regions. 
In comparison, our PG methods target a bridge-specific dimension that is absent in standard diffusion models: the strong exploitation of an informative prior. 
A more comprehensive discussion of these related works is provided in Appendix~\ref{app:related_work}.

\paragraph{Bridge models.}
Recent efforts have explored the generative dynamics~\cite{liu2023i2sb,zhou2024ddbm,zhang2025exploring}, parameterization method~\cite{zhou2024ddbm,bridge-tts}, prior representation~\cite{framebridge,RefineBridge}, data space~\cite{audiolbm,voicebridge}, and sampling methods~\cite{zheng2025diffusion,zhang2025exploring} of tractable bridge models.
In this work, GuidedBridge makes the first attempt to improve bridge generation quality with a guidance method that further emphasizes their advantage, namely prior exploitation, which is orthogonal to most previous methods.

\section{Conclusion}

In this work, we propose PG and its extension FMPG, two training-free guidance methods tailored for diffusion bridge models that unlock the untapped potential of the source prior as an informative guidance signal. 
By analyzing the underlying physics of the bridge generative process, FMPG dynamically adapts the guidance scale to match the inherent mechanism of prior exploitation across sampling steps. 
Our framework is orthogonal to existing guidance methods for condition alignment and integrates seamlessly with methods such as CFG, enabling cumulative gains through complementary mechanisms,~\textit{i.e.}, CFG-FMPG. 
Extensive experiments validate that our guidance methods achieve superior generation quality without sacrificing sampling efficiency. 
We hope that this work inspires further exploration into the structural properties of bridge priors and physics-informed guidance mechanisms.

\section*{Acknowledgement}
This work is supported by the National Natural Science Foundation of China (62550004, U24A20342, U25B6003, 92570001).
The authors sincerely thank Yuji Wang at Tsinghua University, China and Chang Li at University of Science and Technology of China for the insightful discussions and suggestions.


\section*{Impact Statement}

This work accelerates Diffusion Bridge Models (DBMs) for image restoration and translation. 

\textbf{Positive Societal Impacts:} 
Our method supports \textit{Green AI} by significantly reducing inference energy and FLOPs, facilitating edge deployment and lowering carbon footprints. In restoration, it aids medical imaging (e.g., low-dose scans) and cultural heritage preservation through high-fidelity recovery.

\textbf{Potential Negative Societal Impacts:} 
Enhanced generation quality and speed inherently increase the risk of misuse, such as creating Deepfakes or generating misleading visual content at scale.

\textbf{Mitigation and Responsibility:} 
We advocate for integrating digital watermarking and detection tools. Furthermore, as models may amplify training biases (e.g., racial or gender), practitioners must perform fairness audits before sensitive deployments.

\bibliographystyle{icml2026}
\bibliography{references}

\appendix
\onecolumn

\setcounter{section}{0}
\renewcommand{\thesection}{\Alph{section}}

\section{Extended Related Work}
\label{app:related_work}

\setlength{\parskip}{0.8em}
\setlength{\parindent}{0pt}

\subsection{Generative Sampling}

\textbf{Guided Sampling.}
Recent perspectives have conceptualized the sampling process of diffusion and bridge models as a search problem within a high-dimensional latent space. Standard diffusion models rely on the learned score function as a gradient field to navigate this space stochastically~\cite{SGM, NEURIPS2020_4c5bcfec}. 
However, in data-to-data translation tasks, the search space is constrained by the source domain. Our proposed Prior Guidance (PG) injects an explicit guidance term derived from the contrast between a clean and a corrupted prior into the optimization landscape. This guidance term effectively prunes the search space by penalizing trajectories that deviate towards the weak prior manifold, thereby accelerating convergence to the target distribution from the prior distribution.

\textbf{Training-Free Guidance of Distribution-Transforming Models.}
Diffusion Bridge Models~\cite{zhou2024ddbm, liu2023i2sb, zheng2025diffusion, CDBM} and Flow Matching represent two prominent classes of distribution-transforming generative models. Currently, research in this domain primarily focuses on guiding Flow Matching via energy functions~\cite{feng2025on}. However, Flow Matching typically relies on a simple Gaussian prior, and bridge models have shown distinctive advantages of data-to-data transformation,~\textit{i.e.}, prior exploitation.

\textbf{Guidance with A Weak Model.}
Existing literature has extensively explored guidance using a weak model~\cite{NEURIPS2024_5ee7ed60, ho2022classifier, 10378223}. However, these studies are predominantly tailored to general diffusion models. As a result, they fail to exploit the unique structural properties of the diffusion bridge Model, leaving its potential for specialized guidance largely unexplored.

\subsection{Guidance Mechanisms in Diffusion Models}

Recent advances in training-free guidance have primarily explored two popular design spaces: mining the external condition information $\bm c$ to maximize alignment, and mining the inherent potential of the pre-trained model $\bm \theta$ to enhance generation quality.

\subsubsection{Condition Mining}
This stream is dedicated to mining the potential of external conditions (e.g., text prompts) to maximize alignment. It aims to suppress unwanted semantic attributes by explicitly manipulating the conditioning signal.

\begin{itemize}
    \item \textbf{Normalized Attention Guidance (NAG)}~\cite{chen2025normalized} addresses the instability of Classifier-Free Guidance (CFG)~\cite{ho2022classifier} in few-step sampling regimes . It observes that standard output-space extrapolation leads to signal saturation and artifacts when sampling steps are aggressive. NAG moves the guidance operation into the attention mechanism with an L1-based normalization stabilizer.
    \item \textbf{Foresight Guidance (FSG)}~\cite{wang2025towards} challenges the linear extrapolation assumption of standard CFG. It reframes the guidance process as a fixed-point iteration problem, aiming to identify a ``Golden Path'' where conditional and unconditional latents achieve consistency through iterative refinement.
\end{itemize}

\subsubsection{Model Potential Mining}
This stream operates without modifying external conditions. Instead, it focuses on mining the potential of the pre-trained model $\bm \theta$ itself by constructing an internal ``adversarial'' state to guide the generation.

\begin{itemize}
    \item \textbf{Self-Attention Guidance (SAG)}~\cite{10378223} pioneers the training-free self-guidance paradigm by leveraging the rich structural information embedded in intermediate self-attention maps. It constructs a negative branch by applying Gaussian blur to the attended content, effectively suppressing high-frequency details and guiding the generation toward enhanced stability and coherence.
        \item \textbf{Perturbed-Attention Guidance (PAG)}~\cite{10.1007/978-3-031-73464-9_1} identifies that the self-attention mechanism is crucial for establishing global structural coherence. By replacing the attention map with an identity matrix during the negative pass, PAG effectively severs the contextual dependencies between tokens.
    \item \textbf{Smoothed Energy Guidance (SEG)}~\cite{NEURIPS2024_7b3f7b66} reinterprets the self-attention mechanism through the lens of Energy-Based Models (EBMs). SEG constructs a negative branch by blurring the Query ($Q$) projection, which effectively ``smoothes'' the energy curvature to mitigate saturation artifacts.
    \item \textbf{Entropy Rectifying Guidance (ERG)}~\cite{berrada2025entropy} takes an information-theoretic approach. It perturbs the attention distribution by increasing the softmax temperature ($\tau$), driving the attention mechanism towards a high-entropy state where the model fails to focus on relevant features.
    \item \textbf{Stochastic Self-Guidance ($S^2$-Guidance)}~\cite{chen2026stochastic} extends the self-guidance paradigm to Diffusion Transformers (DiTs). Unlike U-Net-based methods that mask attention, $S^2$-Guidance constructs a weak model by randomly dropping transformer blocks during the negative pass. This ``perturbed pass'' filters out high-frequency details, allowing the guidance signal to enhance texture quality.
\end{itemize}

\subsubsection{Dynamic and Frequency-Adaptive Guidance}
Beyond the construction of the guidance direction, we also acknowledge a series of works that innovate on the guidance \textit{scale}. These methods, which dynamically modulate the guidance strength, have provided valuable inspiration for our work.

\begin{itemize}
    \item \textbf{Dynamic Classifier-Free Guidance (DCFG)}~\cite{papalampidi2026dynamic} reformulates the guidance scale selection as an online optimization problem. Instead of a fixed schedule, it introduces a quantitative feedback mechanism that evaluates sample quality at each timestep. By maximizing alignment while strictly penalizing deviations (e.g., saturation), DCFG determines the optimal scale adaptively based on the model's instantaneous capacity.
    \item \textbf{Frequency-Decoupled Guidance (FDG)}~\cite{sadat2025guidancefrequencydomainenables} identifies a limitation in uniform scaling: low frequencies (structure) are prone to saturation, while high frequencies (detail) require stronger boosting. FDG decomposes the guidance signal in the spectral domain, applying lower scales to low-frequency components and higher scales to high-frequency components to achieve high-fidelity generation without compromising diversity.
\end{itemize}

\section{Details of Corruption Methods}
\label{app:corruption_details}

In our PG framework, the construction of a weak prior, $\mathcal{H}(\bm x_t)$, is essential for providing the negative gradient signal. 
To systematically study this, we investigate the performance of different corruption operator $\mathcal{H}(\cdot)$. Specifically, given the current state $\bm{x}_t$, the negative prior is training-freely constructed. We instantiate $\mathcal{H}$ with four distinct primitives:

\paragraph{1. Gaussian Noise.}
We define the noise injection operator $\mathcal{H}_{\text{noise}}$ as:
\begin{equation}
    \mathcal{H}_{\text{noise}}(\boldsymbol{x}_t; \sigma) \coloneqq \boldsymbol{x}_t + \boldsymbol{\epsilon}, \quad \text{where } \boldsymbol{\epsilon} \sim \mathcal{N}(\boldsymbol{0}, \sigma^2 \boldsymbol{I}).
\end{equation}
This operator acts as a high-entropy filter that disrupts the signal coherence. By injecting unstructured isotropic Gaussian noise, $\mathcal{H}_{\text{noise}}$ effectively masks high-frequency details without altering the global semantic layout. Empirically, this serves as a "noisy observation" prior, forcing the model to distinguish between the structured signal in $\boldsymbol{x}_t$ and the purely stochastic component in $\mathcal{H}_{\text{noise}}(\bm x_t)$.

\paragraph{2. Gaussian Blur.}
The blurring operator $\mathcal{H}_{\text{blur}}$ is formalized as a convolution with a Gaussian kernel $G_{\sigma}$:
\begin{equation}
    \mathcal{H}_{\text{blur}}(\boldsymbol{x}_t; k) \coloneqq \boldsymbol{x}_t * G_{k, \sigma}.
\end{equation}
Physically, $\mathcal{H}_{\text{blur}}$ functions as a low-pass filter in the frequency domain. It aggressively suppresses high-frequency components, such as sharp edges, textures, and fine-grained details, while preserving the low-frequency structural approximations. Using this as a negative prior explicitly penalizes "over-smoothness," thereby encouraging the guidance mechanism to recover sharpness and high-frequency details.

\paragraph{3. JPEG Artifacts.}
We denote the compression operator as $\mathcal{H}_{\text{jpeg}}$, which involves a lossy encoding-decoding cycle:
\begin{equation}
    \mathcal{H}_{\text{jpeg}}(\boldsymbol{x}_t; Q) \coloneqq \text{Decode}(\text{Encode}(\boldsymbol{x}_t, Q)).
\end{equation}
This operator simulates the quantization errors inherent in JPEG compression with a low Quality Factor ($Q$). Unlike Gaussian noise or blur, $\mathcal{H}_{\text{jpeg}}$ introduces specific structural degradations, including block-wise discontinuities ($8\times8$ blocking artifacts) and Gibbs ringing effects around edges. This guides the model to reject characteristic compression artifacts and improves robustness.

\paragraph{4. Super-Resolution / Pooling (SR4x).}
The resolution degradation operator $\mathcal{H}_{\text{sr}}$ is defined by a downsampling-upsampling coupled process:
\begin{equation}
    \mathcal{H}_{\text{sr}}(\boldsymbol{x}_t; s) \coloneqq \text{Upsample}_s(\text{AvgPool}_s(\boldsymbol{x}_t)).
\end{equation}
With a scaling factor $s=4$, this operator effectively removes all sub-pixel fine-grained information. The naive upsampling step (e.g., nearest-neighbor or bilinear) results in a "pixelated" or aliased image. By using this as a negative reference, the prior guidance is incentivized to reconstructs missing high-frequency details during the generation process.
\begin{table}[h]
    \centering
    \caption{Quantitative Comparison of Corruption Methods (FID $\downarrow$). We report FID scores on \textbf{Edges$\to$Handbags} (\textbf{10,000 sampled images}) across different inference budgets (NFE). Notably, all results in this table are obtained using PG alone, \textbf{without FMPG}, providing a fair comparison of the corruption primitives under a unified setting.}
    \label{tab:corruption_benchmark}
    \vskip 0.15in
    \begin{small}
    \begin{sc}
    \begin{tabular}{lcccc}
        \toprule
        & \multicolumn{4}{c}{\textbf{NFE}} \\
        \cmidrule(r){2-5}
        \textbf{Corruption Method} & \textbf{10} & \textbf{20} & \textbf{40} & \textbf{100} \\
        \midrule
        \textbf{Gaussian Noise} & \textbf{3.75} & \textbf{3.42} & 3.33 & 3.20 \\
        \textbf{Gaussian Blur}  & 4.01 & 3.50 & \textbf{3.26} & \textbf{3.15} \\
        JPEG Artifacts          & 3.64 & 3.80 & 3.48 & 3.35 \\
        SR / Pooling (4x)       & 4.72 & 3.98 & 3.56 & 3.42 \\
        \bottomrule
    \end{tabular}
    \end{sc}
    \end{small}
\end{table}
\begin{table}[H]
    \centering
    \caption{Quantitative Comparison of Corruption Methods on \textbf{DIODE} (FID $\downarrow$). We evaluate the corruption primitives on the DIODE dataset across different inference budgets (NFE). Notably, all results in this table are obtained using PG alone, \textbf{without FMPG}, providing a fair comparison of the corruption primitives under a unified setting.}
    \label{tab:diode_corruption_benchmark}
    \vskip 0.15in
    \begin{small}
    \begin{sc}
    \begin{tabular}{lcccc}
        \toprule
        & \multicolumn{4}{c}{\textbf{NFE}} \\
        \cmidrule(r){2-5}
        \textbf{Corruption Method} & \textbf{10} & \textbf{20} & \textbf{40} & \textbf{100} \\
        \midrule
        \textbf{Gaussian Noise} & \textbf{6.25} & \textbf{3.77} & 2.96 & 2.63 \\
        \textbf{Gaussian Blur}            & 7.33 & 3.89 & \textbf{2.64} & \textbf{2.06} \\ 
        JPEG Artifacts          & 6.58 & 4.25 & 3.42 & 3.15 \\ 
        SR / Pooling (4x)       & 8.45 & 5.60 & 4.15 & 3.72 \\ 
        \bottomrule
    \end{tabular}
    \end{sc}
    \end{small}
\end{table}
\section{Quantitative Analysis on Guidance Parameters}
\label{app:parameter_tables}

In this appendix, we provide a detailed hyperparameter sensitivity analysis on the \textbf{DIODE} dataset across four computational budgets: \textbf{10, 20, 40, and 100 NFE}. Throughout this analysis, we uniformly adopt adding noise as the degradation operator.

\subsection{Low-Budget Regime (10 NFE)}
\label{app:param_10nfe}

Under the low-NFE setting, strong guidance is crucial. The top two performing scales are $w=38.0$ (FID 6.59) and $w=39.0$ (FID 6.64).

\begin{figure}[H]
    \centering
    \begin{minipage}[c]{0.38\linewidth}
        \centering
        \makeatletter\def\@captype{table}\makeatother 
        \caption{\textbf{Guidance Scale} (10 NFE). Best at $w \in \{38.0, 39.0\}$.}
        \label{tab:guidance_sweep_diode}
        
        \begin{small}
        \begin{sc}
        \setlength{\tabcolsep}{3pt} 
        \renewcommand{\arraystretch}{0.9} 
        \begin{tabular}{cc|cc}
            \toprule
            Scale ($w$) & FID & Scale ($w$) & FID \\
            \midrule
            26.0 & 9.18 & 36.0 & 6.73 \\
            28.0 & 8.62 & 37.0 & 6.65 \\
            30.0 & 8.12 & \textbf{38.0} & \textbf{6.59} \\
            32.0 & 7.80 & \textbf{39.0} & \textbf{6.64} \\
            34.0 & 7.50 & 41.0 & 6.93 \\
            \bottomrule
        \end{tabular}
        \end{sc}
        \end{small}
    \end{minipage}
    \hfill
    \begin{minipage}[c]{0.58\linewidth}
        \centering
        \begin{tikzpicture}
            \begin{axis}[
                width=\linewidth,
                height=5.0cm, 
                xlabel={\textbf{Corruption Scale ($\sigma$)}},
                ylabel={\textbf{FID $\downarrow$}},
                grid=major,
                xmin=0.265, xmax=0.345,
                ymin=6.20, ymax=6.75,
                xtick={0.27, 0.29, 0.31, 0.33},
                ylabel near ticks, xlabel near ticks,
                tick label style={font=\footnotesize},
                label style={font=\small},
                title style={font=\small}
            ]
                \addplot[color=red!80!black, mark=*, mark size=1.5pt, thick, smooth, tension=0.55] coordinates {
                    (0.27, 6.59) (0.28, 6.44) (0.29, 6.32)
                    (0.30, 6.28) (0.32, 6.28) (0.34, 6.35)
                };
                \node[anchor=south] at (axis cs: 0.31, 6.28) {\scriptsize \textbf{Optimal Range}};
            \end{axis}
        \end{tikzpicture}
        \caption{\textbf{Corruption Scale (10 NFE).} Evaluated with $w=38.0$. Optimal $\sigma \in [0.30, 0.32]$.}
        \label{fig:corrupt_sweep_diode}
    \end{minipage}
\end{figure}

\subsection{Standard Regime (20 NFE)}
\label{app:param_20nfe}

With 20 NFE, the optimal guidance decreases. The best performance is achieved at $w=20.0$ (FID 3.79) and $w=21.0$ (FID 3.81).

\begin{figure}[H]
    \centering
    \begin{minipage}[c]{0.38\linewidth}
        \centering
        \makeatletter\def\@captype{table}\makeatother 
        \caption{\textbf{Guidance} (20 NFE). Best: $w \in \{20.0, 21.0\}$.}
        \label{tab:scale_20nfe_detail}
        
        \begin{small} \begin{sc} \setlength{\tabcolsep}{3pt} \renewcommand{\arraystretch}{0.9}
        \begin{tabular}{cc|cc}
            \toprule
            Scale ($w$) & FID & Scale ($w$) & FID \\ \midrule
            16.0 & 4.52 & \textbf{21.0} & \textbf{3.81} \\
            18.0 & 3.94 & 22.0 & 3.85 \\
            19.0 & 3.85 & 24.0 & 4.12 \\
            \textbf{20.0} & \textbf{3.79} & 26.0 & 4.68 \\
            \bottomrule
        \end{tabular}
        \end{sc} \end{small}
    \end{minipage}
    \hfill
    \begin{minipage}[c]{0.58\linewidth}
        \centering
        \begin{tikzpicture}
            \begin{axis}[
                width=\linewidth, height=5.0cm, 
                xlabel={\textbf{Guidance Scale ($w$)}}, ylabel={\textbf{FID $\downarrow$}},
                grid=major, xmin=15.5, xmax=26.5, ymin=3.5, ymax=4.8,
                xtick={16, 18, 20, 22, 24, 26},
                ylabel near ticks, xlabel near ticks,
                tick label style={font=\footnotesize}, label style={font=\small}
            ]
                \addplot[color=blue!80!black, mark=*, mark size=1.5pt, thick, smooth, tension=0.6] coordinates {
                    (16.0, 4.52) (17.0, 4.21) (18.0, 3.94) (19.0, 3.85) (20.0, 3.79) 
                    (21.0, 3.81) (22.0, 3.85) (23.0, 3.96) (24.0, 4.12) (25.0, 4.35) (26.0, 4.68)
                };
                \node[anchor=north] at (axis cs: 20.0, 3.79) {\scriptsize \textbf{Optimal (3.79)}};
            \end{axis}
        \end{tikzpicture}
        \captionof{figure}{\textbf{Guidance Scale (20 NFE).} Convex shape confirms optimality at $w=20.0$.}
    \end{minipage}
\end{figure}

\begin{figure}[H]
    \centering
    \begin{minipage}[c]{0.38\linewidth}
        \centering
        \makeatletter\def\@captype{table}\makeatother 
        \caption{\textbf{Corruption} (20 NFE). Best: $\sigma=0.30$.}
        \label{tab:corr_20nfe_detail}
        
        \begin{small} \begin{sc} \setlength{\tabcolsep}{4pt} \renewcommand{\arraystretch}{0.9}
        \begin{tabular}{cc}
            \toprule
            Corr. ($\sigma$) & FID $\downarrow$ \\ \midrule
            0.20 & 4.35 \\
            0.25 & 4.04 \\
            \textbf{0.30} & \textbf{3.77} \\
            0.35 & 4.10 \\
            0.40 & 4.62 \\
            \bottomrule
        \end{tabular}
        \end{sc} \end{small}
    \end{minipage}
    \hfill
    \begin{minipage}[c]{0.58\linewidth}
        \centering
        \begin{tikzpicture}
            \begin{axis}[
                width=\linewidth, height=5.0cm,
                xlabel={\textbf{Corruption Scale ($\sigma$)}}, ylabel={\textbf{FID $\downarrow$}},
                grid=major, xmin=0.18, xmax=0.42, ymin=3.6, ymax=4.8,
                xtick={0.2, 0.25, 0.3, 0.35, 0.4},
                ylabel near ticks, xlabel near ticks,
                tick label style={font=\footnotesize}, label style={font=\small}
            ]
                \addplot[color=red!80!black, mark=square*, mark size=1.5pt, thick, smooth, tension=0.6] coordinates {
                    (0.20, 4.35) (0.25, 4.04) (0.28, 3.79) (0.30, 3.77) 
                    (0.32, 3.85) (0.35, 4.10) (0.40, 4.62)
                };
                \node[anchor=south] at (axis cs: 0.30, 3.77) {\scriptsize \textbf{Best (3.77)}};
            \end{axis}
        \end{tikzpicture}
        \captionof{figure}{\textbf{Corruption Scale (20 NFE).} Robust around $\sigma=0.30$.}
    \end{minipage}
\end{figure}

\subsection{High-Fidelity Regime (40 NFE)}
\label{app:param_40nfe}

Optimal guidance scale shifts to $w \approx 14.0$. The top two scales are $w=14.0$ (FID 2.96) and $w=13.0$ (FID 3.05).

\begin{figure}[H]
    \centering
    \begin{minipage}[c]{0.38\linewidth}
        \centering
        \makeatletter\def\@captype{table}\makeatother 
        \caption{\textbf{Guidance} (40 NFE). Best: $w \in \{13.0, 14.0\}$.}
        \label{tab:scale_40nfe_detail}
        
        \begin{small} \begin{sc} \setlength{\tabcolsep}{3pt} \renewcommand{\arraystretch}{0.9}
        \begin{tabular}{cc|cc}
            \toprule
            Scale ($w$) & FID & Scale ($w$) & FID \\ \midrule
            10.0 & 3.85 & 15.0 & 3.08 \\
            12.0 & 3.18 & 16.0 & 3.26 \\
            \textbf{13.0} & \textbf{3.05} & 18.0 & 3.79 \\
            \textbf{14.0} & \textbf{2.96} & 20.0 & 4.67 \\
            \bottomrule
        \end{tabular}
        \end{sc} \end{small}
    \end{minipage}
    \hfill
    \begin{minipage}[c]{0.58\linewidth}
        \centering
        \begin{tikzpicture}
            \begin{axis}[
                width=\linewidth, height=5.0cm,
                xlabel={\textbf{Guidance Scale ($w$)}}, ylabel={\textbf{FID $\downarrow$}},
                grid=major, xmin=9.5, xmax=20.5, ymin=2.8, ymax=4.8,
                xtick={10, 12, 14, 16, 18, 20},
                ylabel near ticks, xlabel near ticks,
                tick label style={font=\footnotesize}, label style={font=\small}
            ]
                \addplot[color=blue!80!black, mark=*, mark size=1.5pt, thick, smooth, tension=0.6] coordinates {
                    (10.0, 3.85) (12.0, 3.18) (13.0, 3.05) (14.0, 2.96) 
                    (15.0, 3.08) (16.0, 3.26) (17.0, 3.52) (18.0, 3.79) (20.0, 4.67)
                };
                \node[anchor=south] at (axis cs: 14.0, 2.96) {\scriptsize \textbf{Optimal (2.96)}};
            \end{axis}
        \end{tikzpicture}
        \captionof{figure}{\textbf{Guidance Scale (40 NFE).} Shifted lower to $w=14.0$.}
    \end{minipage}
\end{figure}

\begin{figure}[H]
    \centering
    \begin{minipage}[c]{0.38\linewidth}
        \centering
        \makeatletter\def\@captype{table}\makeatother 
        \caption{\textbf{Corruption} (40 NFE). Best: $\sigma=0.25$.}
        \label{tab:corr_40nfe_detail}
        
        \begin{small} \begin{sc} \setlength{\tabcolsep}{4pt} \renewcommand{\arraystretch}{0.9}
        \begin{tabular}{cc}
            \toprule
            Corr. ($\sigma$) & FID $\downarrow$ \\ \midrule
            0.15 & 3.80 \\
            0.20 & 3.25 \\
            \textbf{0.25} & \textbf{2.96} \\
            0.30 & 3.51 \\
            0.35 & 4.20 \\
            \bottomrule
        \end{tabular}
        \end{sc} \end{small}
    \end{minipage}
    \hfill
    \begin{minipage}[c]{0.58\linewidth}
        \centering
        \begin{tikzpicture}
            \begin{axis}[
                width=\linewidth, height=5.0cm,
                xlabel={\textbf{Corruption Scale ($\sigma$)}}, ylabel={\textbf{FID $\downarrow$}},
                grid=major, xmin=0.13, xmax=0.37, ymin=2.8, ymax=4.4,
                xtick={0.15, 0.20, 0.25, 0.30, 0.35},
                ylabel near ticks, xlabel near ticks,
                tick label style={font=\footnotesize}, label style={font=\small}
            ]
                \addplot[color=red!80!black, mark=square*, mark size=1.5pt, thick, smooth, tension=0.6] coordinates {
                    (0.15, 3.80) (0.20, 3.25) (0.23, 3.02) (0.25, 2.96) 
                    (0.27, 3.10) (0.30, 3.51) (0.35, 4.20)
                };
                \node[anchor=south] at (axis cs: 0.25, 2.96) {\scriptsize \textbf{Best (2.96)}};
            \end{axis}
        \end{tikzpicture}
        \captionof{figure}{\textbf{Corruption Scale (40 NFE).} Best at clean prior $\sigma=0.25$.}
    \end{minipage}
\end{figure}

\subsection{Converged Regime (100 NFE)}
\label{app:param_100nfe}

At 100 NFE, best FID $\approx 2.67$. The optimal guidance scales are $w=9.0$ (FID 2.67) and $w=10.0$ (FID 2.77).

\begin{figure}[H]
    \centering
    \begin{minipage}[c]{0.38\linewidth}
        \centering
        \makeatletter\def\@captype{table}\makeatother 
        \caption{\textbf{Guidance} (100 NFE). Best: $w \in \{9.0, 10.0\}$.}
        \label{tab:scale_100nfe_detail}
        
        \begin{small} \begin{sc} \setlength{\tabcolsep}{4pt} \renewcommand{\arraystretch}{0.9}
        \begin{tabular}{cc}
            \toprule
            Scale ($w$) & FID $\downarrow$ \\ \midrule
            6.0 & 3.65 \\
            8.0 & 2.96 \\
            \textbf{9.0} & \textbf{2.67} \\
            \textbf{10.0} & \textbf{2.77} \\
            12.0 & 3.42 \\
            \bottomrule
        \end{tabular}
        \end{sc} \end{small}
    \end{minipage}
    \hfill
    \begin{minipage}[c]{0.58\linewidth}
        \centering
        \begin{tikzpicture}
            \begin{axis}[
                width=\linewidth, height=5.0cm,
                xlabel={\textbf{Guidance Scale ($w$)}}, ylabel={\textbf{FID $\downarrow$}},
                grid=major, xmin=5.5, xmax=12.5, ymin=2.5, ymax=3.8,
                xtick={6, 8, 9, 10, 12},
                ylabel near ticks, xlabel near ticks,
                tick label style={font=\footnotesize}, label style={font=\small}
            ]
                \addplot[color=blue!80!black, mark=*, mark size=1.5pt, thick, smooth, tension=0.6] coordinates {
                    (6.0, 3.65) (7.0, 3.21) (8.0, 2.96) (9.0, 2.67) 
                    (10.0, 2.77) (11.0, 3.05) (12.0, 3.42)
                };
                \node[anchor=south] at (axis cs: 9.0, 2.67) {\scriptsize \textbf{Optimal (2.67)}};
            \end{axis}
        \end{tikzpicture}
        \captionof{figure}{\textbf{Guidance Scale (100 NFE).} Converged at $w=9.0$.}
    \end{minipage}
\end{figure}

\begin{figure}[H]
    \centering
    \begin{minipage}[c]{0.38\linewidth}
        \centering
        \makeatletter\def\@captype{table}\makeatother 
        \caption{\textbf{Corruption} (100 NFE). Best: $\sigma=0.22$.}
        \label{tab:corr_100nfe_detail}
        
        \begin{small} \begin{sc} \setlength{\tabcolsep}{4pt} \renewcommand{\arraystretch}{0.9}
        \begin{tabular}{cc}
            \toprule
            Corr. ($\sigma$) & FID $\downarrow$ \\ \midrule
            0.15 & 3.12 \\
            0.20 & 2.77 \\
            \textbf{0.22} & \textbf{2.67} \\
            0.26 & 2.95 \\
            0.30 & 3.45 \\
            \bottomrule
        \end{tabular}
        \end{sc} \end{small}
    \end{minipage}
    \hfill
    \begin{minipage}[c]{0.58\linewidth}
        \centering
        \begin{tikzpicture}
            \begin{axis}[
                width=\linewidth, height=5.0cm,
                xlabel={\textbf{Corruption Scale ($\sigma$)}}, ylabel={\textbf{FID $\downarrow$}},
                grid=major, xmin=0.14, xmax=0.31, ymin=2.5, ymax=3.6,
                xtick={0.15, 0.20, 0.22, 0.26, 0.30},
                ylabel near ticks, xlabel near ticks,
                tick label style={font=\footnotesize}, label style={font=\small}
            ]
                \addplot[color=red!80!black, mark=square*, mark size=1.5pt, thick, smooth, tension=0.6] coordinates {
                    (0.15, 3.12) (0.18, 2.85) (0.20, 2.77) (0.22, 2.67) 
                    (0.24, 2.75) (0.26, 2.95) (0.30, 3.45)
                };
                \node[anchor=south] at (axis cs: 0.22, 2.67) {\scriptsize \textbf{Best (2.67)}};
            \end{axis}
        \end{tikzpicture}
        \captionof{figure}{\textbf{Corruption Scale (100 NFE).} Very clean prior required ($\sigma=0.22$).}
    \end{minipage}
\end{figure}

\section{Frequency-Specific Guidance Analysis}
\label{app:freq_analysis}

To validate our Frequency-Modulated Prior Guidance (FMPG), we conducted an ablation study by restricting the guidance signal to specific frequency bands using FFT decomposition.

\begin{algorithm}[H]
   \caption{Frequency-modulated Prior Guidance for Bridge Models}
   \label{alg:fdg_sampling}
\begin{algorithmic}[1]
    \small 
    \STATE {\bfseries Input:} Pre-trained denoiser $D_{\boldsymbol{\theta}}$, source image $\boldsymbol{x}_T$, steps $\{t_N, \dots, t_0\}$, corruption $\mathcal{H}$, scales $w_{\text{low}}, w_{\text{high}}$.
    \STATE {\bfseries Output:} Sampled image $\boldsymbol{x}_0$.
      
    \STATE $\boldsymbol{x} \leftarrow \boldsymbol{x}_T$
    \FOR{$i = N$ {\bfseries to} $1$}
       
        \STATE $\boldsymbol{x}^{\text{bad}} \leftarrow \mathcal{H}(\boldsymbol{x})$ \quad $\rhd$ Construct weak prior (bad state)
       
        \STATE $\hat{\boldsymbol{x}}_0^{\text{good}}, \hat{\boldsymbol{x}}_0^{\text{bad}} \leftarrow D_{\boldsymbol{\theta}}(\boldsymbol{x}, t_i), D_{\boldsymbol{\theta}}(\boldsymbol{x}^{\text{bad}}, t_i)$ \quad $\rhd$ Predict $\boldsymbol{x}_0$ from both states
       
        \STATE $\boldsymbol{\Delta} \leftarrow \hat{\boldsymbol{x}}_0^{\text{good}} - \hat{\boldsymbol{x}}_0^{\text{bad}}$ \quad $\rhd$ Calculate guidance residual
       
        \STATE $\mathcal{F}_{\boldsymbol{\Delta}} \leftarrow \text{FFT}(\boldsymbol{\Delta})$ \quad $\rhd$ Transform to frequency domain
       
        \STATE $\boldsymbol{\Delta}_{\text{low}} \leftarrow \text{iFFT}(\text{LowPass}(\mathcal{F}_{\boldsymbol{\Delta}}))$ \quad $\rhd$ Extract low-frequency component
       
        \STATE $\boldsymbol{\Delta}_{\text{high}} \leftarrow \boldsymbol{\Delta} - \boldsymbol{\Delta}_{\text{low}}$ \quad $\rhd$ Extract high-frequency component
       
        \STATE $\boldsymbol{\Delta}_{\text{guided}} \leftarrow w_{\text{low}} \cdot \boldsymbol{\Delta}_{\text{low}} + w_{\text{high}} \cdot \boldsymbol{\Delta}_{\text{high}}$ \quad $\rhd$ Apply frequency-specific scales
       
        \STATE $\hat{\boldsymbol{x}}_0^{\text{target}} \leftarrow \hat{\boldsymbol{x}}_0^{\text{bad}} + \boldsymbol{\Delta}_{\text{guided}}$ \quad $\rhd$ Rectify prediction with guidance
       
        \STATE $\boldsymbol{x} \leftarrow \text{DBIMStep}(\boldsymbol{x}, \hat{\boldsymbol{x}}_0^{\text{target}}, \boldsymbol{x}_T, t_i, t_{i-1})$ \quad $\rhd$ Update state with 5 params
    \ENDFOR
    \STATE \textbf{return} $\boldsymbol{x}$
\end{algorithmic}
\end{algorithm}
\subsection{Static vs. Dynamic Frequency Modulation}

To isolate the contribution of our dynamic scheduling strategy, we compare our \textbf{FMPG} (Frequency-Modulated Prior Guidance) against a static baseline, denoted as \textbf{Static-PG}.
In the Static-PG setting, the guidance scale is kept constant and identical for both high and low-frequency components throughout the sampling trajectory (i.e., $w_{\text{hf}}(t) = w_{\text{lf}}(t) = \lambda_{\text{const}}$). This baseline assumes that prior information is equally reliable across all frequencies and time steps.

However, our analysis reveals that this assumption is suboptimal for Bridge processes. As illustrated in the main text, the Signal-to-Noise Ratio (SNR) of a Bridge process follows a U-shaped curve~\cite{zhou2024ddbm, zheng2025diffusion}, implying that the intermediate steps ($t \approx 0.5$) are dominated by noise, while the trajectory boundaries ($t \to 0$ and $t \to 1$) contain cleaner signal.

\textbf{1. The Failure of Static Guidance:}
Static-PG forces a compromise. If the scale is set high to capture details, it amplifies noise in the intermediate steps, leading to high-frequency artifacts. If set low to ensure stability, it fails to sufficiently correct the structural alignment or refine fine textures at the endpoints.

\textbf{2. The Advantage of FMPG (Ours):}
Our FMPG decouples the frequency bands to align with the physics of the Bridge process:
\begin{itemize}
    \item \textbf{High-Frequency (HF) as a U-Shape:} We adopt a U-shaped schedule for HF components. Guidance is strengthened at the endpoints ($t=0 $ and $t=1$) to perform fine-grained texture refinement on the high-SNR data, but relaxed in the noisy intermediate region to prevent artifact amplification.
    \item \textbf{Low-Frequency (LF) as an Inverted U-Shape:} Conversely, LF components follow an inverted U-shaped schedule. We amplify LF guidance in the high-uncertainty intermediate region. This acts as a strong ``structural anchor,'' ensuring the global layout remains consistent when the signal is weakest, before tapering off at the boundaries where the data is already structurally sound.
\end{itemize}

In this section, we investigate the sensitivity of the baseline \textbf{Static-PG} method to the guidance scale parameter $w$. We performed a fine-grained grid search on the \textit{Edges$\to$Handbags} dataset~\cite{Isola_2017_CVPR} to identify the optimal static scale.

\begin{figure}[h]
    \centering
    \begin{tikzpicture}
        \begin{axis}[
            width=0.7\linewidth,
            height=6cm,
            xlabel={\textbf{Guidance Scale} ($w$)},
            ylabel={\textbf{FID} $\downarrow$},
            xmin=14.5, xmax=21.5,
            ymin=3.4, ymax=4.4, 
            grid=major,
            grid style={dashed, gray!30},
            title={\textbf{Static-PG Performance vs. Guidance Scale}},
            title style={font=\small, yshift=-1ex},
            label style={font=\small},
            tick label style={font=\footnotesize},
            legend pos=north west
        ]
            \addplot[
                color=blue!80!black,
                mark=*,
                mark size=2pt,
                line width=1.5pt,
            ] coordinates {
                (15.0, 4.21)
                (16.0, 3.75)
                (16.5, 3.58)
                (17.0, 3.53)
                (17.5, 3.47)
                (18.0, 3.44)
                (18.5, 3.49)
                (19.0, 3.56)
                (20.0, 3.85)
                (21.0, 4.30)
            };
            \addlegendentry{FID Score}
            \addplot[
                only marks,
                mark=*,
                mark size=3pt,
                color=red
            ] coordinates {(18, 3.44)};
            
            \addplot[
                only marks,
                mark=x,
                mark size=3pt,
                color=black!70
            ] coordinates {(16, 3.75) (20, 3.85)};
            \node[anchor=north, align=center, color=red!80!black, font=\footnotesize] 
                at (axis cs: 18, 3.42) 
                {\textbf{Optimal}\\($w=18$)};
            \draw[dashed, red, thin] (axis cs: 18, 3.4) -- (axis cs: 18, 3.44);
        \end{axis}
    \end{tikzpicture}
    \caption{\textbf{Guidance Scale Search for Static-PG (Edges2Handbags}~\cite{Isola_2017_CVPR}). Ten discrete scale points are connected by line segments, forming a roughly U-shaped unimodal curve with the optimum at $w=18$.}
    \label{fig:static_scale_search}
\end{figure}
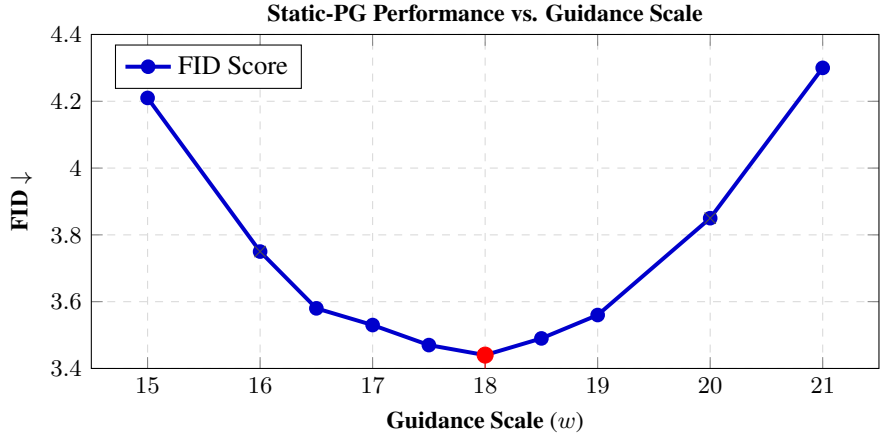
\subsection{Calibration of Dynamic FMPG Schedules}
\label{app:fmpg_calibration}

Building upon the optimal static baseline ($w_{\text{base}}=18$), we further unlock the potential of our method by calibrating the dynamic modulation intensity. We independently fine-tune the modulation range for Low-Frequency (LF) and High-Frequency (HF) components to find the optimal deviation from the base scale.

\textbf{1. Low-Frequency (Structural) Calibration:}
For the LF component, we employ an \textbf{Inverted U-shaped} schedule to strengthen structural guidance in the intermediate diffusion steps. We fix the endpoints at $w=18$ and search for the optimal \textbf{Peak Scale} ($w_{\text{lf}}^{\text{max}}$) in the range of $[18.5, 20.5]$.

\textbf{2. High-Frequency (Texture) Calibration:}
For the HF component, we employ a \textbf{U-shaped} schedule to relax guidance in the noisy intermediate steps. We fix the endpoints at $w=18$ and search for the optimal \textbf{Minimum Scale} ($w_{\text{hf}}^{\text{min}}$) in the range of $[15.5, 17.5]$.

The ablation results are recorded in Table~\ref{tab:lf_calibration} and Table~\ref{tab:hf_calibration} respectively.

\begin{table}[h]
    \centering
    \caption{\textbf{Ablation study on Dynamic Modulation Intensity.} The base scale is fixed at $w_{\text{base}}=18$. We report FID scores~\cite{NIPS2017_8a1d6947} ($\downarrow$) for different modulation amplitudes.}
    \label{tab:fmpg_calibration}
    
    \begin{minipage}[t]{0.48\linewidth}
        \centering
        \subcaption{\textbf{Low-Frequency (LF) Tuning}. Increasing the peak of the Inverted U-shape.}
        \label{tab:lf_calibration}
        \begin{small}
        \begin{sc}
        \setlength{\tabcolsep}{4pt}
        \begin{tabular}{ccc}
            \toprule
            \textbf{Base Scale} & \textbf{LF Peak} ($w_{\text{lf}}^{\text{max}}$) & \textbf{FID} \\
            \midrule
            18.0 & 18.5 & 3.42 \\
            18.0 & 19.0 & 3.37 \\
            18.0 & 19.5 & 3.35 \\
            18.0 & 20.0 & 3.30 \\
            18.0 & 20.5 & 3.28 \\
            18.0 & 21.0 & 3.31 \\
            \bottomrule
        \end{tabular}
        \end{sc}
        \end{small}
    \end{minipage}
    \hfill
    \begin{minipage}[t]{0.48\linewidth}
        \centering
        \subcaption{\textbf{High-Frequency (HF) Tuning}. Decreasing the trough of the U-shape.}
        \label{tab:hf_calibration}
        \begin{small}
        \begin{sc}
        \setlength{\tabcolsep}{4pt}
        \begin{tabular}{ccc}
            \toprule
            \textbf{Base Scale} & \textbf{HF Min} ($w_{\text{hf}}^{\text{min}}$) & \textbf{FID} \\
            \midrule
            18.0 & 17.5 & 3.43 \\
            18.0 & 17.0 & 3.37 \\
            18.0 & 16.5 & 3.35 \\
            18.0 & 16.0 & 3.30 \\
            18.0 & 15.5 & 3.26 \\
            18.0 & 15.0 & 3.28 \\
            \bottomrule
        \end{tabular}
        \end{sc}
        \end{small}
    \end{minipage}
\end{table}

\noindent \textbf{Synergy of Optimal Schedules.} Finally, by simultaneously applying these optimized modulation strategies, specifically the \textbf{Inverted U-shape} for LF (peaking at $w_{\text{lf}}^{\text{max}}=20.5$) and the \textbf{U-shape} for HF (dipping to $w_{\text{hf}}^{\text{min}}=15.5$) as illustrated in the main text, our method achieves its peak performance reported in the repository: an FID of \textbf{3.19} (evaluated on 10,000 images) and \textbf{1.07} (evaluated on the full dataset of 138,567 images).

\section{Hyperparameter Tuning and Sensitivity Analysis}
\label{app:hyperparameter_tuning}

In this section, we provide detailed experimental evidence regarding the optimization of our baselines and the sensitivity of our proposed hyperparameters.

\subsection{Optimization of DBIM+CFG Baseline on ImageNet}
\label{app:baseline_tuning_sub}

We first disclose the tuning process for the strong baseline: \textbf{DBIM}~\cite{zheng2025diffusion} combined with Classifier-Free Guidance (DBIM+CFG)~\cite{ho2022classifier} on ImageNet~\cite{5206848} ($256\times256$). To ensure a rigorous comparison, we exhaustively searched for the optimal scale $w$ at each computational budget (10, 20, and 40 NFE).

\begin{figure}[h!] 
    \centering
    \begin{minipage}[c]{0.38\linewidth}
        \centering
        \makeatletter\def\@captype{table}\makeatother 
        \caption{\textbf{10 NFE (ImageNet).} Best at $w=2.0$.}
        \label{tab:scale_10nfe}
        
        \begin{small}
        \begin{sc}
        \begin{tabular}{cc}
            \toprule
            Scale ($w$) & FID $\downarrow$ \\
            \midrule
            1.5 & 4.93 \\
            1.7 & 4.66 \\
            \textbf{2.0} & \textbf{4.33} \\
            2.3 & 4.37 \\
            2.6 & 4.42 \\
            3.0 & 8.24 \\
            \bottomrule
        \end{tabular}
        \end{sc}
        \end{small}
    \end{minipage}
    \hfill
    \begin{minipage}[c]{0.58\linewidth}
        \centering
        \begin{tikzpicture}
            \begin{axis}[
                width=\linewidth, height=5.0cm,
                xlabel={\textbf{Guidance Scale ($w$)}},
                ylabel={\textbf{FID $\downarrow$}},
                grid=major,
                ymin=3.6, ymax=8.5,
                xtick={1.5, 2.0, 2.5, 3.0},
                ylabel near ticks, xlabel near ticks,
                tick label style={font=\scriptsize},
                label style={font=\footnotesize},
                title style={font=\small}
            ]
            \addplot[color=red, mark=square*, mark size=1.5pt, thick] coordinates {
                (1.5, 4.93) (1.7, 4.66) (2.0, 4.33)
                (2.3, 4.37) (2.6, 4.42) (3.0, 8.24)
            };
            \node[anchor=north] at (axis cs: 2.0, 4.33) {\scriptsize \textbf{4.33}};
            \end{axis}
        \end{tikzpicture}
        \caption{\textbf{10 NFE Baseline.} Optimal $w=2.0$.}
        \label{fig:curve_10nfe}
    \end{minipage}
\end{figure}

\begin{figure}[h!]
    \centering
    \begin{minipage}[c]{0.38\linewidth}
        \centering
        \makeatletter\def\@captype{table}\makeatother 
        \caption{\textbf{20 NFE (ImageNet).} Best at $w=1.5$.}
        \label{tab:scale_20nfe}
        
        \begin{small}
        \begin{sc}
        \begin{tabular}{cc}
            \toprule
            Scale ($w$) & FID $\downarrow$ \\
            \midrule
            1.2 & 3.72 \\
            1.4 & 3.69 \\
            \textbf{1.5} & \textbf{3.68} \\
            1.6 & 3.69 \\
            1.8 & 3.71 \\
            2.0 & 3.72 \\
            2.5 & 3.77 \\
            \bottomrule
        \end{tabular}
        \end{sc}
        \end{small}
    \end{minipage}
    \hfill
    \begin{minipage}[c]{0.58\linewidth}
        \centering
        \begin{tikzpicture}
            \begin{axis}[
                width=\linewidth, height=5.0cm,
                xlabel={\textbf{Guidance Scale ($w$)}},
                ylabel={\textbf{FID $\downarrow$}},
                grid=major,
                ymin=3.65, ymax=3.80,
                xtick={1.2, 1.5, 1.8, 2.0, 2.5},
                ylabel near ticks, xlabel near ticks,
                tick label style={font=\scriptsize},
                label style={font=\footnotesize},
                title style={font=\small}
            ]
            \addplot[color=blue, mark=triangle*, mark size=2pt, thick] coordinates {
                (1.2, 3.72) (1.4, 3.69) (1.5, 3.68)
                (1.6, 3.69) (1.8, 3.71) (2.0, 3.72)
                (2.5, 3.77)
            };
            \node[anchor=north] at (axis cs: 1.5, 3.68) {\scriptsize \textbf{3.68}};
            \end{axis}
        \end{tikzpicture}
        \caption{\textbf{20 NFE Baseline.} Optimal $w=1.5$.}
        \label{fig:curve_20nfe}
    \end{minipage}
\end{figure}

\begin{figure}[h!]
    \centering
    \begin{minipage}[c]{0.38\linewidth}
        \centering
        \makeatletter\def\@captype{table}\makeatother 
        \caption{\textbf{40 NFE (ImageNet).} Best at $w=1.0$--$1.1$.}
        \label{tab:scale_40nfe}
        
        \begin{small}
        \begin{sc}
        \begin{tabular}{cc}
            \toprule
            Scale ($w$) & FID~\cite{NIPS2017_8a1d6947} $\downarrow$ \\
            \midrule
            0.7 & 3.57 \\
            0.9 & 3.55 \\
            \textbf{1.0} & \textbf{3.54} \\
            \textbf{1.1} & \textbf{3.54} \\
            1.4 & 3.55 \\
            1.5 & 3.57 \\
            \bottomrule
        \end{tabular}
        \end{sc}
        \end{small}
    \end{minipage}
    \hfill
    \begin{minipage}[c]{0.58\linewidth}
        \centering
        \begin{tikzpicture}
            \begin{axis}[
                width=\linewidth, height=5.0cm,
                xlabel={\textbf{Guidance Scale ($w$)}},
                ylabel={\textbf{FID $\downarrow$}},
                grid=major,
                ymin=3.53, ymax=3.58,
                xtick={0.7, 0.9, 1.0, 1.1, 1.4, 1.5},
                ylabel near ticks, xlabel near ticks,
                tick label style={font=\scriptsize},
                label style={font=\footnotesize},
                title style={font=\small}
            ]
            \addplot[color=black, mark=*, mark size=1.5pt, thick] coordinates {
                (0.7, 3.57) (0.9, 3.55) (1.0, 3.54)
                (1.1, 3.54) (1.4, 3.55) (1.5, 3.57)
            };
            \node[anchor=north] at (axis cs: 1.05, 3.54) {\scriptsize \textbf{3.54}};
            \end{axis}
        \end{tikzpicture}
        \caption{\textbf{40 NFE Baseline.} Optimal $w \approx 1.0$.}
        \label{fig:curve_40nfe}
    \end{minipage}
\end{figure}

\newpage
\subsection{Sensitivity Analysis of FMPG Parameters}
\label{app:fmpg_sensitivity}

We conducted fine-grained ablation studies on the \textbf{Start Ratio} ($\tau_{start}$) to identify the optimal configuration for Frequency-Modulated Prior Guidance (FMPG) under different computational budgets.

\textbf{Analysis at NFE = 10.}
For the low-budget regime, we fixed the hyperparameters as follows:
\begin{itemize}
    \setlength\itemsep{0em}
    \item \textbf{CFG}~\cite{ho2022classifier} Guidance Scale ($w$): 3.0
    \item \textbf{FMPG Low-Frequency Scale:} 1.45
    \item \textbf{FMPG High-Frequency Scale:} 1.35
\end{itemize}

Table~\ref{tab:fmpg_ablation_10} reports the FID scores~\cite{NIPS2017_8a1d6947}. The model achieves the best performance (FID = \textbf{3.86}) when frequency modulation starts at ratio \textbf{0.4}.

\begin{table}[h!] 
    \centering
    \caption{\textbf{Ablation on FMPG Start Ratio (NFE = 10).} 
    The nominal 10-NFE budget corresponds to 5 sampling steps under two-branch guidance. Since the first transition from $t=1.0$ to $t=0.9999$ is a booting step, the remaining four steps are effective denoising steps. We therefore keep only the discrete effective switching points present in our sweep. Fixed scales: CFG=2.0, Low=1.45, High=1.35.}
    \label{tab:fmpg_ablation_10}
    \begin{small}
    \begin{sc}
        \begin{tabular}{lcccccc}
            \toprule
            Start Ratio ($\tau$) 
            & 0.0 & 0.2 & \textbf{0.4} & 0.6 & 0.8 & 1.0 \\
            \midrule
            FID~\cite{NIPS2017_8a1d6947} ($\downarrow$) 
            & 4.33 & 3.89 & \textbf{3.86} & 4.05 & 4.51 & 5.08 \\
            \bottomrule
        \end{tabular}
    \end{sc}
    \end{small}
\end{table}

\textbf{Analysis at NFE = 20.}
For the standard budget regime, the hyperparameters were set to:
\begin{itemize}
    \setlength\itemsep{0em}
    \item \textbf{CFG}~\cite{ho2022classifier} Guidance Scale ($w$): 2.5
    \item \textbf{FMPG Low-Frequency Scale:} 1.2
    \item \textbf{FMPG High-Frequency Scale:} 1.11
\end{itemize}

Table~\ref{tab:fmpg_ablation_20} presents the results. The optimal sweet spot shifts slightly, achieving the best FID~\cite{NIPS2017_8a1d6947} (\textbf{3.53}) at start ratio \textbf{0.3}.

\begin{table}[h!] 
    \centering
    \caption{\textbf{Ablation on FMPG Start Ratio (NFE = 20).} Fixed scales: CFG=2.5, Low=1.2, High=1.11. Optimal start ratio is 0.3.}
    \label{tab:fmpg_ablation_20}
    \begin{small}
    \begin{sc}
        \begin{tabular}{lcccccccc}
            \toprule
            Start Ratio ($\tau$) & 0.0 & 0.2 & \textbf{0.3} & 0.4 & 0.5 & 0.6 & 0.7 & 1.0 \\
            \midrule
            FID~\cite{NIPS2017_8a1d6947} ($\downarrow$)    & 3.69 & 3.55 & \textbf{3.53} & 3.54 & 3.55 & 3.58 & 3.61 & 3.74 \\
            \bottomrule
        \end{tabular}
    \end{sc}
    \end{small}
\end{table}

\textbf{Analysis at NFE = 40.}
For the high-fidelity regime, the hyperparameters were configured as:
\begin{itemize}
    \setlength\itemsep{0em}
    \item \textbf{CFG}~\cite{ho2022classifier} Guidance Scale ($w$): 2.1
    \item \textbf{FMPG Low-Frequency Scale:} 1.1
    \item \textbf{FMPG High-Frequency Scale:} 1.04
\end{itemize}

Table~\ref{tab:fmpg_ablation_40} summarizes the FID scores~\cite{NIPS2017_8a1d6947}. Similar to the low-budget regime, we observe an optimal start ratio at \textbf{0.4} with an FID of \textbf{3.47}.
\begin{table}[h!] 
    \centering
    \caption{\textbf{Ablation on FMPG Start Ratio (NFE = 40).} Fixed scales: CFG=2.1, Low=1.1, High=1.04. Optimal start ratio is 0.4.}
    \label{tab:fmpg_ablation_40}
    \begin{small}
    \begin{sc}
        \begin{tabular}{lcccccccccc}
            \toprule
            Start Ratio ($\tau$) & 0.0 & 0.1 & 0.2 & 0.3 & \textbf{0.4} & 0.5 & 0.6 & 0.8 & 0.9 & 1.0 \\
            \midrule
            FID~\cite{NIPS2017_8a1d6947} ($\downarrow$)    & 3.54 & 3.53 & 3.51 & 3.49 & \textbf{3.47} & 3.50 & 3.58 & 3.75 & 3.82 & 3.94 \\
            \bottomrule
        \end{tabular}
    \end{sc}
    \end{small}
\end{table}

\section{Qualitative Results}
\label{app:qualitative_results}

We provide additional visual results to demonstrate the effectiveness of our method across different domains: DIODE (outdoor scenes), Edges2Handbags (edge-guided generation), and ImageNet (class-conditional in-painting). For all experiments, we consistently use the pre-trained checkpoints provided by DBIM~\citep{zheng2025diffusion} and a first-order ODE sampler. 
\subsection{DIODE: Generation Quality across Different NFEs}
We visualize the evolution of generation quality on the DIODE dataset.

\begin{figure}[htbp]
    \centering
    \begin{minipage}[t]{0.46\linewidth}
        \centering
        \includegraphics[width=\linewidth]{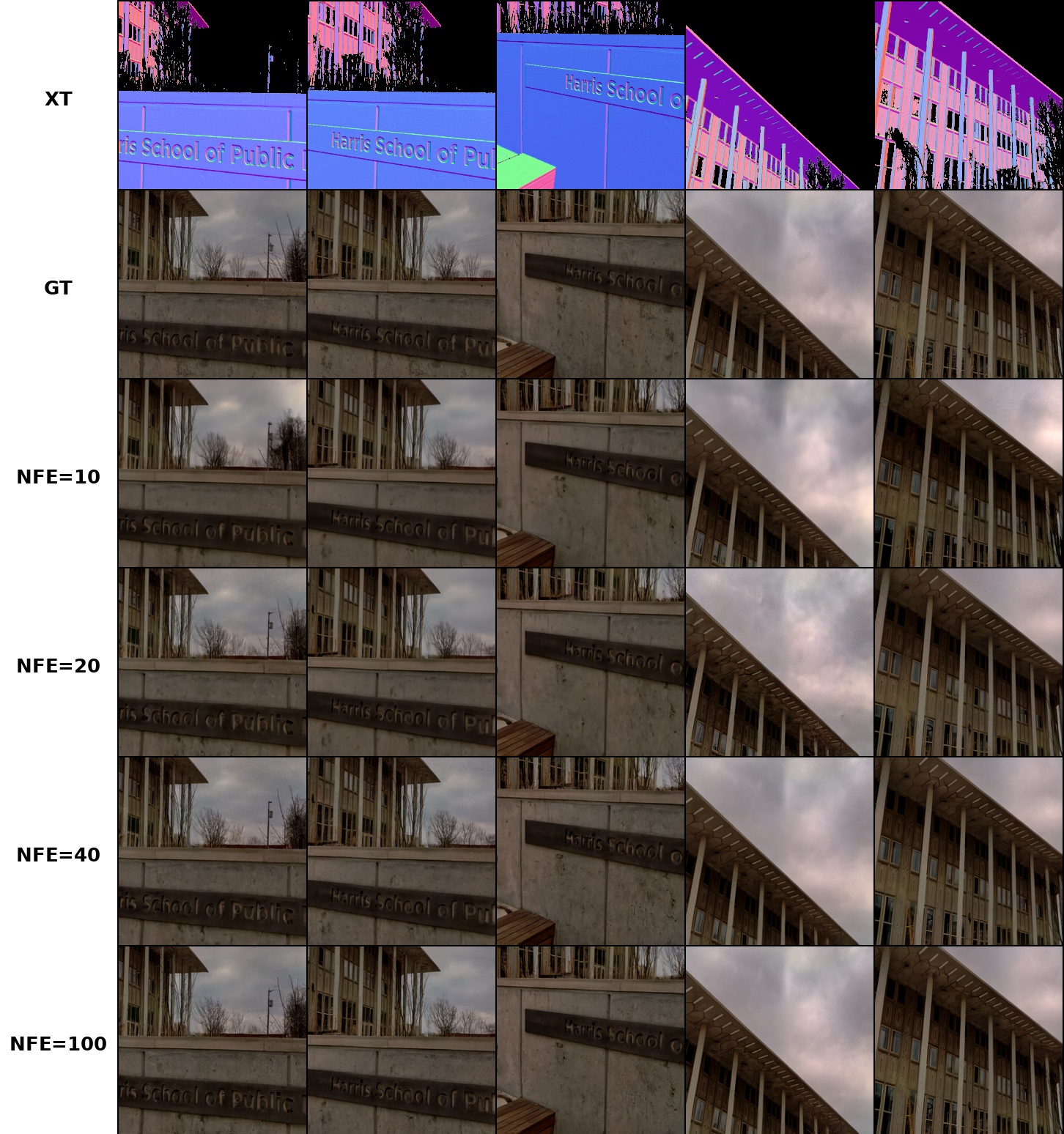}
    \end{minipage}
    \hfill 
    \begin{minipage}[t]{0.46\linewidth}
        \centering
        \includegraphics[width=\linewidth]{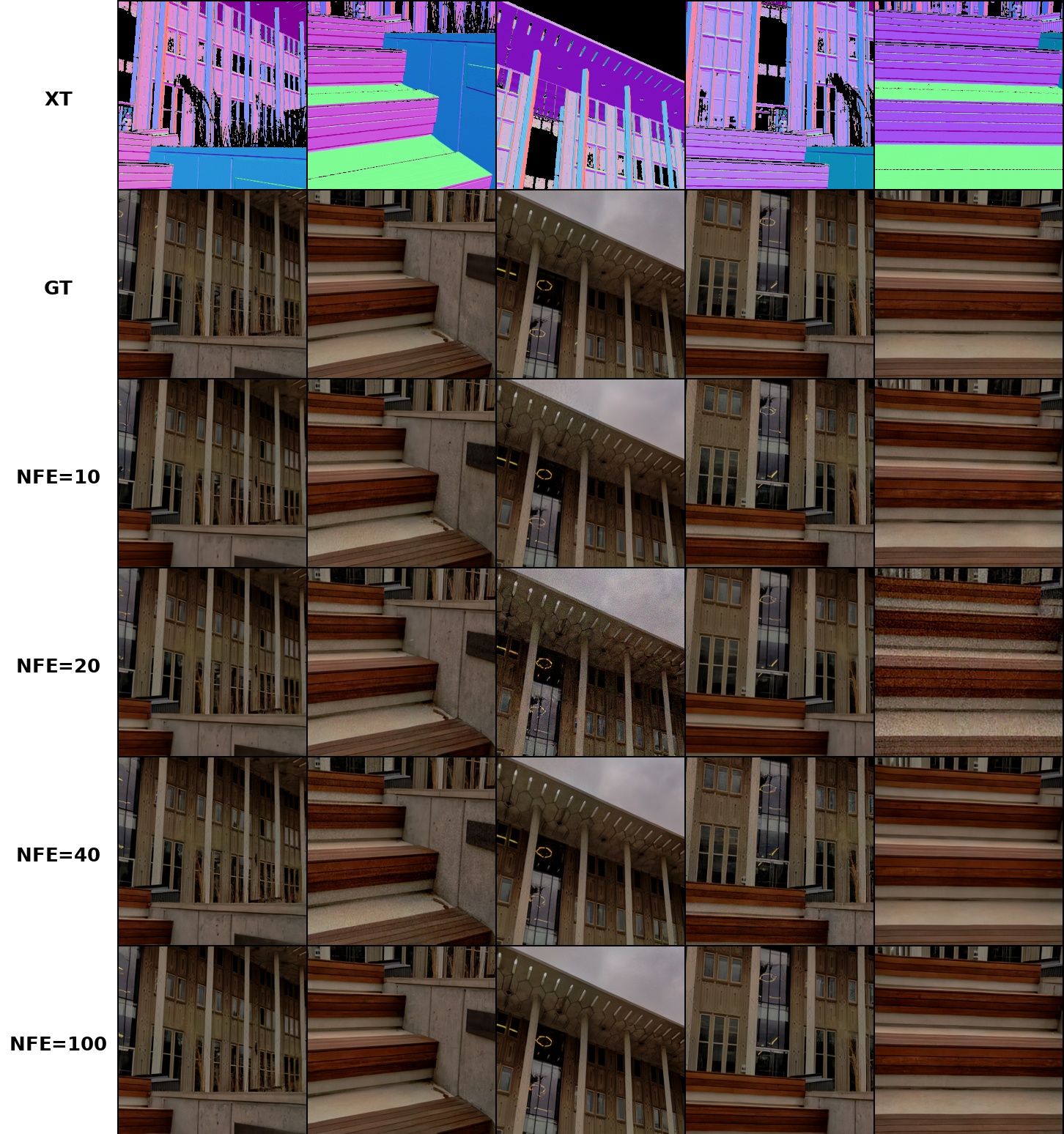}
    \end{minipage}
    \caption{\textbf{Qualitative Results on DIODE (Part I).} Visual comparison of the first two samples.}
    \label{fig:diode_part1}
\end{figure}

\begin{figure}[htbp]
    \centering
    \begin{minipage}[t]{0.495\linewidth}
        \centering
        \includegraphics[width=\linewidth]{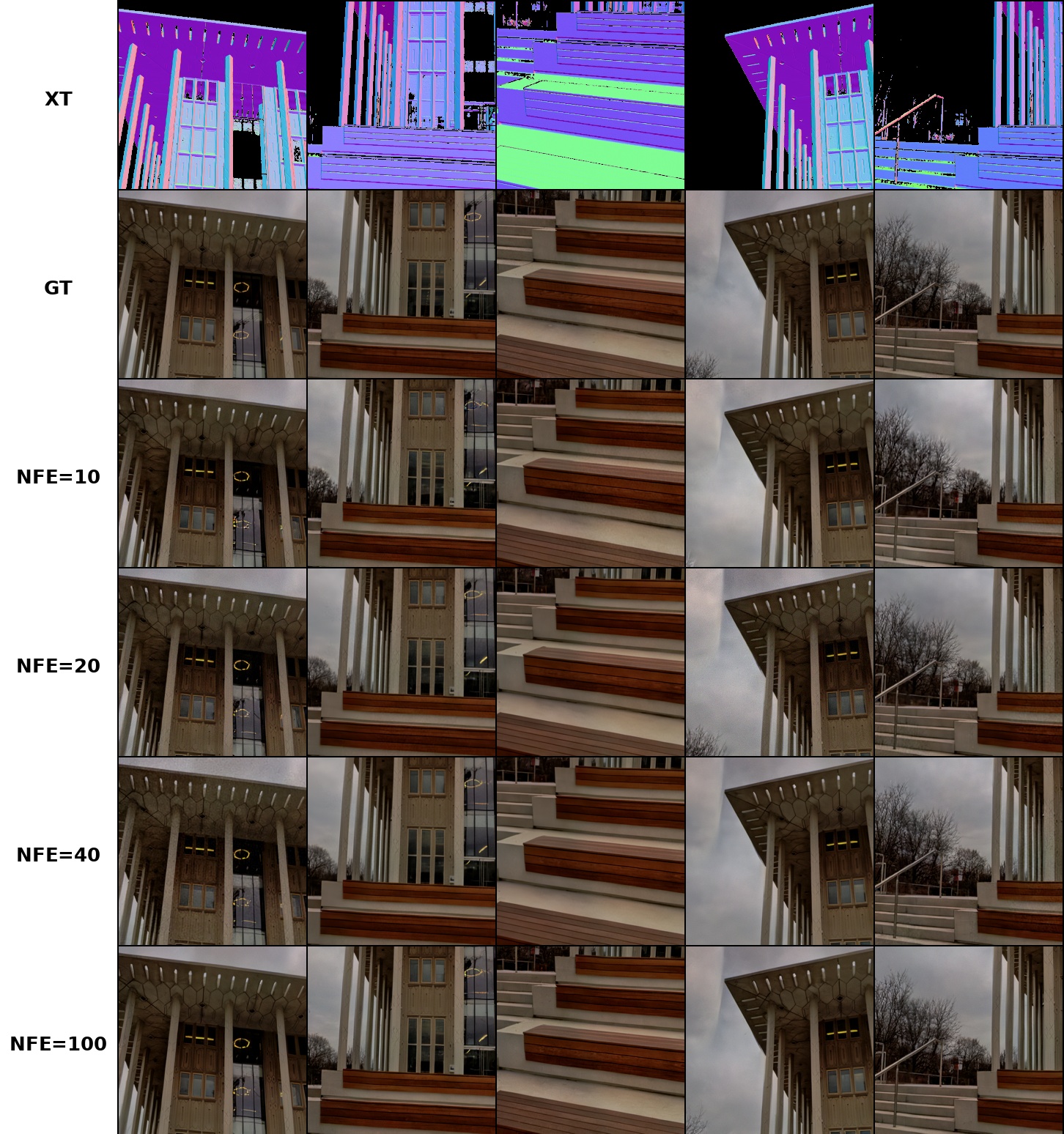}
    \end{minipage}
    \hfill 
    \begin{minipage}[t]{0.495\linewidth}
        \centering
        \includegraphics[width=\linewidth]{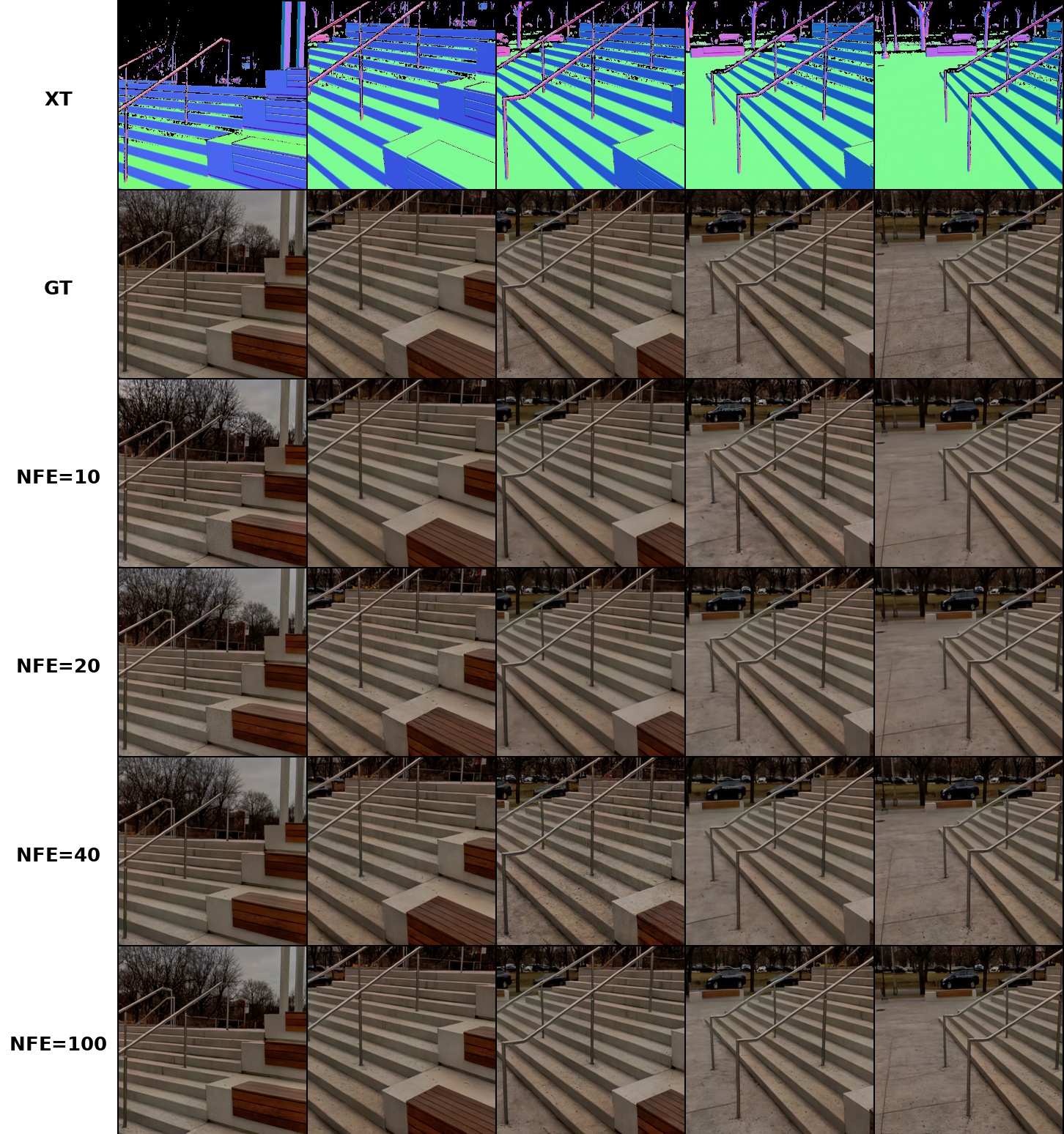}
    \end{minipage}
    
    \vspace{2mm}
    
    \begin{minipage}[t]{0.495\linewidth}
        \centering
        \includegraphics[width=\linewidth]{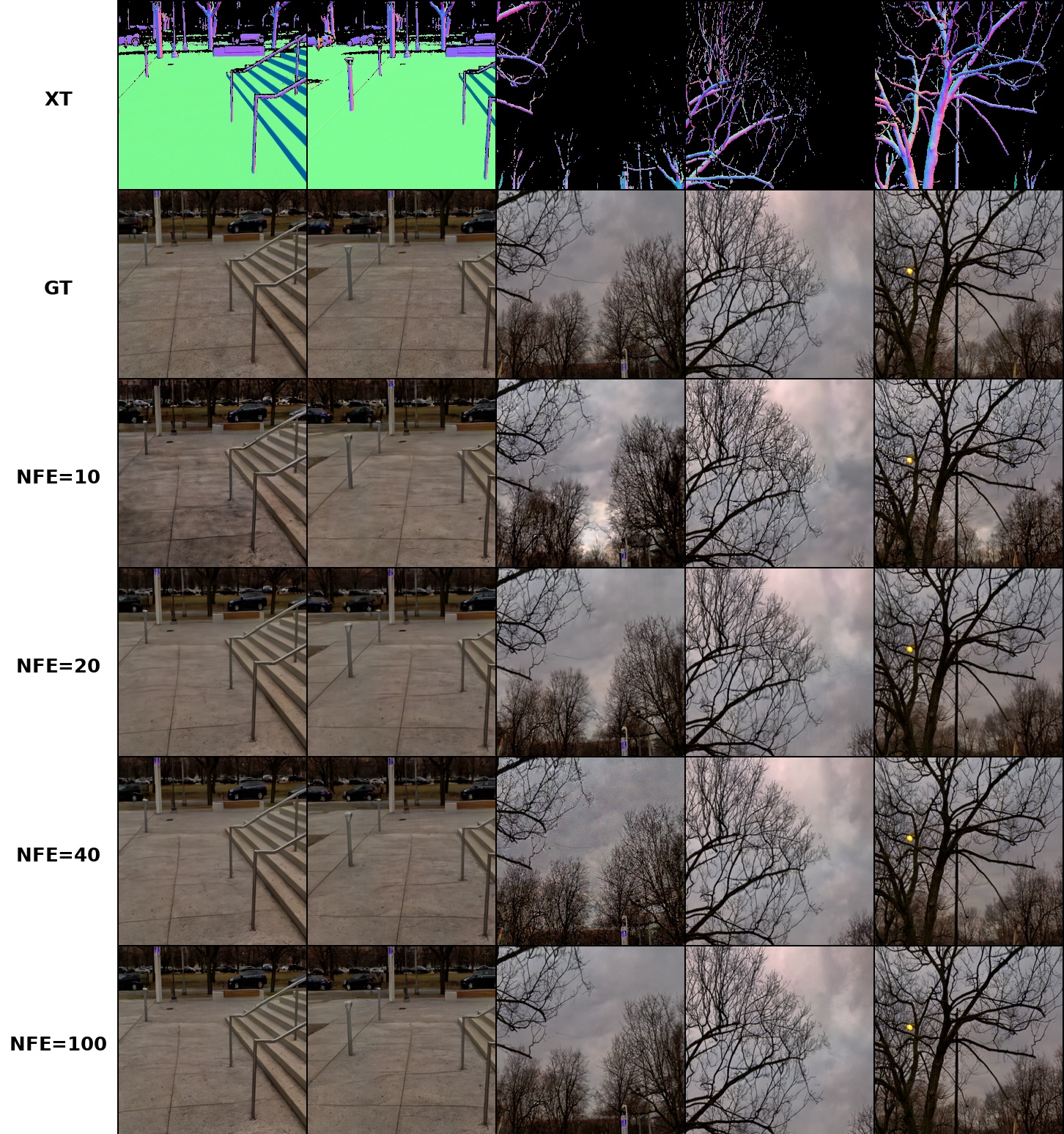}
    \end{minipage}
    \hfill
    \begin{minipage}[t]{0.495\linewidth}
        \centering
        \includegraphics[width=\linewidth]{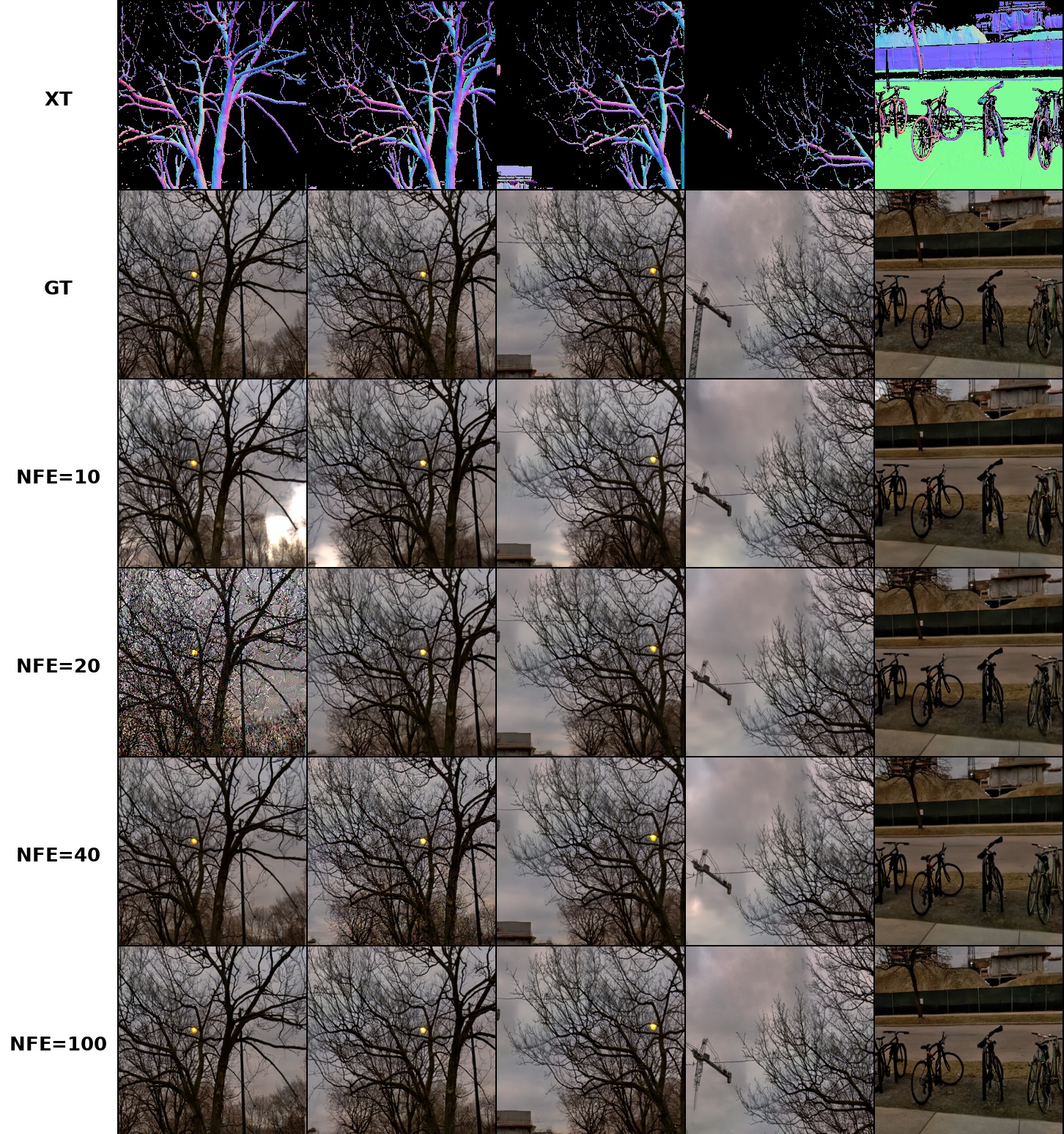}
    \end{minipage}
    
    \caption{\textbf{Qualitative Results on DIODE (Part II).} Visual comparison of additional samples.}
    \label{fig:diode_part2}
\end{figure}

\subsection{Edges2Handbags: Texture Evolution}
We further demonstrate robustness on the Edges2Handbags dataset.

\begin{figure}[htbp]
    \centering
    \begin{minipage}[t]{0.495\linewidth}
        \centering
        \includegraphics[width=\linewidth]{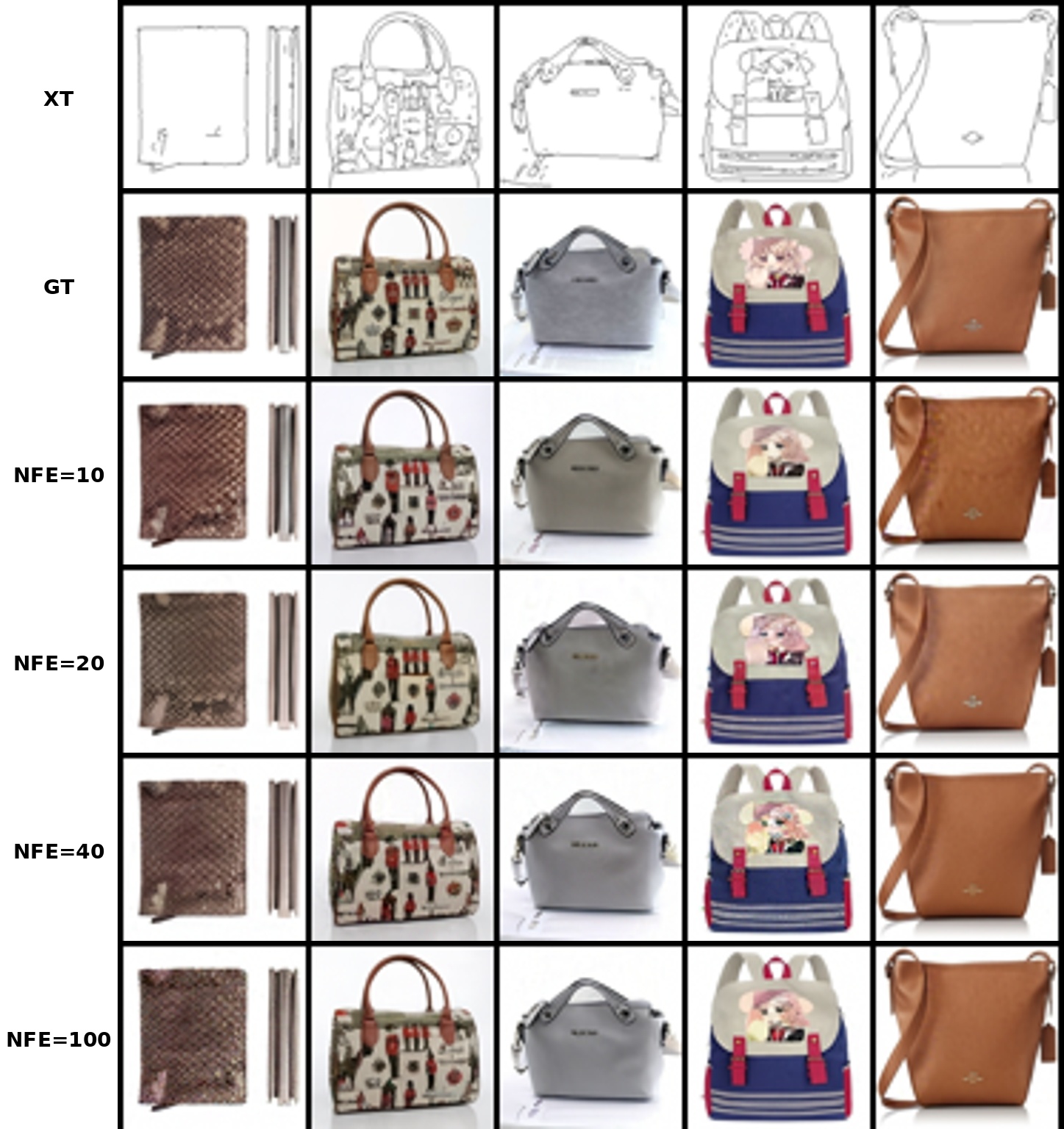}
    \end{minipage}
    \hfill 
    \begin{minipage}[t]{0.495\linewidth}
        \centering
        \includegraphics[width=\linewidth]{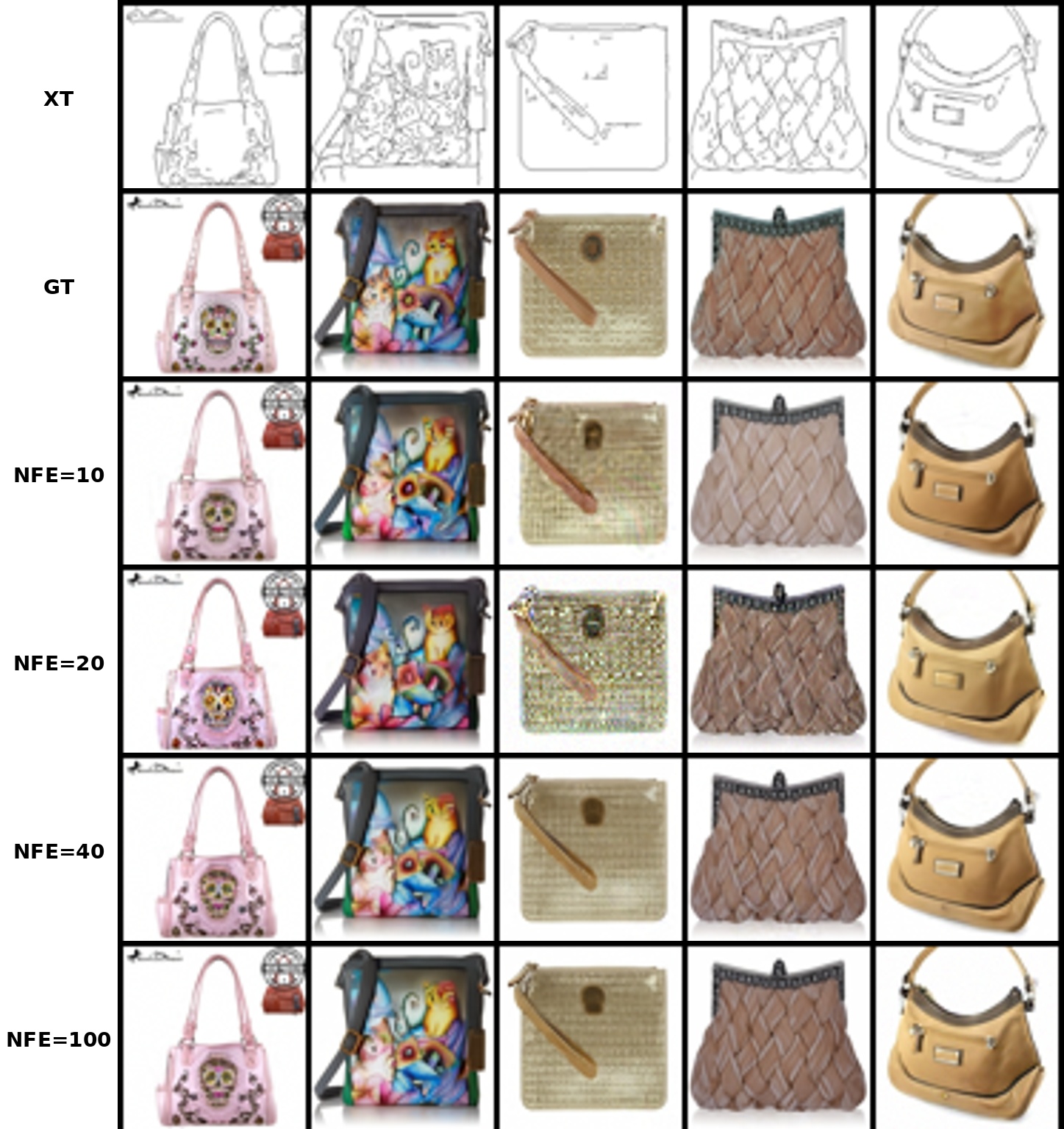}
    \end{minipage}
    
    \vspace{2mm}
    
    \begin{minipage}[t]{0.495\linewidth}
        \centering
        \includegraphics[width=\linewidth]{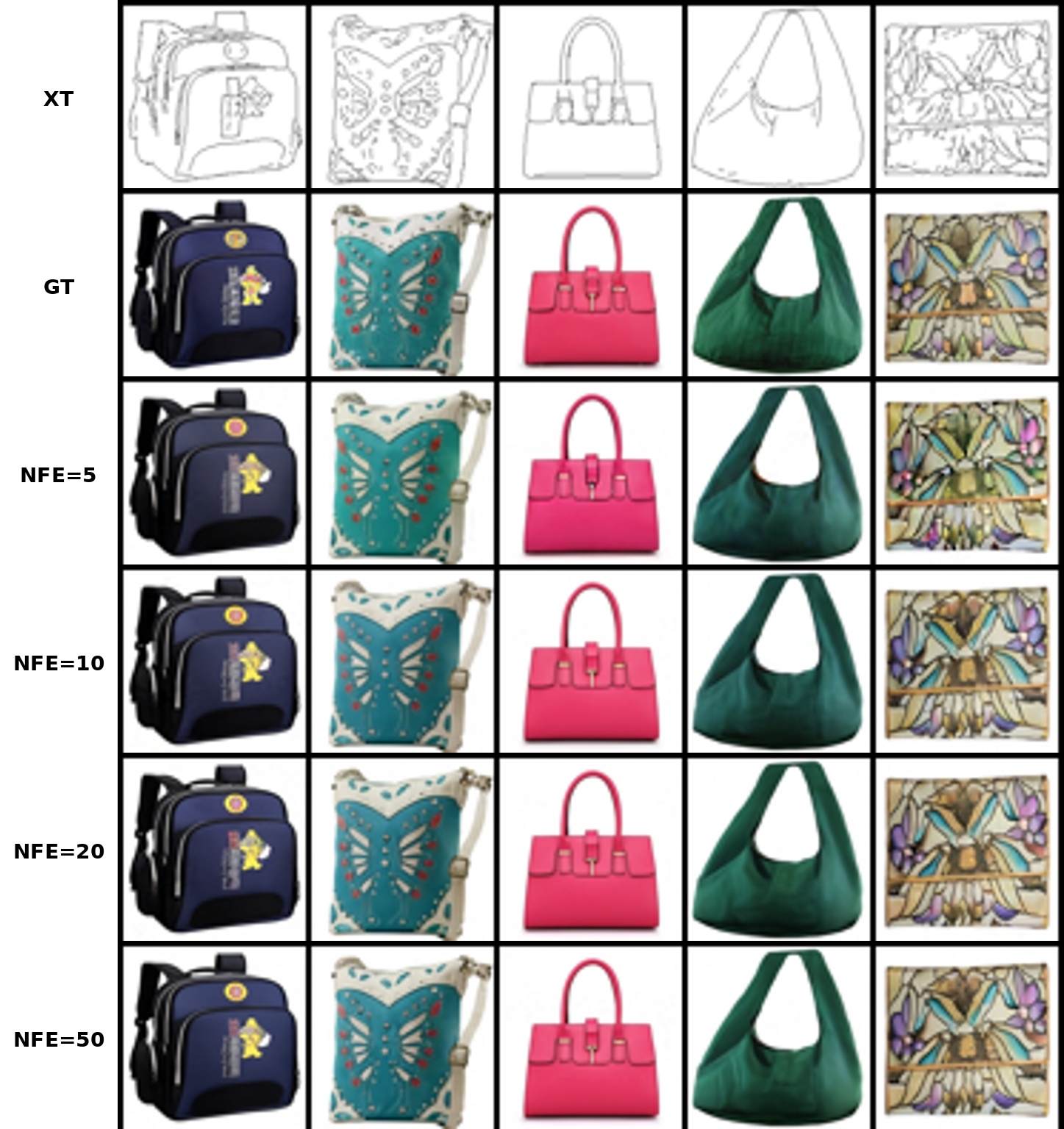}
    \end{minipage}
    \hfill
    \begin{minipage}[t]{0.495\linewidth}
        \centering
        \includegraphics[width=\linewidth]{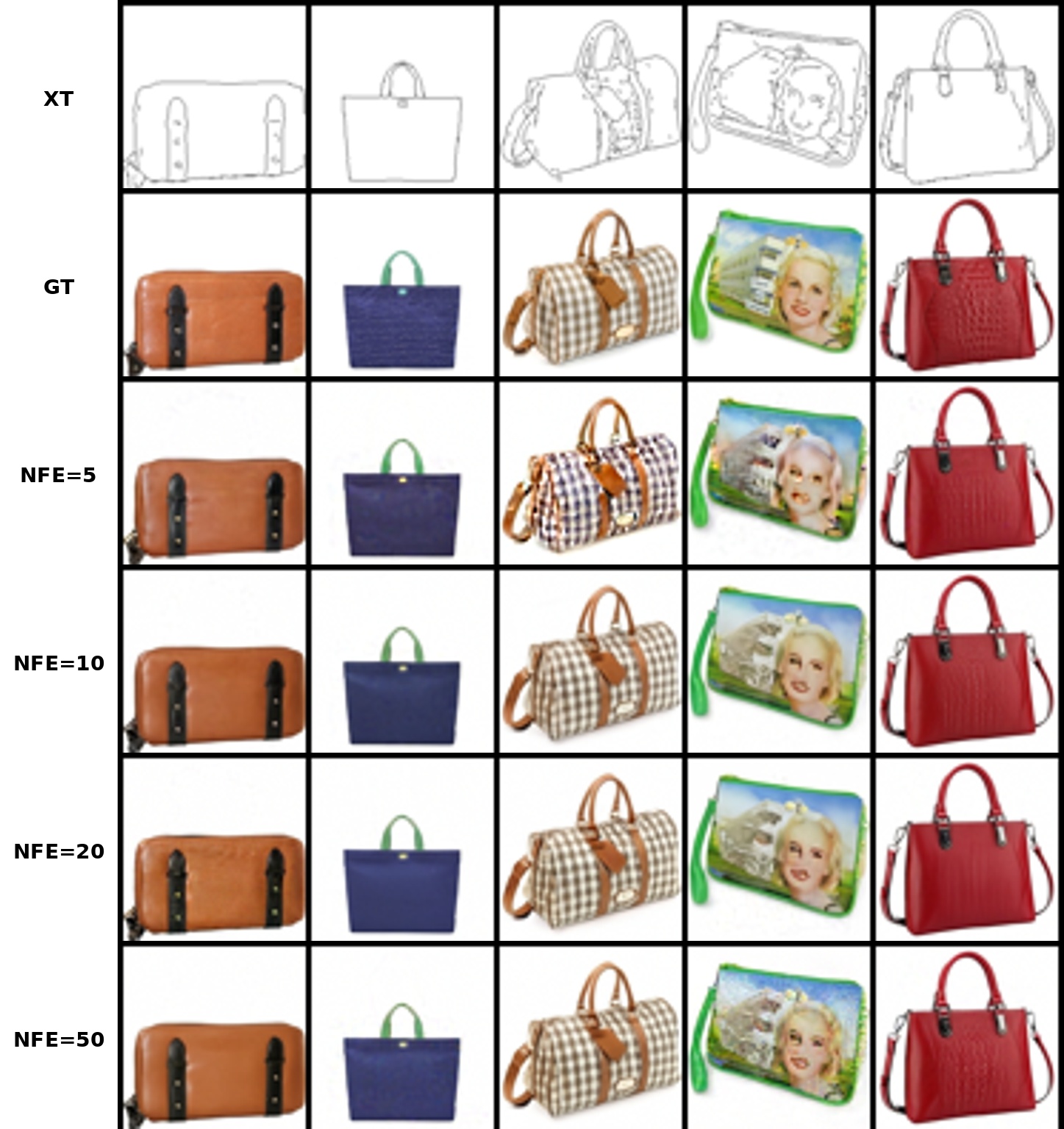}
    \end{minipage}
    
    \caption{\textbf{Qualitative Results on Edges2Handbags (Part I).} Visual comparison of additional samples.}
    \label{fig:e2h_part2}
\end{figure}

\begin{figure}[htbp]
    \centering
    \begin{minipage}[t]{0.495\linewidth}
        \centering
        \includegraphics[width=\linewidth]{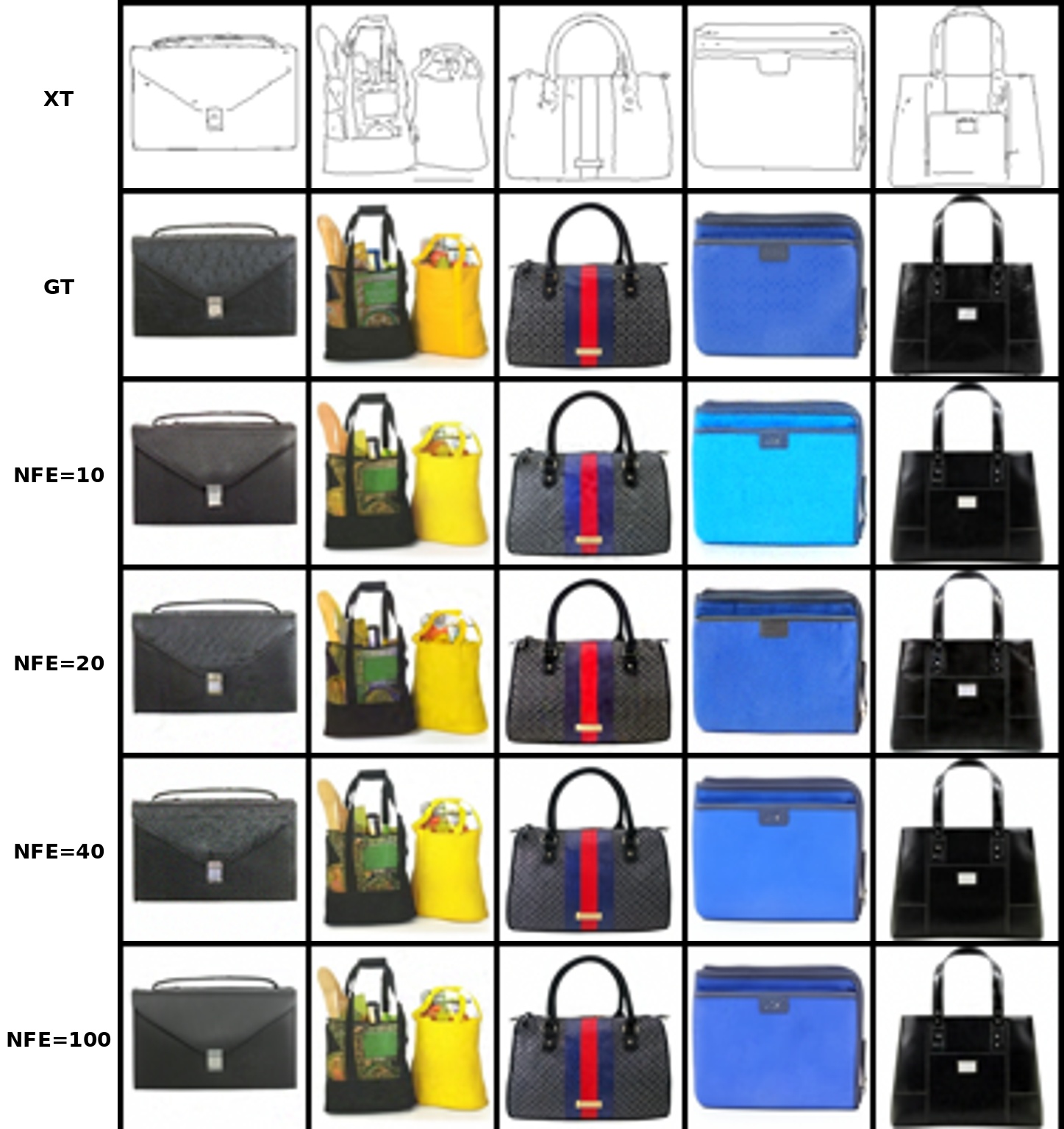}
    \end{minipage}
    \hfill 
    \begin{minipage}[t]{0.495\linewidth}
        \centering
        \includegraphics[width=\linewidth]{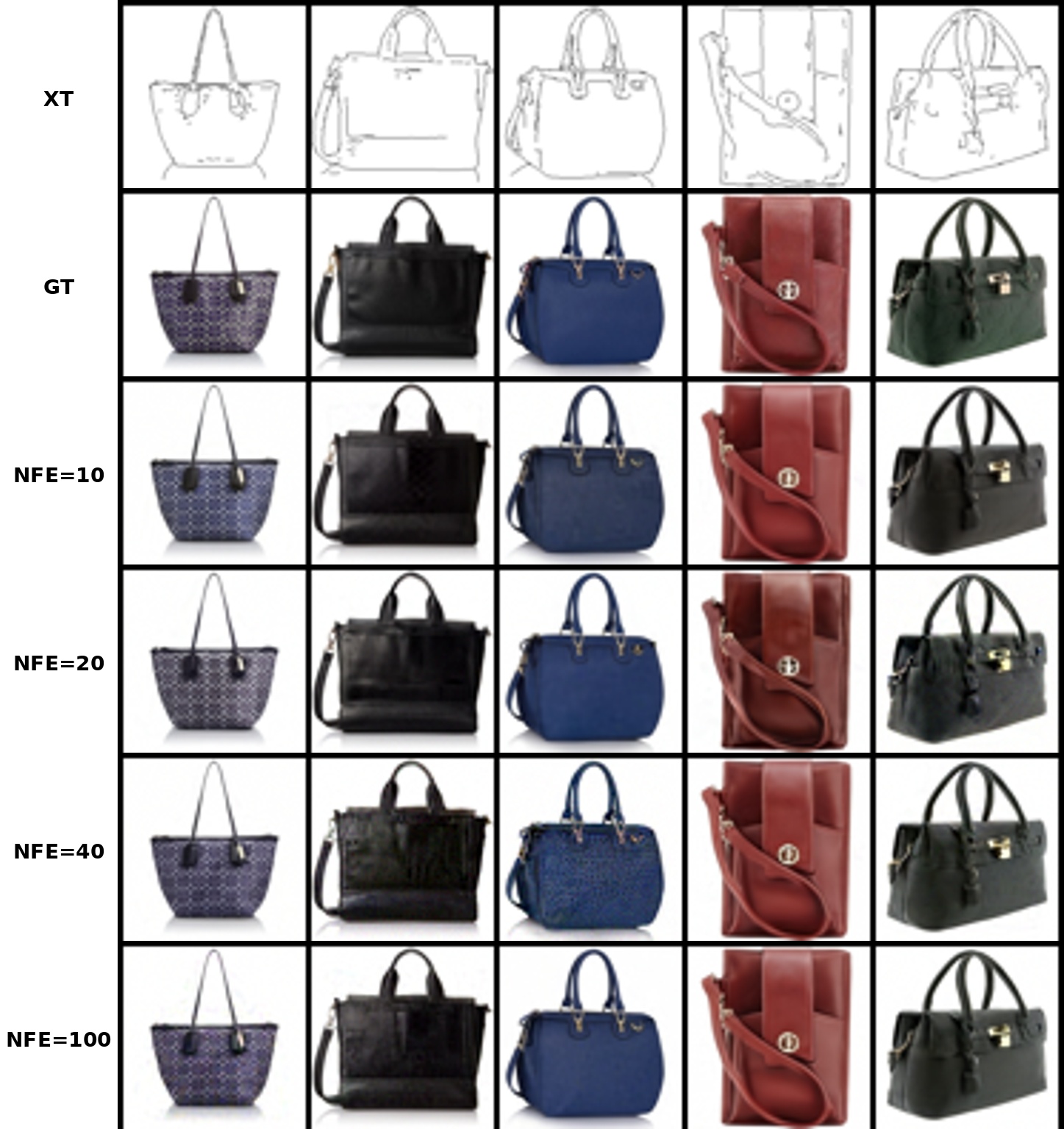}
    \end{minipage}
    
    \vspace{2mm}
    
    \begin{minipage}[t]{0.495\linewidth}
        \centering
        \includegraphics[width=\linewidth]{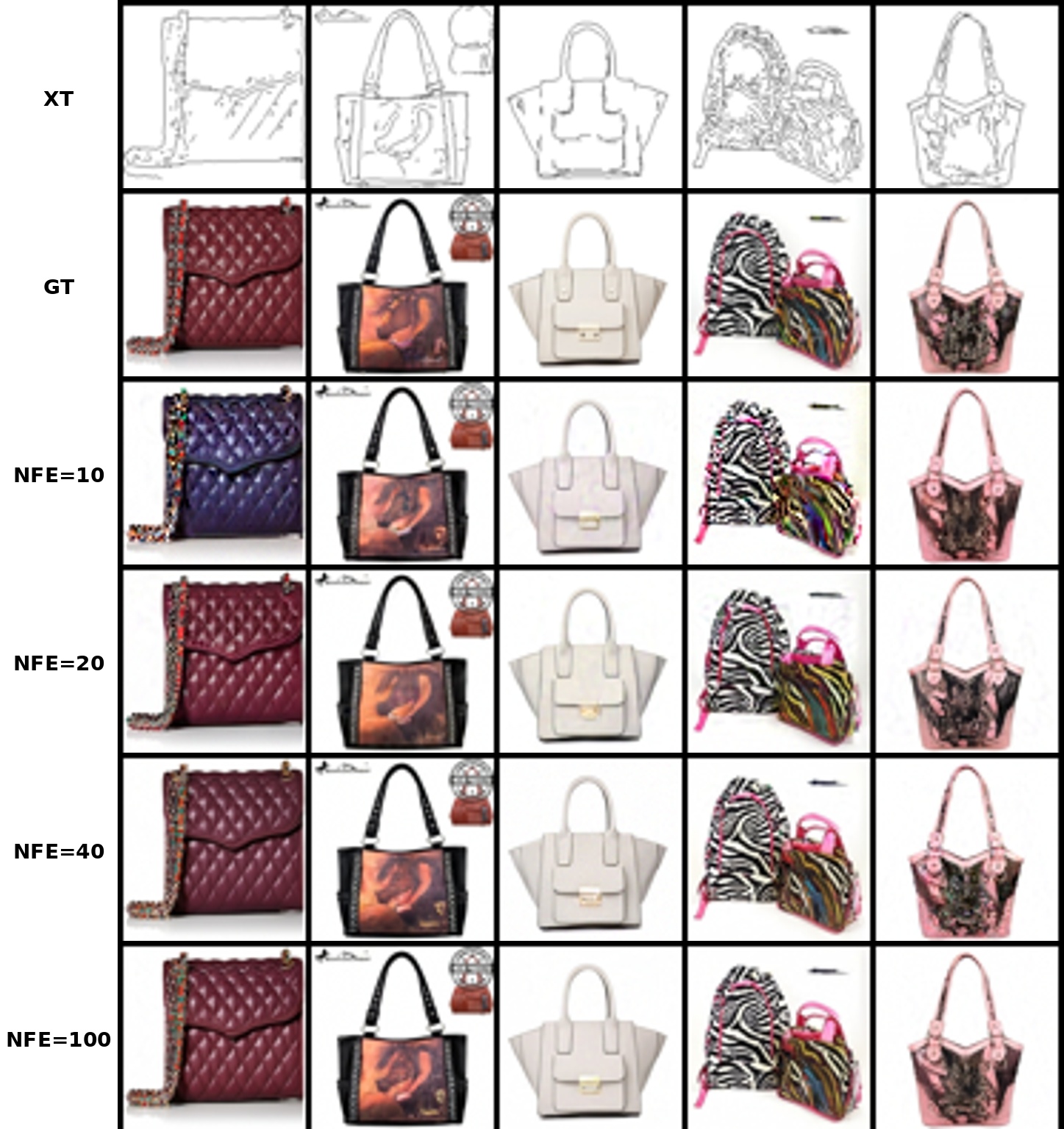}
    \end{minipage}
    \hfill
    \begin{minipage}[t]{0.495\linewidth}
        \centering
        \includegraphics[width=\linewidth]{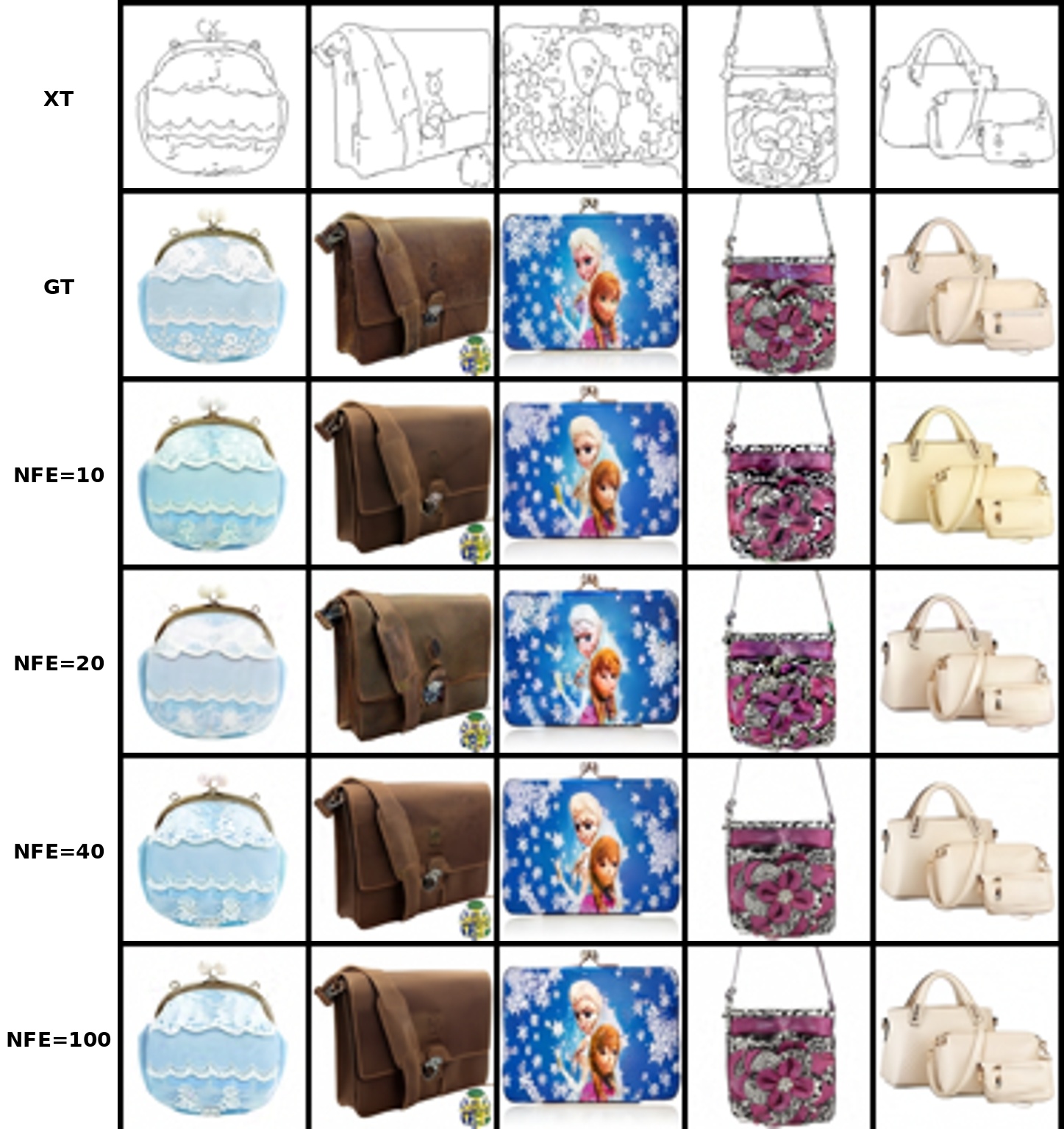}
    \end{minipage}
    
    \caption{\textbf{Qualitative Results on Edges2Handbags (Part II).} Visual comparison of additional samples.}
    \label{fig:e2h_part2}
\end{figure}

\subsection{ImageNet: High-Fidelity Synthesis}
Finally, we present results on the challenging ImageNet ($256\times256$) dataset. 

\begin{figure}[htbp]
    \centering
    \begin{minipage}[t]{0.495\linewidth}
        \centering
        \includegraphics[width=\linewidth]{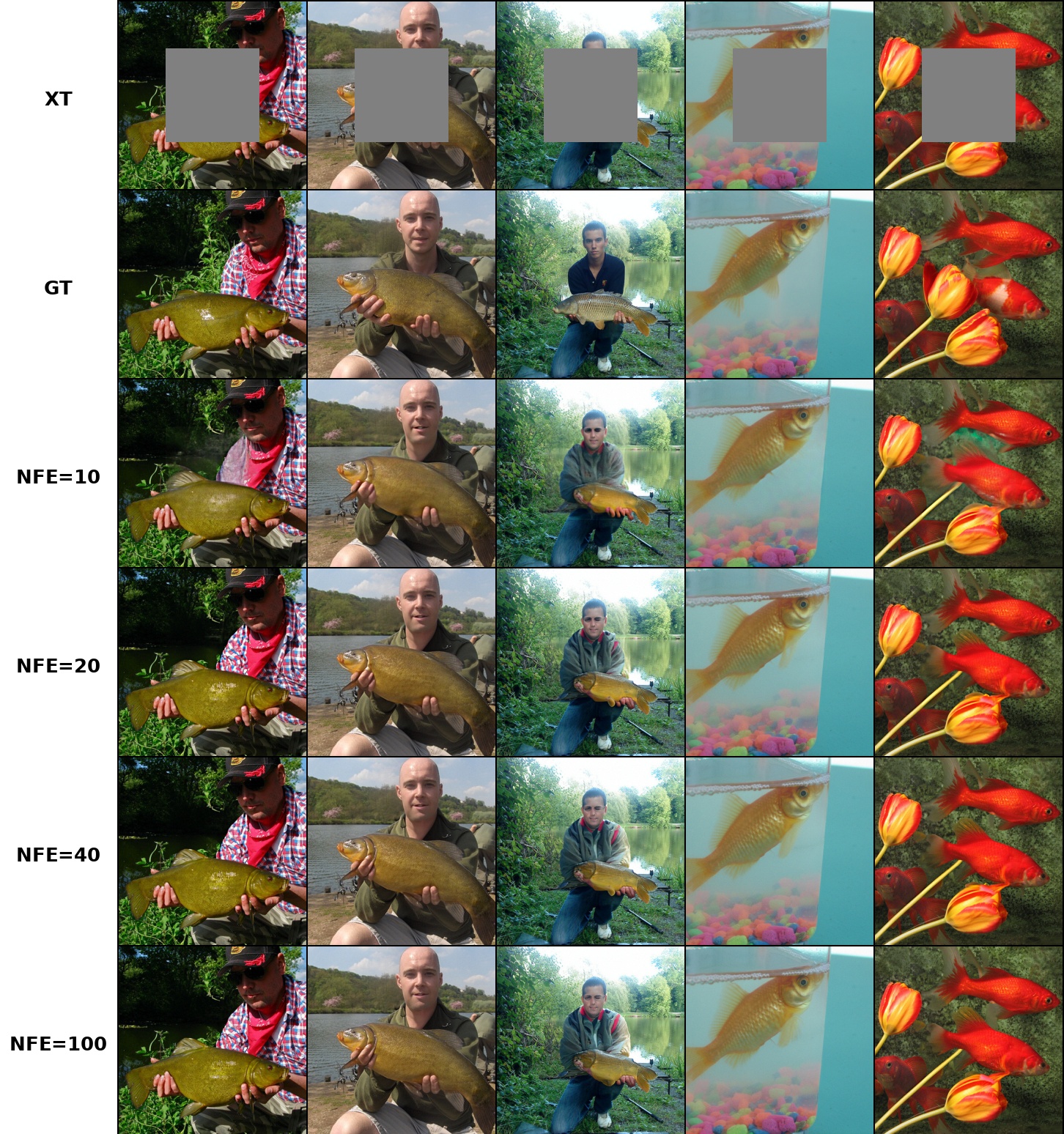}
    \end{minipage}
    \hfill 
    \begin{minipage}[t]{0.495\linewidth}
        \centering
        \includegraphics[width=\linewidth]{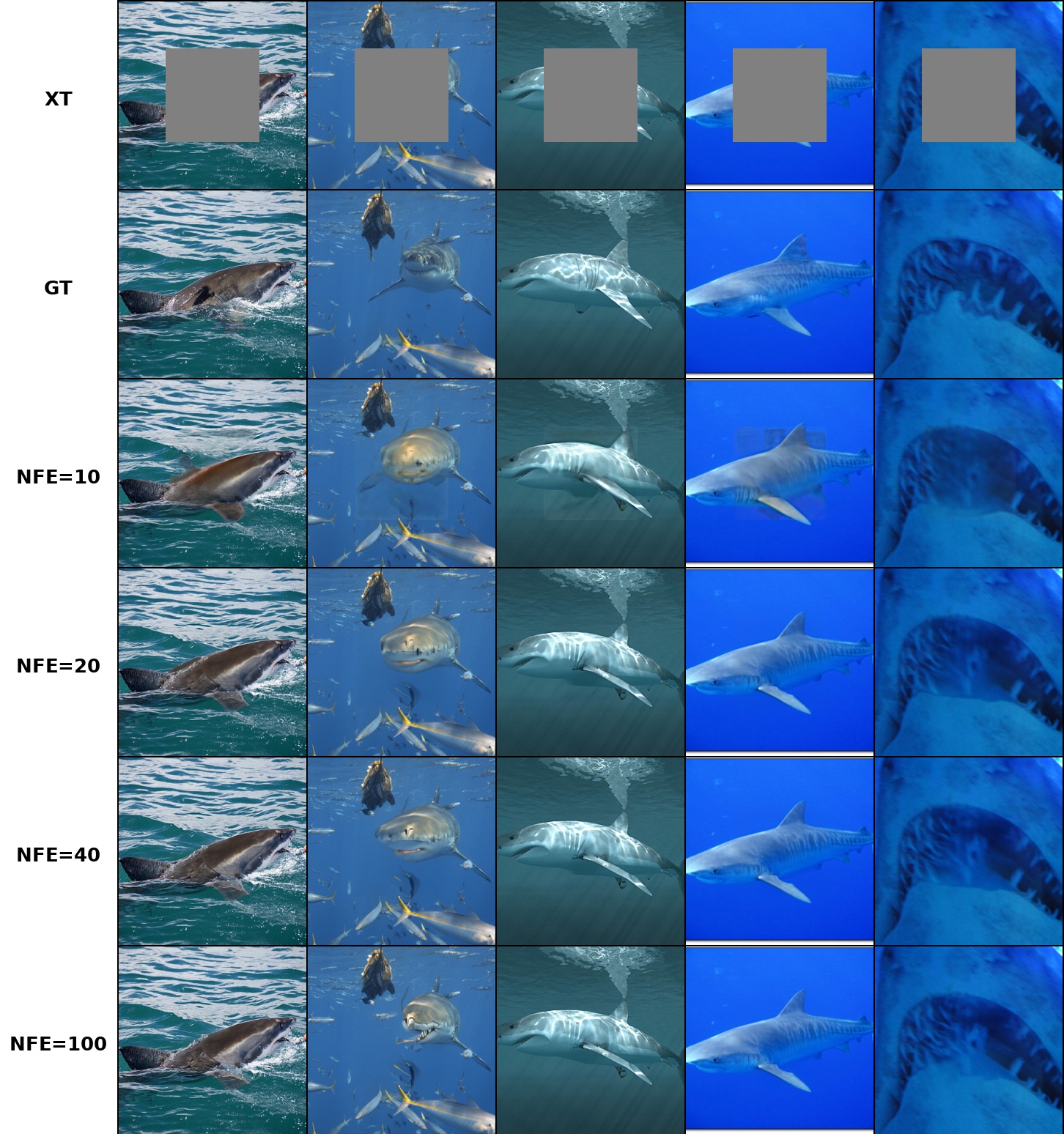}
    \end{minipage}
    
    \vspace{2mm}
    
    \begin{minipage}[t]{0.495\linewidth}
        \centering
        \includegraphics[width=\linewidth]{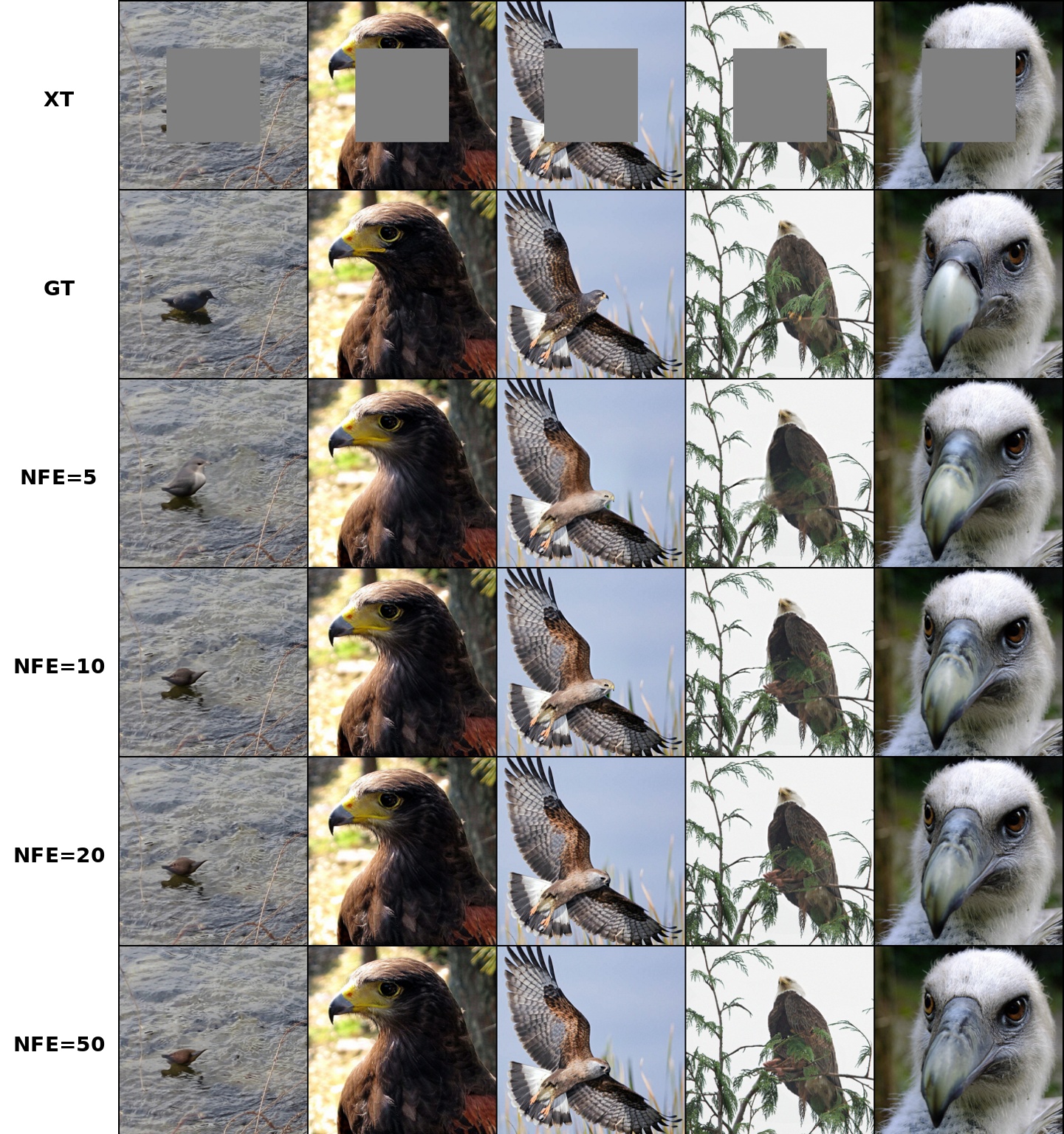}
    \end{minipage}
    \hfill
    \begin{minipage}[t]{0.495\linewidth}
        \centering
        \includegraphics[width=\linewidth]{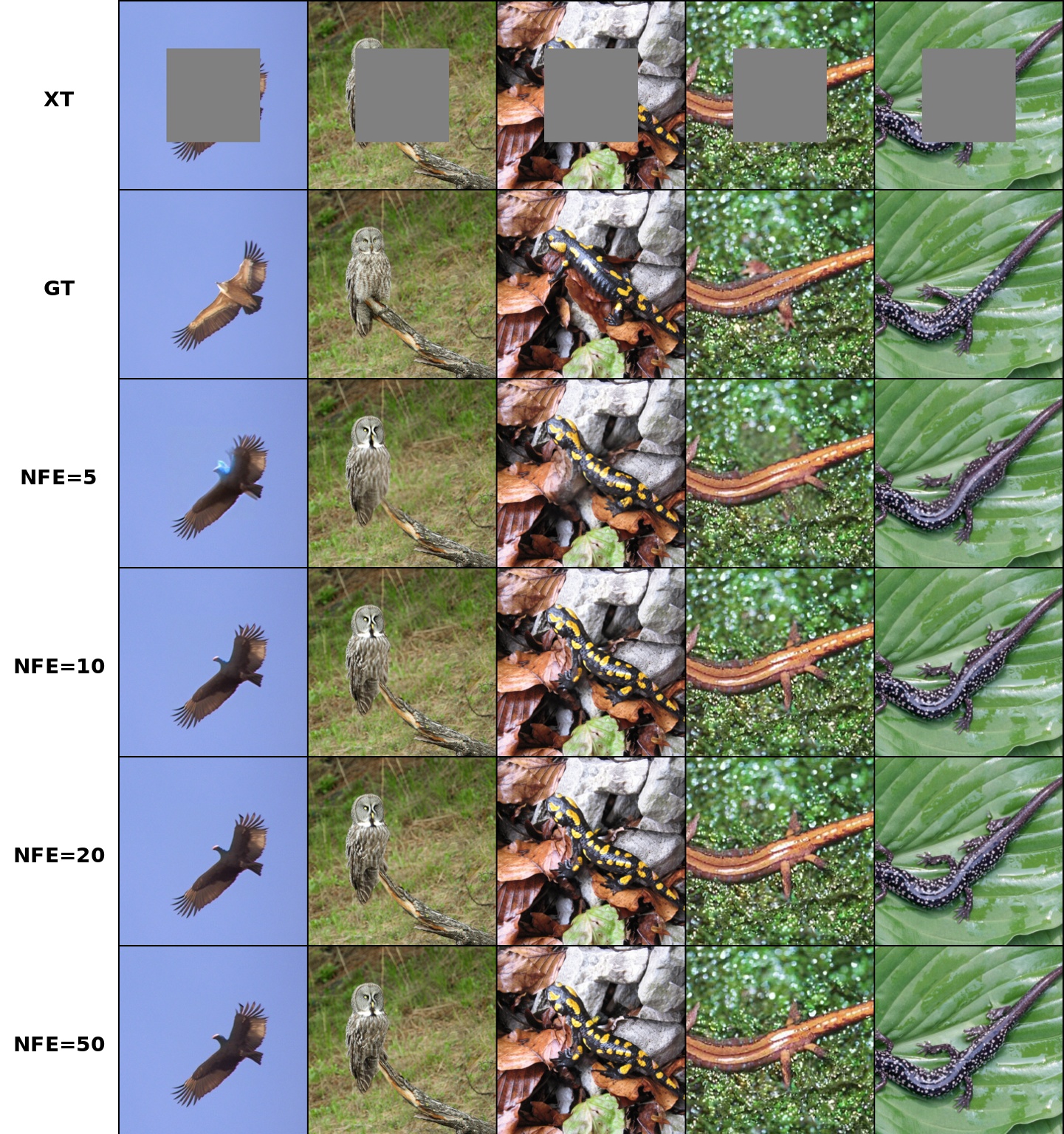}
    \end{minipage}
    
    \caption{\textbf{Qualitative Results on ImageNet (Part I).} Visual comparison of additional samples.}
    \label{fig:imagenet_part2}
\end{figure}

\begin{figure}[htbp]
    \centering
    \begin{minipage}[t]{0.495\linewidth}
        \centering
        \includegraphics[width=\linewidth]{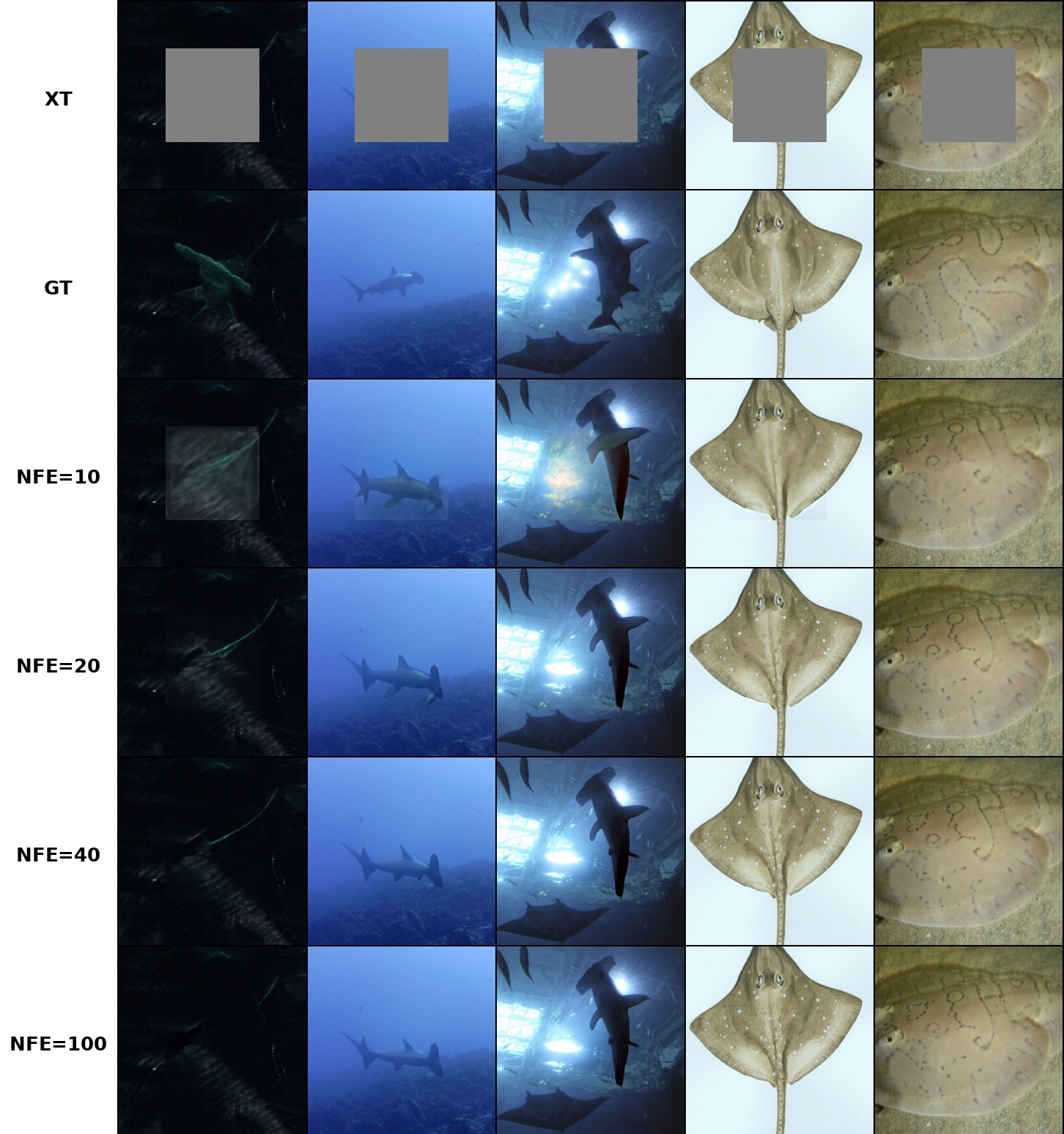}
    \end{minipage}
    \hfill 
    \begin{minipage}[t]{0.495\linewidth}
        \centering
        \includegraphics[width=\linewidth]{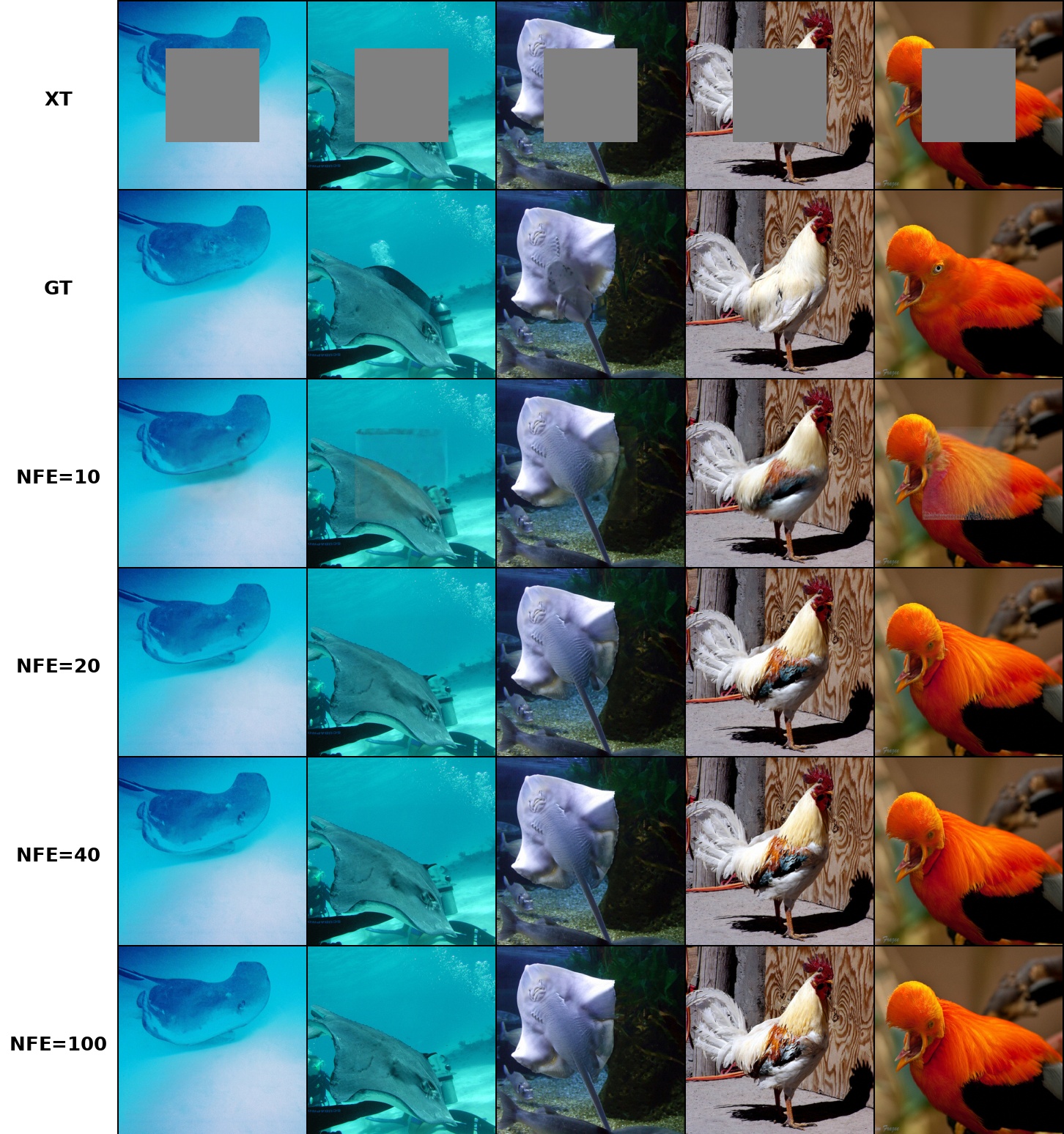}
    \end{minipage}
    
    \vspace{2mm}
    
    \begin{minipage}[t]{0.495\linewidth}
        \centering
        \includegraphics[width=\linewidth]{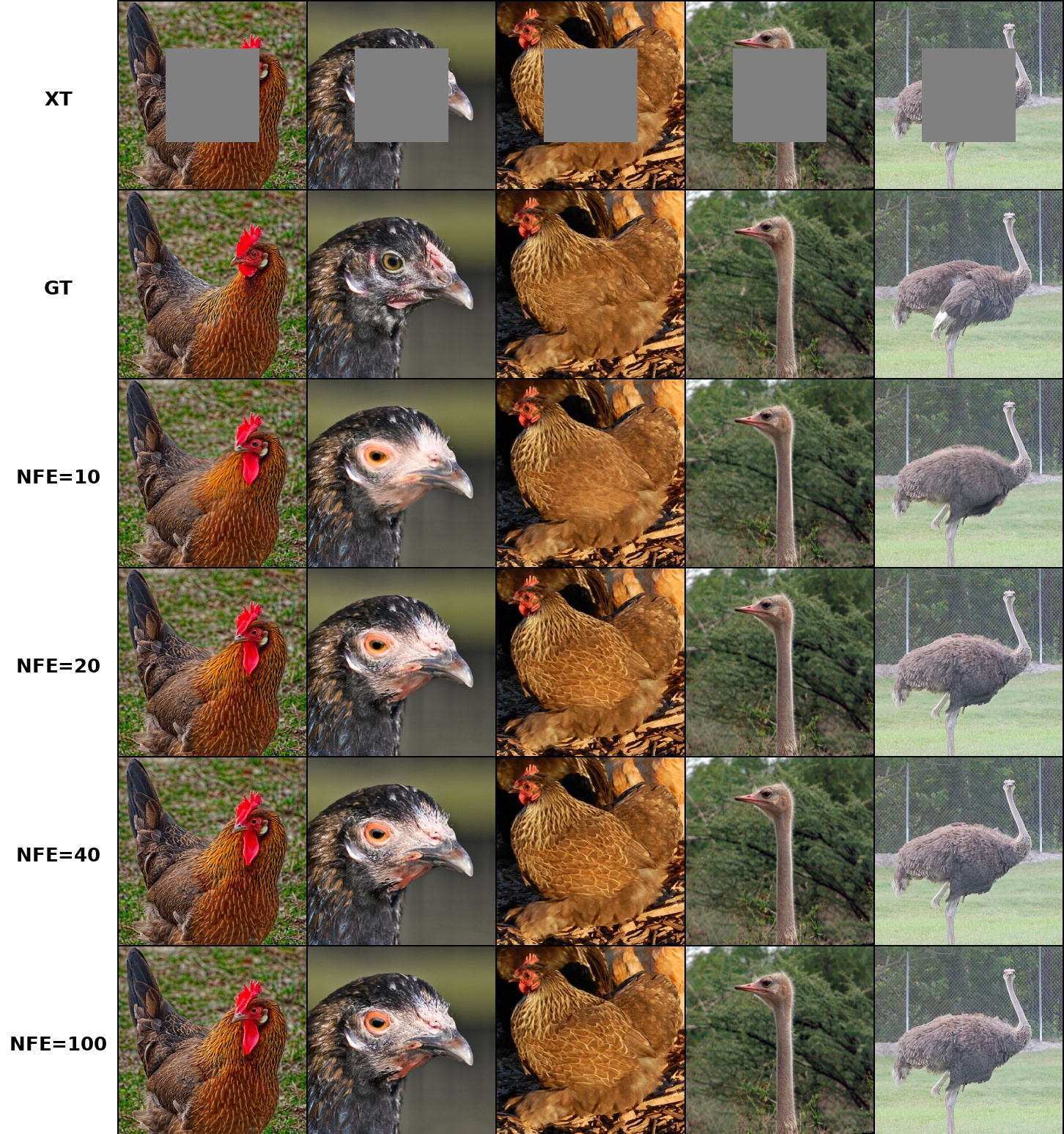}
    \end{minipage}
    \hfill
    \begin{minipage}[t]{0.495\linewidth}
        \centering
        \includegraphics[width=\linewidth]{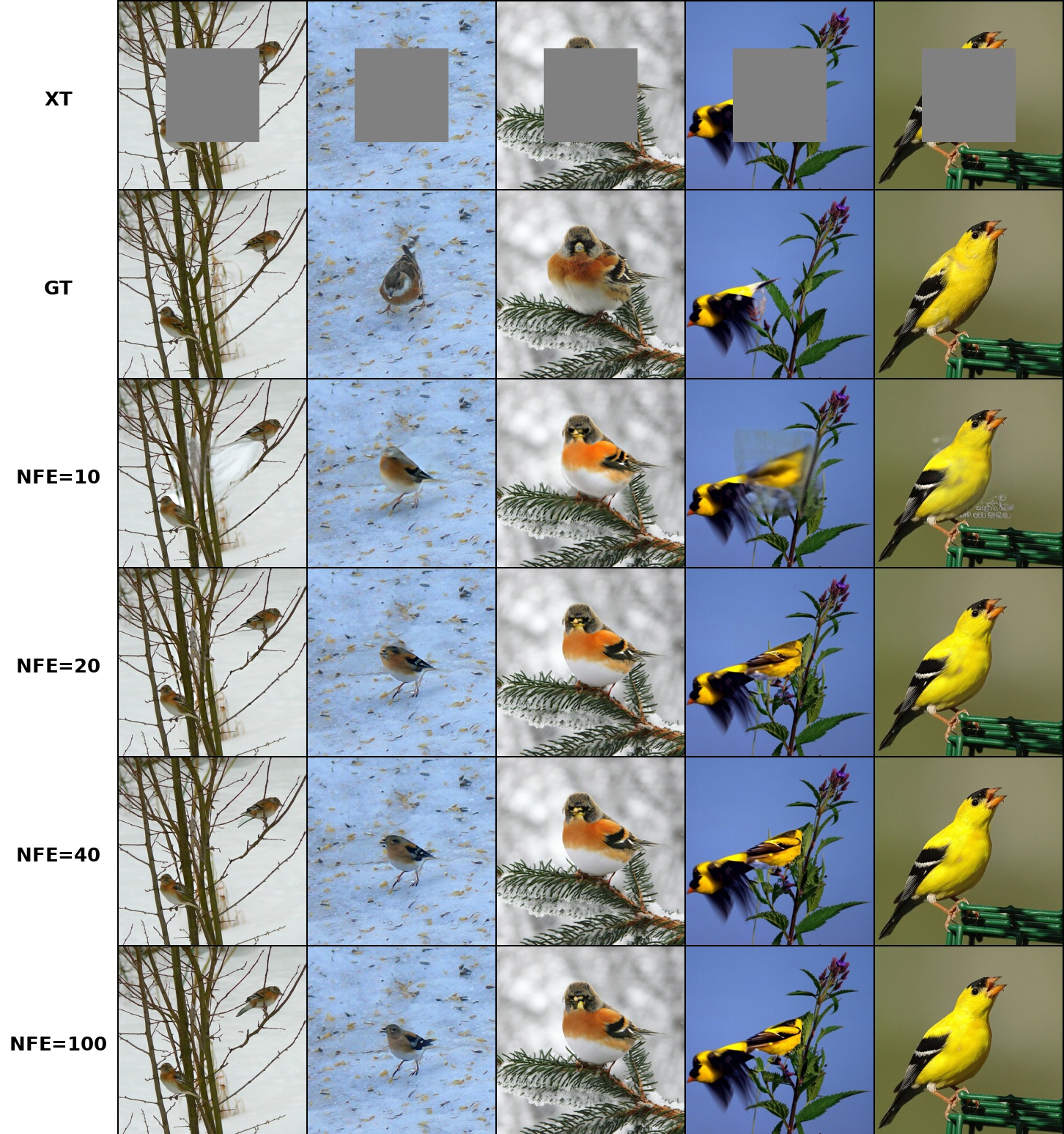}
    \end{minipage}
    
    \caption{\textbf{Qualitative Results on ImageNet (Part II).} Visual comparison of additional samples.}
    \label{fig:imagenet_part2}
\end{figure}

\end{document}